\documentclass[10pt,twocolumn,letterpaper]{article}

\usepackage{iccv}
\usepackage{times}
\usepackage{epsfig}
\usepackage{graphicx}
\usepackage{amsmath}
\usepackage{amssymb}
\usepackage{xcolor}
\usepackage{subcaption}
\usepackage{multirow}
\usepackage{booktabs}
\usepackage{makecell}
\usepackage{colortbl}
\usepackage{cuted}
\usepackage{textcomp}
\usepackage{enumerate}
\usepackage{algorithm}
\usepackage{algorithmic}
\usepackage{footnote}
\usepackage{enumitem}
\usepackage{indentfirst}
\usepackage{url}
\definecolor{mygray}{gray}{.90}
\graphicspath{{./figs/}}
\usepackage[accsupp]{axessibility}  % Improves PDF readability for those with disabilities.
\usepackage[breaklinks=true,bookmarks=false]{hyperref}

\iccvfinalcopy % *** Uncomment this line for the final submission

\ificcvfinal\fi

\newcommand{\name}{CAME}

\begin{document}

%%%%%%%%% TITLE
\title{CAME: Contrastive Automated Model Evaluation}
\author{
  Ru Peng$^{1}$\thanks{Both authors contributed equally to this research.},
  Qiuyang Duan$^{1}$\footnotemark[1],
  Haobo Wang$^{1}$, 
  Jiachen Ma$^{1,2}$, 
  Yanbo Jiang$^{1}$,\\
  Yongjun Tu$^{3}$,
  Xiu Jiang$^{3}$,
  Junbo Zhao$^{1}$\thanks{Corresponding author.}\and
  \textnormal{$^1$Zhejiang University \quad $^2$DI-Lab, Zhejiang University \quad $^3$OPPO} 
  \\
  \texttt{\small \{rupeng,duanqiuyang,wanghaobo,mjc\_zjdx,yanbojiang,j.zhao\}@zju.edu.cn}\\
  \texttt{\small \{tuyongjun1,jiangxiu\}@oppo.com}  
  % \\
  % \texttt{\small \urlstyle{tt}\textcolor{url_color}{\url{https://zju3dv.github.io/pvo/}}}
}

% \author{
%   Ru Peng$^{1}$\thanks{Both authors contributed equally to this research.}\and
%   Qiuyang Duan$^{1}$\footnotemark[1]\and 
%   Haobo Wang$^{1}$\and
%   Jiachen Ma$^{1,2}$ \and 
%   Yanbo Jiang$^{1}$\and
%   Yongjun Tu$^{3}$\and
%   Xiu Jiang$^3$\and
%   Junbo Zhao$^{1}$\thanks{Corresponding author.}\and
%   \\
%   \textnormal{$^1$Zhejiang University \quad $^2$DI-Lab, Zhejiang University \quad $^3$OPPO} 
%   \\ 
%   \texttt{\small \{rupeng,duanqiuyang,wanghaobo,mjc\_zjdx,yanbojiang,j.zhao\}@zju.edu.cn}
%   \\
%   \texttt{\small \{tuyongjun1,jiangxiu\}@oppo.com}  
%   % {\url{https://zju3dv.github.io/pvo/}}
% }

\maketitle
\ificcvfinal\thispagestyle{empty}\fi

%%%%%%%%% ABSTRACT
\begin{abstract}
The Automated Model Evaluation (AutoEval) framework entertains the possibility of evaluating a trained machine learning model without resorting to a labeled testing set.
Despite the promise and some decent results, the existing AutoEval methods heavily rely on computing distribution shifts between the unlabelled testing set and the training set. 
We believe this reliance on the training set becomes another obstacle in shipping this technology to real-world ML development.
In this work, we propose \textbf{C}ontrastive \textbf{A}utomatic \textbf{M}odel \textbf{E}valuation (\textbf{\name{}}), a novel AutoEval framework that is rid of involving training set in the loop.
The core idea of \name{} bases on a theoretical analysis which bonds the model performance with a contrastive loss.
Further, with extensive empirical validation, we manage to set up a predictable relationship between the two, simply by deducing on the unlabeled/unseen testing set.
The resulting framework \name{} establishes a new SOTA results for AutoEval by surpassing prior work significantly. \protect\footnotemark 
\end{abstract}
\footnotetext{Our code is publicly available at:  
\url{https://github.com/pengr/Contrastive_AutoEval}}

%%%%%%%%% BODY TEXT
\section{Introduction}
% autoeval 
During the last decade, the technological advancement of artificial intelligence and machine learning has attained unprecedented achievements, affecting a variety of domains or verticals.
Ubiquitously, off these milestones, to properly evaluate, assess and benchmark the trained models is undoubtedly pivotal, particularly when considering the deployment towards the production in real-world scenarios.
To do that, the traditional means often relies on a pre-split and static testing set for model evaluation, which is principally left out of the sight during training or validation phase.
However, several recent works has pointed out the drawback of this standardized scheme due to its requirement of careful sample selection, randomization due to the sample set split, the OOD gap between the deployment environment, and (somewhat) expensive label annotation~\cite{deng2020labels,deng2021does,chen2021mandoline}, etc.
Most recently, we see that \emph{Automated Model Evaluation} (AutoEval) has emerged to tackle these problems~\cite{deng2020labels}.

\begin{figure}[t]
    \centering
    \includegraphics[width=0.485\textwidth]{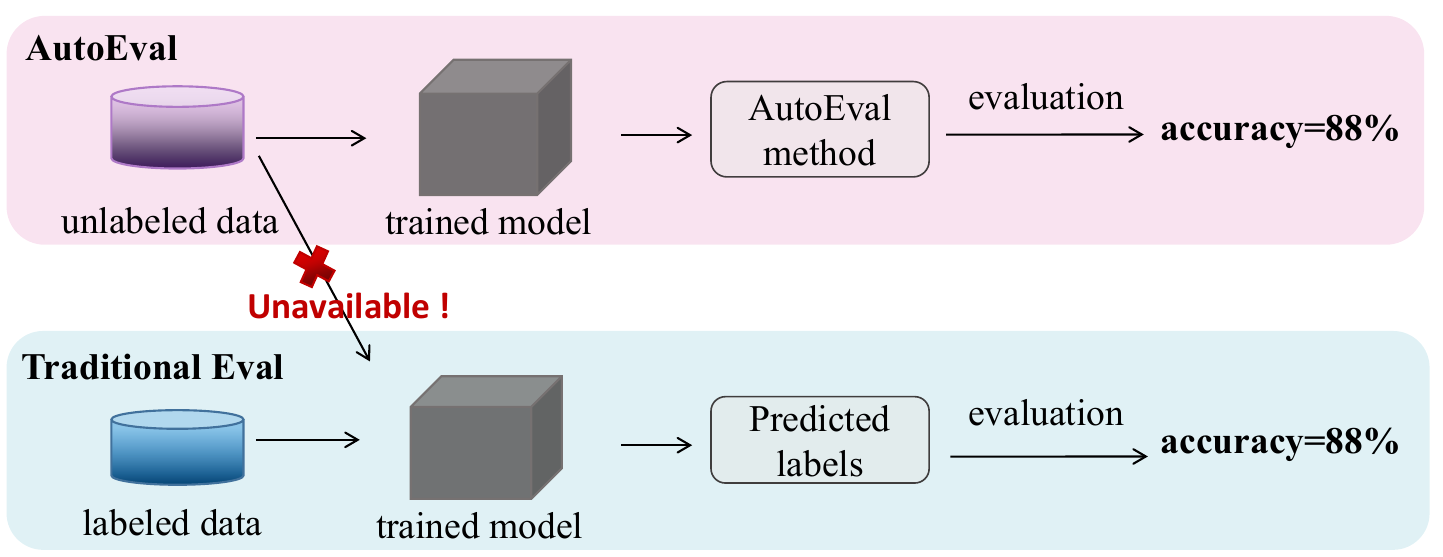}
    \caption{Illustration of workflow differences between AutoEval and traditional model evaluation.}
    \label{desc}
\end{figure}

% autoeval intro slightly
In particular, the vanilla prototype of the Automated Model Evaluation approaches aim at estimating a provided model's performance on an unlabeled testing set.
Notably, these approaches first generate a few \emph{meta-sets} by adopting pre-selected data augmentations on the training set. In what follows, one can estimate a certain distance --- for instance, the Frechet distance~\cite{dowson1982frechet} --- averaged between the meta-sets with the testing set.
As a result, the prior work has proactively shown that this averaged distance measurement is related to the final model performance on the testing set.
Indeed, we believe this setup of AutoEval on the testing set possesses positive prospects because it manifests a high similarity towards real production --- where the testing set is acquired on the fly in the real-world, leaving no time/space for these samples to be annotated or persist.
A graphical illustration of the AutoEval against the conventional static testing set evaluation is depicted in Figure \ref{desc}.

% problem of autoeval.
Despite its promise and prospect, we realize that the current paradigm of AutoEval may still fail in its real-world deployment, under certain conditions.
On one hand, it is widely acknowledged that the prior works are dedicated to avoiding annotating the testing samples and to amortizing the vexing randomness through the massive generation of meta-sets offline. 
On the other hand, however, these techniques still demand the full presence of the sample input from the training set, which in many --- if not most --- of the occasions probably imply expensive storage and computation cost.
Hence, we argue that this requirement cannot be easily ensured in many scenarios, most notably on limited-capacity, limited-storage, or low-power platforms such as edge devises for IOT or autonomous driving.
Hereby, we pose the core motivation of the design of this work: can we establish an  AutoEval framework without keeping the training set in the loop?

To reach this target is not trivial, and it cannot be achieved by incrementally changing the prior method.
This is mostly due to the heavy bond of the final model performance regressor with the meta-sets induced from the training data. 
In this work, we hope to break this paradigm commonly used in prior work, and propose a novel paradigm --- \textbf{C}ontrastive \textbf{A}utomatic \textbf{M}odel \textbf{E}valuation, dubbed  \textbf{(\name)}. 
Unlike the previous approaches, \name{} aims to regress to the final model performance that assumes the absence of the training data.
In particular, \name{} is very much motivated by the following series of the theories:
\newtheorem{assumption}{Assumption}[section]
\newtheorem{theorem}{Theorem}[section]
\newtheorem{lemma}[theorem]{Lemma}

\begin{theorem} \label{Guarantees for the Optimal Encoder}
\cite{wang2022chaos}
Given a model $f$ with an optimal functional  minimizer $f^*=\arg\min\mathcal{L}_{NCE}(f)$, its classification risk can be upper- and lower-bounded by its contrastive learning risk as
\begin{equation}
\begin{aligned}
\mathcal{L}_{NCE}(f^*)-\mathcal{O}(M^{-1/2})&\leq \mathcal{L}_{CE}^{\mu}(f^*)+\log(M/K) \\
&\leq\mathcal{L}_{NCE}(f^*)+\mathcal{O}(M^{-1/2})
\end{aligned}
\end{equation}
where $M$ is the number of negative samples in contrastive learning, $K$ is the number of classes, $\mathcal{L}_{NCE}$~\cite{oord2018representation} is the InfoNCE loss for contrastive learning, and $\mathcal{L}_{CE}^{\mu}$~\cite{wang2022chaos} is the mean CE Loss used to indicate the downstream classification risk (definitions in section 3).
\end{theorem}
Theorem \ref{Guarantees for the Optimal Encoder} indicate that under mild assumptions, the contrastive loss constantly bounded the cross-entropy loss and thus, can reflect the overall trends of generalization. 
Moreover, analogous theoretical guarantees of bounding CE loss through CL risk are also evident in \cite{saunshi2019theoretical} and \cite{bao2022surrogate}.
Notably, in the AutoEval problem, with distribution shifts and the absence of ground-truth labels on the test sets, the cross-entropy loss $\mathcal{L}_{CE}$ is inaccessible. Fortunately however, $\mathcal{L}_{NCE}$ is self-supervised and can be inferred purely from testing inputs.
Based on the theoretical analysis, we cast a hypothesis as follows. The contrastive loss --- calculated from the testing set alone --- is informative towards predicting the performance of the provided model.

To this regard, we briefly introduce our framework, \name{}. It is prerequisite composed of two conditions: (i)-the model is trained jointed of a normal task loss together with a contrastive loss and (ii)-the model performance is not affected by jointly contrastive learning.
Based on the model yielded this way, we conduct a rigorous empirical study --- guided by the theories we pose above --- that we prove the correlation between the contrastive loss on the testing set with its performance truly exists.
The AutoEval established this way enjoys the following two major merits: (i)-it shreds the need for the training set during the evaluation phase which further extends the AutoEval technique towards production in real-world; (ii)-\name{} sets a new record of testing performance estimation by exceeding the prior work by significant margins.

\begin{figure*}[t]
    \centering
    \begin{subfigure}{0.195\textwidth}
        \centering
        \includegraphics[width=\textwidth]{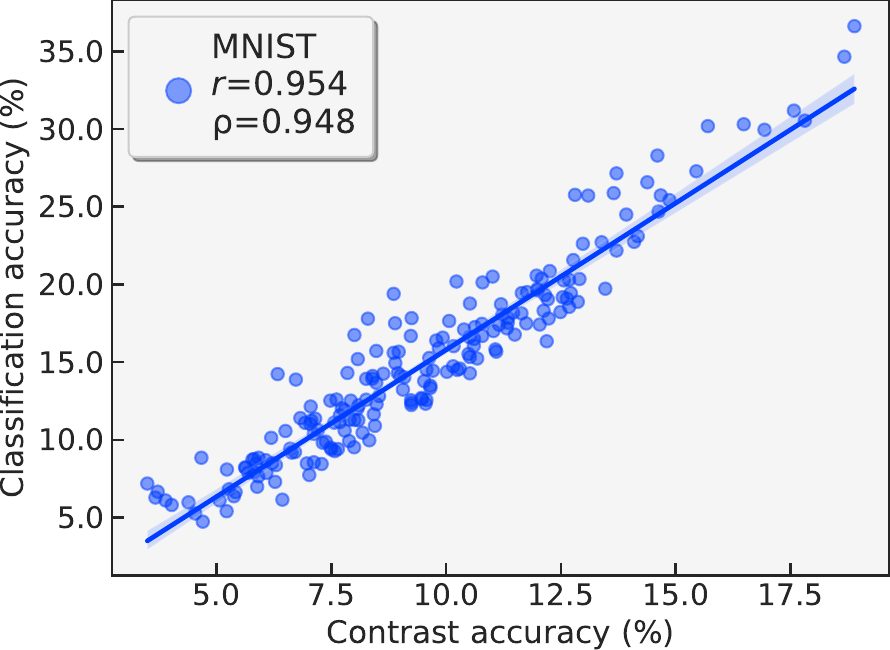}
        \caption{MNIST}
    \end{subfigure}
    \begin{subfigure}{0.195\textwidth}
        \centering
        \includegraphics[width=\textwidth]{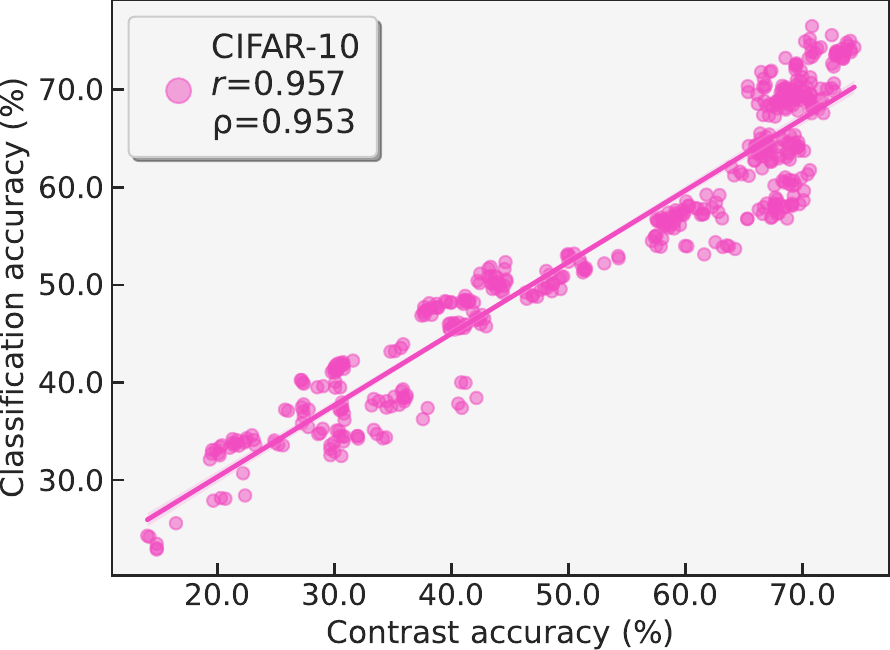}
        \caption{CIFAR-10}
    \end{subfigure}
    \begin{subfigure}{0.195\textwidth}
        \centering
        \includegraphics[width=\textwidth]{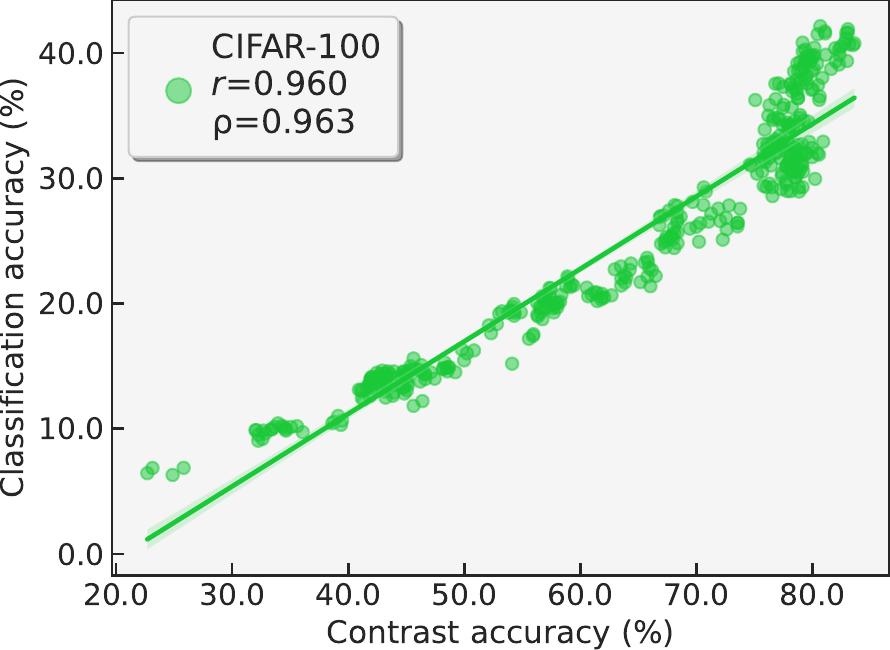}
        \caption{CIFAR-100}
    \end{subfigure}
    \begin{subfigure}{0.195\textwidth}
        \centering
        \includegraphics[width=\textwidth]{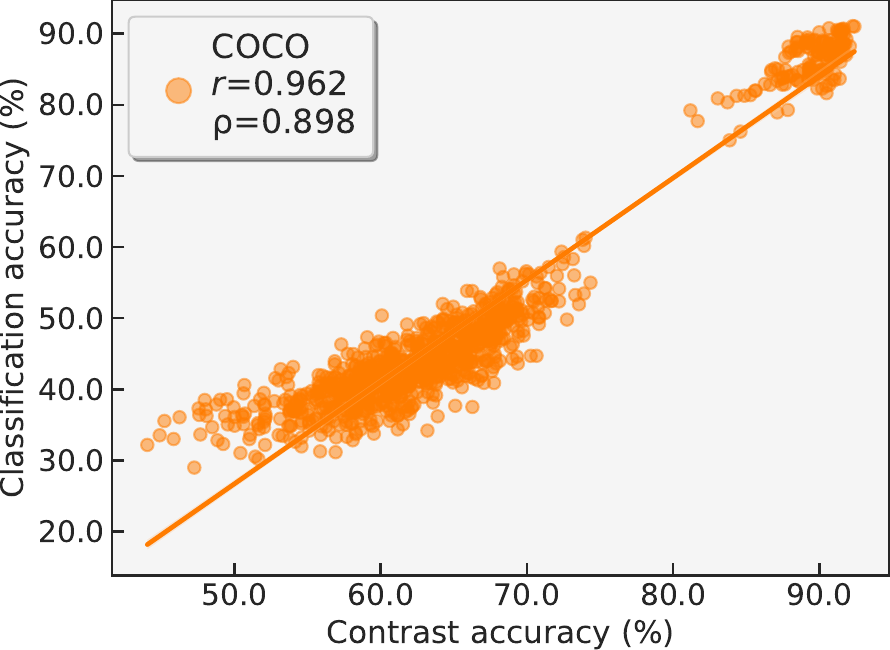}
        \caption{COCO}
    \end{subfigure}
    \begin{subfigure}{0.195\textwidth}
        \centering
        \includegraphics[width=\textwidth]{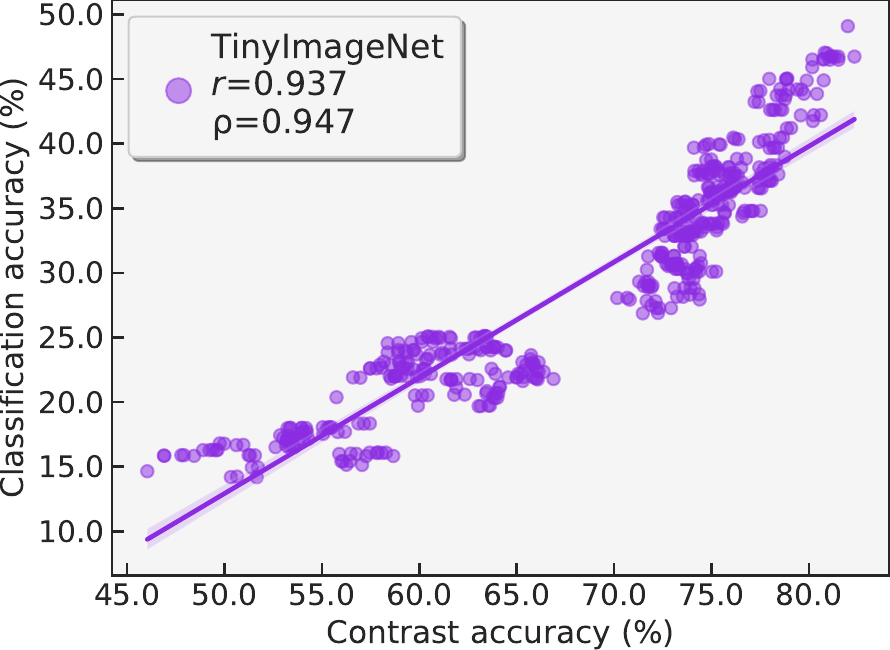}
        \caption{TinyImageNet}
    \end{subfigure}
    \caption{Experimental results about the correlation study. On the above different datasets, we show that there exists strong linear correlation between contrastive accuracy (x-axis) and classification accuracy (y-axis). The symbols $r$ represents Pearson's correlation coefficient and $\rho$ indicates Spearman's rank correlation coefficient.}
    \label{corr}
    % \vspace{-0.5cm}
\end{figure*}

\section{Related Works}
\textbf{Automated Model Evaluation.}
\cite{deng2020labels,deng2021does,sun2021label,yu2022predicting} build regression models on many test sets with distribution shifts to predict model's accuracy on an unlabeled test set.
\cite{guillory2021predicting,garg2022leveraging} use confidence based methods, .
Our work does not use distribution shift measurements. Instead, we directly use contrastive accuracy to regress classification accuracy.

\textbf{Model Generalization Prediction.}
This problem is to estimate the generalization gap and predict the generalization error. From model perspective, \cite{corneanu2020computing,jiang2018predicting,jiang2021assessing} predict generalization error by leveraging model parameters. 
From data perspective, \cite{chuang2020estimating,zilly2019frechet} predict the generalization gap under distribution shifts via data representations.
Different from these works, our work aims to directly predict a model's accuracy on unseen unlabeled test sets.

\textbf{Out-of-Distribution Detection.}
The OOD Detection task \cite{hendrycks2016baseline,liang2017enhancing,liu2020energy,hendrycks2019scaling} is to detect test samples subject to a distribution different from the training data distribution.
It focuses on the data distribution of training data and testing data.
Different from this, our work focuses on estimating model's accuracy on unlabeled OOD test set.

\textbf{Contrastive Learning.}
Contrastive Learning (CL) \cite{chen2020simple,he2019moco,chen2020mocov2,grill2020bootstrap,caron2020unsupervised,chen2021exploring} is a typical self-supervised learning paradigm to learn effective representation of input samples.
Used as a pre-training task, it can significantly enhance the downstream semantic classification task, which means CL learns information-rich features for object classification. 
Inspired by these pioneering efforts, we choose contrastive learning as the auxiliary learning task, and take the classical CL framework (SimCLR) to validate the feasibility of CL in this work.

\textbf{Unsupervised Domain Adaptation.} 
Unsupervised domain adaptation is also an active research field, which aims to use labeled source data and unlabeled target data to learn a model generalizing well from the source domain to the target domain. 
In recent years, many researches such as \cite{sun2016return,long2015learning,tanwisuth2021prototype,li2021implicit,huang2021model} have proposed different measures and frameworks to promote the development of this field. 
In this work, given a model trained on the source dataset, we focus on obtaining a precise estimation of its accuracy on unlabeled target sets.

\section{Contrastive Automated Model Evaluation}
In this section, we describe our proposed method, \name{}, in detail.
The core idea of \name{} is simple.
We first formulate a multi-task learning framework by integrating a normal task loss with a contrastive learning loss objective.
After attaining the model through regular optimization process, on an unseen and unlabeled testing set, we build a simple and separate neural network to regress from the contrastive loss to a proximal model performance.
A pseudo code is provided in algorithm \ref{alg1}.
% \vspace{-0.3cm}
\begin{algorithm}[t]
   %\textsl{}\setstretch{1.8}  % ajust line spacing
   \renewcommand{\algorithmicrequire}{\textbf{Input:}}
   \renewcommand{\algorithmicensure}{\textbf{Output:}}
   \caption{Contrastive Automated Model Evaluation}  
   \label{alg1}
   \begin{algorithmic}[1]
        \REQUIRE Training set $\mathcal{D}_{o}$, seed set $\mathcal{D}_{e}$, unseen test set $\mathcal{D}_{t}$; shared encoder $f$, projection heads $h$, $g$; contrastive loss weight $\lambda$; data transformations $t \sim \mathcal{T}_{s}$
        
        \textcolor{cyan}{// Multi-task learning}
        \FOR{$epoch = 1, 2, ... T$}
        \STATE sample the mini-batch $\mathcal{D}_{l}$, $\mathcal{D}_{u}$ from $\mathcal{D}_{o}$
        \FOR{$x\in\mathcal{D}_l$}
        \STATE $z = g(f(x)) \in\mathbb{S}^{2N-1}$
        \STATE $Wf(x) = h(f(x)) \in\mathbb{R}^{K}$
        \STATE $\mathcal{L}=\mathcal{L}_{CE}+\lambda\mathcal{L}_{NCE}$, $\mathcal{L}_{CE}$ and $\mathcal{L}_{NCE}$ in Sec. \ref{eq:5}
        \ENDFOR{}
        \ENDFOR{}

        \textcolor{cyan}{// Synthesizing Sample Sets}
        \FOR{$i = 1,2,..., a$}
        \STATE    $\mathcal{D}_{s}^{i} = {t}_{i}(\mathcal{D}_{e})$
        \STATE    $Acc_{con}^{i}= \sum_{i=0}^{N-1} \mathbb{I}\left[i = \mathop{\arg\max}\limits_{j \in [0, N-1]} g_{j}(f(D_{s}^{i}))\right] /N$
        \STATE    $Acc_{cla}^{i}= \sum_{i=0}^{N-1} \mathbb{I}\left[y = \mathop{\arg\max}\limits_{i \in \{1,2,\cdots,K\}} h_{i}(f(D_{s}^{i}))\right] /N$
        \ENDFOR{}
        \STATE Correlation Coefficients: $r$, $\rho$ statistic from $\mathcal{D}_{s} = \{(Acc_{con}^1,Acc_{cla}^1), \ldots, (Acc_{con}^a,Acc_{cla}^a)\}$
        \STATE Linear regressor $R: Acc_{cla} = W^\top(Acc_{con}) + b$

       \textcolor{cyan}{// Automated Model Evaluation via Regression}
        \STATE In $\mathcal{D}_{t}: \hat{Acc_{cla}} = R(Acc_{con})$; $Acc_{con}$, $Acc_{cla}$ in Eq. \ref{eq:7}
        \STATE Mean Absolute Error: $\varepsilon=|Acc_{cla}-\hat{Acc_{cla}}|$
        \ENSURE Pearson’s correlation $r$, Spearman’s rank correlation $\rho$ and Mean Absolute Error $\varepsilon$.
    \end{algorithmic}
\end{algorithm}

\subsection{Problem Definition}
Consider a image classification task, we aim to estimate the classification performance of a trained classifier on different unseen and unlabeled test sets automatically. 
We denote the training set as $\mathcal{D}_{o}=\{(x_i,y_i)\}_{i=1}^I$, where $y_i\in \{1,2,\cdots,K\}$ denotes the  label of the image $x_i$. 
These unseen and unlabeled test set are denoted as 
$D_{t}=\{\{x_j\}_{j=1}^{J_1}, \ldots, \{x_j\}_{j=1}^{J_M}\}$, 
where $\{x_j\}_{j=1}^{J_M}$ represent the $m$-th unseen test set.
Based on the theoretical anaylsis in Theorem \ref{Guarantees for the Optimal Encoder} --- ``\textit{For the contrastive learning model, its classification risk can be upper and lower bounded by its contrastive risk in any unseen test data distribution}'', which means that it is feasible to predict classification accuracy with contrastive accuracy. 
Motivated by this exhilarating finding, in unseen test set $D_{t}^{m}=\{x_j\}_{j=1}^{J_m}$, we fit a linear regressors \cite{huber2004robust} $R: (f, g, D_{t}^{m}) \rightarrow \hat{Acc}$ to predict a classifier's accuracy $\hat{Acc_{cla}}$ by its contrastive accuracy:
\begin{equation}
    \hat{Acc_{cla}}=R(f, g, D_{t}^{m})=R(Acc_{con}),\label{eq:3}
\end{equation}
where $f, g$ are the shared encoder and contrastive projection head in Section \ref{section3_3}.

\subsection{Correlation Analysis}
To further corroborate the feasibility of predicting classification accuracy from contrastive learning accuracy, we analyze the correlation between the two sources of accuracies among various data setup in Figure \ref{corr}.
For each data setup, we train \name{} on the training set, then evaluate its accuracy pairs on each synthetic sample set (mentioned in Section \ref{section3_4}).
Finally, we report the scatter plot and both Pearson's correlation coefficient $r$ and Spearman's rank correlation coefficient $\rho$. 
Here, each data point in the scatter plot corresponds to the accuracy pair of a synthetic sample set.
From Figure \ref{corr}, we can see that among various data environments, the contrastive learning accuracy and classification accuracy exhibit strong linear correlation ($r>0.937$ and $\rho>0.898$). 
This finding prompts us to train a linear regressor to predict classification accuracy on unlabeled unseen test distribution, rather than building a non-linear MLP like previous work.

\subsection{Multi-task Learning}\label{section3_3} % co-training with Contrastive Learning
\begin{figure}[t]
    \centering
    \includegraphics[width=0.48\textwidth]{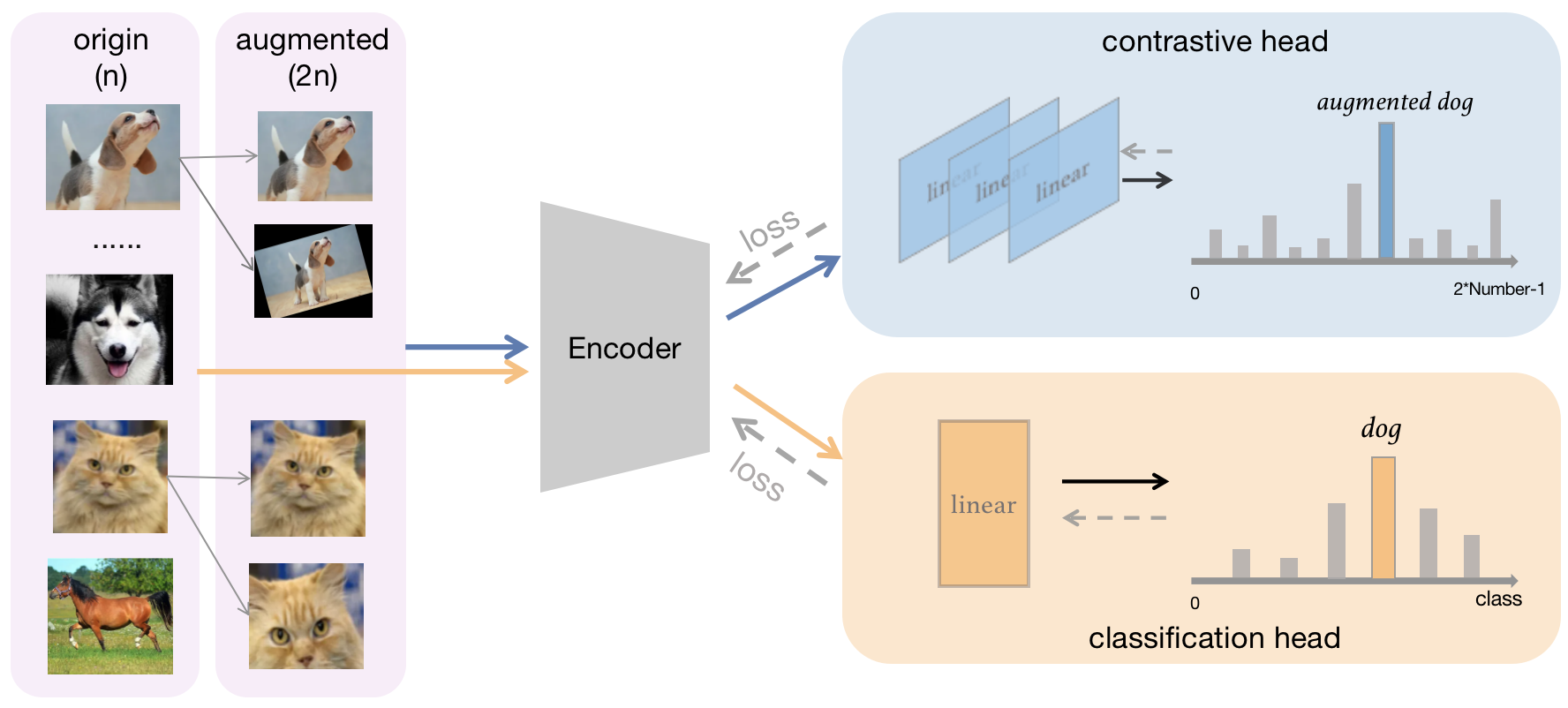}
    \caption{The model architecture of our multi-task learning. Here, we adopt SimCLR \cite{chen2020simple} as the contrastive learning framework. 
    % A CNN backbone is used as a shared encoder to extract image features, followed by two projection heads for contrastive learning and semantic classification task. 
    Note that we only consider the original batch when calculating the classification accuracy.}
    \label{framework}
    \vspace{-0.5cm}
\end{figure}

In this section, we begin by introducing how CAME trains the model via pivotal multi-task learning, to reveal the inherent strong correlation between classification accuracy and contrastive learning accuracy.
% In proposed CAME, to learn a model that reveals the inherent strong correlation between classification accuracy and contrastive learning accuracy, we begin by introducing the basic notations and pipeline of the pivotal multi-task learning. 
In summary, the co-training paradigm consists of the supervised classification and the self-supervised contrastive learning.
Specifically, given a mini-batch of \textit{N} labeled samples $\mathcal{D}_l=\left\{\left(x_i, y_i\right)\right\}_{i=1}^N$ where labels $y_i \in\{1, \ldots, K\}$ and \textit{N} unlabeled samples $\mathcal{D}_u=\left\{x_i\right\}_{i=1}^N$.
We adopt a network-unrestricted encoder $f \in \mathcal{F}: \mathbb{R}^n \rightarrow \mathbb{R}^d$ to extract image representations from input samples. 
Then, we apply two projection heads $g: \mathbb{R}^d \rightarrow \mathbb{S}^{2N-1}$
and $h: \mathbb{R}^d \rightarrow \mathbb{R}^K$ to map the shared representations into two spaces where contrastive loss and classification loss are applied, respectively.
The illustration of our model is shown in Figure \ref{framework}.

\textbf{Contrastive Learning.}
Taken a training sample $x\in\mathcal{D}_u$, we apply a set of random data augmentation $t \sim \mathcal{T}$ to generate its positive sample $x^{+} = t(x)$, and treat the other augmented samples $\left\{x_i^{-}\right\}_{i=1}^{2(N-1)}$ within a mini-batch as its negative samples.
Then, the shared encoder $f$ can be learned by the InfoNCE contrastive loss \cite{oord2018representation} which mapping from the \textit{d}-dimensonal image representations to a unit hypersphere:
\begin{gather}
\mathcal{L}_{\mathrm{NCE}}(f, g)=\mathop{\mathbb{E}}\limits_{p\left(x, x^{+}\right)} \mathop{\mathbb{E}}\limits_{\left\{p\left(x_i^{-}\right)\right\}}\left[-\log \frac{\exp \left(z^{\top} z^{+}\right)}{\sum\limits_{i=1}^{2N-1} \exp \left(z^{\top} z_i^{-}\right)}\right],\label{eq:4}
\end{gather}
where $z=g(f(x))$, $p\left(x\right)$ is the data distribution, and $p\left(x, x^{+}\right)$ is the joint distribution of positive data pairs.

\textbf{Classification Learning.} 
Given a labeled data pairs $(x, y) \in \mathcal{D}_l$, we usually use the cross-entropy (CE) loss to train the shared encoder $f$ through classification learning:
\begin{gather}
\mathcal{L}_{\mathrm{CE}}(f, h)=\mathbb{E}_{p(x, y)}\left[-\log \frac{\exp \left(f(x)^{\top} w_y\right)}{\sum_{i=1}^K \exp \left(f(x)^{\top} w_i\right)}\right],
\end{gather}\label{eq:5}
where $h=\left[w_1, w_2, \ldots, w_K\right]$ is the classification head.

Hereafter, we train our classifier based on the aforementioned multi-task learning paradigm, by minimizing the following loss:
\begin{equation}
    \mathcal{L}=\mathcal{L}_{CE}+\lambda\mathcal{L}_{NCE},
\end{equation}
where $\lambda$ is a weighing coefficient.

\subsection{Synthesizing Sample Sets}\label{section3_4}
To fit the linear regressor (Eq. \ref{eq:3}) for performance prediction, we need to collect the contrastive accuracy and the counterpart classification accuracy in various test environments by the multi-task-learning based model.
For this, we require many test sets which should include:
\emph{i)-various distributions,}
\emph{ii)-the same classes as the training set but without image overlap,}
and \emph{iii)-sufficient samples.} 
However, one obstacle still is to collect such testing sets from natural distributions.
As a surrogate, we synthesize these ``sample sets''. 
Overall, we synthesize these sample sets by applying a combination of transformations $t \sim \mathcal{T}_{s}$ on a seed dataset.
Note that all possibilities of the transformation sequences are not calculated by permutation and combination, so there are many random states when applying these transformations.
Therefore, we generate image sets of various distributions.
For all the data transformation setup, we inherit the labels from the seed dataset directly because the transformations do not alter the profound semantics.
Meanwhile, We generate 400 synthetic sample sets to calculate the accuracy pairs $\{(Acc_{con}^1,Acc_{cla}^1), \ldots, (Acc_{con}^a,Acc_{cla}^a)\}$, which is applied to fit a linear regression model and statistic the correlation among these accuracy pairs. $a$ is the amount of synthetic sample sets, the default value is set as 400.
Specific setups are elaborated as follow.

\begin{itemize}[left=0pt]
    \item
    \textbf{MNIST setup.} 
    We synthesize sample sets by applying various transformations on MNIST test dataset. 
    MNIST as simple gray-scale images, we first consider change their black background to color ones. 
    Specifically, for each sample of MNIST, we randomly select an image from COCO dataset and randomly crop a patch, which is then used as the background of the hand-written digit sample. 
    After that, we randomly apply three out of the six transformations on the background-replaced images: \{\emph{autoContrast, rotation, color, brightness, sharpness, translation}\}.
    \item
    \textbf{CIFAR-10 setup.}
    We synthesize sample sets based on CIFAR-10 test set by adopting three image transformations on it. 
    The three transformations are randomly selected from the six transformations of MNIST Setup. 
    \item
    \textbf{CIFAR-100 setup.}
    We study on CIFAR-100 to explore if a strong correlation still exists between contrastive learning and classification accuracy in more semantic classes case. 
    The applied transformations are followed as CIFAR-10 setup. 
    Here, we use the training split of CIFAR-100 dataset to train our model and the testing split to synthesize the various sample sets.
    \item
    \textbf{COCO setup.}
    Following the practice in \cite{deng2021does} and \cite{peng2017visda}, we select 12 common classes: aeroplane, bike, bird, boat, bottle, bus, car, dog, horse, monitor, motorbike and person. 
    Resort to the annotation boxes, we crop out the objects of images belonging to these categories in the COCO training set and validation set to build training and testing set for the classification task.
    Based on the testing set, we use 15 common corruptions from CIFAR-10-C to generate 60 sample sets, and randomly use three transformations in MNIST setup to generate the rest of sample sets.
    \item
    \textbf{TinyImageNet setup.}
    The synthetic sample sets of TinyImageNet are synthesized from its validation split.
    As in the MNIST setup mentioned above, we randomly use three kinds of transformations to synthesize sample sets for linear regression and correlation study.
\end{itemize}

\subsection{Automated Model Evaluation via Regression}
As the final step, we propose to use contrastive learning accuracy to regress to the model's classification accuracy on unseen testing set.
% fuck this drives me crazy...
Notably, for the input and output side of training the regressor, we adopt the soft version attained from contrastive learning and classification tasks.
On the input side, we define it as the probability of an augmented sample being positive-sample.
For the output, we use the confidence value of the sample being corrected predicted.

Consequently we write down the forms as follow:
\begin{gather}
    % Acc_{con}=\sum_{i=0}^{N-1} \mathbb{I}(\hat{i}=i)/N, \\
    % Acc_{cla}=\sum_{i=0}^{N-1} \mathbb{I}(\hat{y}=y))/N, \\
    Acc_{con}= \sum_{i=0}^{N-1} \mathbb{I}\left[i = \mathop{\arg\max}\limits_{j \in [0, N-1]} g_{j}(f(x))\right] /N, \\
    Acc_{cla}= \sum_{i=0}^{N-1} \mathbb{I}\left[y = \mathop{\arg\max}\limits_{i \in \{1,2,\cdots,K\}} h_{i}(f(x))\right] /N \\
    \varepsilon=|Acc_{cla}-\hat{Acc_{cla}}|,
    %=|Acc_{cla}-R(Acc_{con})|
    \label{eq:7}
\end{gather}
where $\mathbb{I}\left[\cdot\right]$ is an indicator function, $N$ is the mini-batch size, $i$ is the index of an image in original batch, $y$ is the ground-truth class label. $\varepsilon$ is the mean absolute error (MAE) for classification accuracy estimation.

\begin{table}[t]
\centering
\caption{Pearson’s correlation ($r$) and Spearman’s rank correlation ($\rho$) on the different data setup (higher is better). ``-'' indicates that the results are not reported in original paper.}\label{table_2}
\renewcommand{\arraystretch}{1.1} %adjust line spacing
\setlength{\tabcolsep}{0.8pt}
\resizebox{0.48\textwidth}{!}
{{
\begin{tabular}{c|cc|cc|cc|cc}
\Xhline{0.8pt}
\multirow{2}{*}{Method} & \multicolumn{2}{c|}{MNIST} & \multicolumn{2}{c|}{CIFAR-10} & \multicolumn{2}{c|}{CIFAR-100} & \multicolumn{2}{c}{COCO} \\ 
\cline{2-9} 
 & \multicolumn{1}{c|}{$\mathnormal{r}$} & $\rho$ & \multicolumn{1}{c|}{$\mathnormal{r}$} & $\rho$ & \multicolumn{1}{c|}{\textbf{$\mathnormal{r}$}} & $\rho$ & \multicolumn{1}{c|}{$\mathnormal{r}$} & $\rho$ 
 \\ \hline
Frechet \cite{deng2020labels} & \multicolumn{1}{c|}{0.912} & - & \multicolumn{1}{c|}{-} &  & \multicolumn{1}{c|}{-} & - & \multicolumn{1}{c|}{0.908} & - \\ 
\hline
Rotation \cite{deng2021does} & \multicolumn{1}{c|}{-} & 0.960 & \multicolumn{1}{c|}{-} & 0.981 & \multicolumn{1}{c|}{-} & 0.950 & \multicolumn{1}{c|}{-} & 0.881 \\ 
\hline
Jigsaw \cite{noroozi2016unsupervised} & \multicolumn{1}{c|}{-} & - & \multicolumn{1}{c|}{-} & 0.958 & \multicolumn{1}{c|}{-} & - & \multicolumn{1}{c|}{-} & - \\ 
\hline
\rowcolor{mygray}
\textbf{\name{} (ours)} & \multicolumn{1}{c|}{0.948} & 0.954 & \multicolumn{1}{c|}{0.953} & 0.957 & \multicolumn{1}{c|}{0.963} & 0.960 & \multicolumn{1}{c|}{0.898} & 0.962 \\ 
\Xhline{0.8pt}
\end{tabular}%
}}
% \vspace{-0.5cm}
\end{table}

\section{Experiments}
\subsection{Experimental Setup}
We conduct extensive experiments on various datasets.
Similar to prior work, the final assessment of \name{} is based on a homogeneous but different unseen test environment.
For instance, in hand-written datasets, we train our model and regressor on MNIST but test them in SVHN~\cite{netzer2011reading} and USPS~\cite{hull1994database}. The differing distribution of the training and testing set is significant but the tasks are indeed homogeneous.
This protocol effectively validates the generalizability of AutoEval approaches.
 
For natural images, we train DenseNet-40-12 \cite{huang2017densely} on CIFAR-10 \cite{krizhevsky2009learning}, then conducting test on CIFAR-10.1 \cite{recht2018cifar} and CIFAR-10-C \cite{hendrycks2019benchmarking}. 
For CIFAR-100 setup \cite{krizhevsky2009learning}, we keep the setttings on CIFAR-10, and test it on CIFAR100-C. 
For COCO setup \cite{lin2014microsoft}, following \cite{deng2021does}, we choose 12 object classes (aeroplane, bicycle, bird, boat, bottle, bus, car, dog, horse, tv-monitor, motorcycle, person). 
To build the training set, object annotation boxes among these 12 classes are cropped from the COCO training images containing them. These images are used to train a ResNet-50 backbone \cite{he2016deep}. 
Similarly, we build unseen testing sets for classification accuracy from Caltech-256 \cite{griffin2007caltech}, PASCAL VOC 2007 \cite{everingham2010pascal}, ImageNet \cite{deng2009imagenet}, which carrying the same 12 categories. 
For TinyImageNet setup \cite{le2015tiny}, we train ResNet-50 and the testing is conducted on TinyImageNet-C.

\begin{table*}[t]
\centering
\caption{
Mean absolute error (MAE) results for evaluating the classifier accuracy on the unseen test sets. The training set of each group is MNIST, CIFAR-10, CIFAR-100, COCO and TinyImageNet, respectively. 
In MNIST group, the merged cells represent the average MAE value among the SVHN and USPS. 
``-'' indicates that the results are not reported in original paper. 
The time cost for different algorithms we count at here\protect\footnotemark.
}
\label{table_1}
\resizebox{\textwidth}{!}{%
\begin{tabular}{c|cc|cc|c|ccc|c}
\Xhline{0.8pt}
\multirow{2}{*}{Method} & \multicolumn{2}{c|}{MNIST} & \multicolumn{2}{c|}{CIFAR-10} & CIFAR-100 & \multicolumn{3}{c|}{COCO} & TinyImageNet \\ \cline{2-10} 
 & \multicolumn{1}{c|}{SVHN} & USPS & \multicolumn{1}{c|}{CIFAR-10.1} & CIFAR10-C & CIFAR100-C & \multicolumn{1}{c|}{Caltech} & \multicolumn{1}{c|}{Pascal} & ImageNet & TinyImageNet-C \\ 
 \hline
Pred ($\tau$ = 0.8) \cite{hendrycks2016baseline,liang2017enhancing} & \multicolumn{1}{c|}{10.58} & 21.18 & \multicolumn{1}{c|}{3.00} & 1.82 & - & \multicolumn{1}{c|}{3.25} & \multicolumn{1}{c|}{2.45} & 2.66 & 6.21 \\ 
\hline
Pred ($\tau$ = 0.9) \cite{hendrycks2016baseline,liang2017enhancing} & \multicolumn{1}{c|}{0.99} & 35.13 & \multicolumn{1}{c|}{1.30} & 1.26 & - & \multicolumn{1}{c|}{8.31} & \multicolumn{1}{c|}{8.43} & 8.00 & 8.54 \\ 
\hline
Entropy ($\tau$ = 0.2) \cite{kendall2017uncertainties} & \multicolumn{1}{c|}{3.57} & 32.29 & \multicolumn{1}{c|}{1.05} & 1.80 & - & \multicolumn{1}{c|}{5.81} & \multicolumn{1}{c|}{6.29} & 5.33 & 8.48 \\ 
\hline
Ens. AC \cite{lakshminarayanan2017simple} & \multicolumn{1}{c|}{80.05} & 12.84 & \multicolumn{1}{c|}{-} & 23.7 & - & \multicolumn{1}{c|}{-} & \multicolumn{1}{c|}{-} & - & - \\ 
\hline
Proxy Risk \cite{chuang2020estimating} & \multicolumn{1}{c|}{13.20} & 1.21 & \multicolumn{1}{c|}{-} & 5.3 & - & \multicolumn{1}{c|}{-} & \multicolumn{1}{c|}{-} & - & - \\ 
\hline
Ens. RI \cite{chen2021detecting} & \multicolumn{1}{c|}{79.56} & 8.01 & \multicolumn{1}{c|}{-} & 14.9 & - & \multicolumn{1}{c|}{-} & \multicolumn{1}{c|}{-} & - & - \\ 
\hline
Ens. RM \cite{chen2021detecting} & \multicolumn{1}{c|}{3.88} & 0.65 & \multicolumn{1}{c|}{-} & 2.2 & - & \multicolumn{1}{c|}{-} & \multicolumn{1}{c|}{-} & - & - \\ 
\hline
Frechet \cite{deng2020labels} & \multicolumn{1}{c|}{0.82} & 13.94 & \multicolumn{1}{c|}{0.96} & 1.94 & - & \multicolumn{1}{c|}{13.63} & \multicolumn{1}{c|}{2.26} & 5.64 & 8.13 \\ 
\hline
Frechet + $\mu$ + $\sigma$ \cite{deng2020labels} & \multicolumn{1}{c|}{2.06} & 0.03 & \multicolumn{1}{c|}{0.83} & 1.83 & - & \multicolumn{1}{c|}{4.01} & \multicolumn{1}{c|}{1.63} & 2.99 & 7.96 \\ 
\hline
Rotation \cite{deng2021does} & \multicolumn{1}{c|}{1.78} & 12.42 & \multicolumn{1}{c|}{3.74} & 1.99 & - & \multicolumn{1}{c|}{1.91} & \multicolumn{1}{c|}{2.86} & 3.15 & 8.21 \\ 
\hline
SSDR \cite{sun2021label} & \multicolumn{1}{c|}{0.76} & - & \multicolumn{1}{c|}{0.74} & 1.28 & - & \multicolumn{1}{c|}{-} & \multicolumn{1}{c|}{-} & - & 5.95 \\
\hline
AC \cite{hendrycks2016baseline, elsahar2019annotate} & \multicolumn{2}{c|}{21.17} & \multicolumn{1}{c|}{9.88} & 16.50 & 23.61 & \multicolumn{1}{c|}{-} & \multicolumn{1}{c|}{-} & - & 32.44 \\ 
\hline
IM \cite{chen2021mandoline} & \multicolumn{2}{c|}{18.48} & \multicolumn{1}{c|}{6.60} & 12.33 & 13.69 & \multicolumn{1}{c|}{-} & \multicolumn{1}{c|}{-} & - & 19.86 \\ 
\hline
DOC \cite{guillory2021predicting} & \multicolumn{2}{c|}{20.19} & \multicolumn{1}{c|}{7.25} & 13.87 & 14.60 & \multicolumn{1}{c|}{-} & \multicolumn{1}{c|}{-} & - & 25.02 \\ 
\hline
GDE \cite{jiang2021assessing} & \multicolumn{2}{c|}{24.42} & \multicolumn{1}{c|}{4.77} & 6.55 & 9.85 & \multicolumn{1}{c|}{-} & \multicolumn{1}{c|}{-} & - & 5.41 \\ 
\hline
ATC-MC \cite{garg2022leveraging} & \multicolumn{2}{c|}{5.02} & \multicolumn{1}{c|}{3.21} & 4.65 & 5.50 & \multicolumn{1}{c|}{-} & \multicolumn{1}{c|}{-} & - & 5.93 \\ 
\hline
ATC-NE \cite{garg2022leveraging} & \multicolumn{2}{c|}{3.14} & \multicolumn{1}{c|}{2.99} & 4.21 & 4.72 & \multicolumn{1}{c|}{-} & \multicolumn{1}{c|}{-} & - & 5.00 \\ 
\hline
\rowcolor{mygray}
\multicolumn{1}{c|}{\textbf{\name{} (ours)}} & \multicolumn{1}{c|}{\textbf{0.52}} & \textbf{0.24} & \multicolumn{1}{c|}{\textbf{0.49}} & \textbf{0.84} & \textbf{2.34} & \multicolumn{1}{c|}{\textbf{0.80}} & \multicolumn{1}{c|}{\textbf{0.85}} & \textbf{0.81} & \textbf{2.50} \\ 
\Xhline{0.8pt}
\end{tabular}%
}
\end{table*}

\subsection{Main Results}
In Table \ref{table_1}, we report the mean absolute error (MAE) results of estimating classifier accuracy on unseen test sets. 
From this table, among all data setup, we conclude that our method reduces the accuracy estimation error by about \textbf{47.2\%} on average upon prior SOTA method. 
Further, \name{} shows strong performance regardless of the image domains or the granularities of the classification categories.
Additionally, in Table \ref{table_2} and \ref{table_3}, we report some more detailed results, regarding statistical correlation scores with other intermediate results.

\textbf{Validity of Multi-task Co-training Setup.}
To guarantee a fair-minded assessment of model performance, we must ensure that auxiliary contrastive learning task satisfies the following criteria:
i)-no extra learning complexity for the main task, 
ii)-minimal network changes and
iii)-does not degrade classification accuracy.
To substantiate the soundness of our co-training buildup, we report the ground-truth accuracies of the co-trained classifiers in Table \ref{table_3}. And below, we give an apple-to-apple comparison with classification-only model -- FD ~\cite{deng2020labels}:
SVHN (23.42 vs. 25.46), USPS (88.64 vs. 64.08), Pascal (85.04 vs. 86.13), Caltech (96.08 vs. 93.40), ImageNet (82.66 vs. 88.83). 
Drawing from these results we can know that co-training as a feasible strategy will not degrade the performance of the model to be evaluated. 
This signifies our adherence to the principle of fairly evaluating a model's performance that deserves testing.

\begin{table*}[t]
\centering
\caption{Results of ground-truth classification accuracy, predicted classification accuracy, contrastive learning accuracy and MAE from \name{} (ours) in different data setups.}\label{table_3}
\resizebox{\textwidth}{!}{%
\begin{tabular}{c|cc|cc|c|ccc|c}
\Xhline{0.8pt}
\multirow{2}{*}{Dataset} & \multicolumn{2}{c|}{MNIST} & \multicolumn{2}{c|}{CIFAR-10} & CIFAR-100 & \multicolumn{3}{c|}{COCO} & TinyImageNet \\ 
\cline{2-10} 
 & \multicolumn{1}{c|}{SVHN} & USPS & \multicolumn{1}{c|}{CIFAR-10.1} & CIFAR10-C & CIFAR100-C & \multicolumn{1}{c|}{Caltech} & \multicolumn{1}{c|}{Pascal} & ImageNet & TinyImageNet-C \\ 
  \hline
Ground-truth Cla. Acc. & \multicolumn{1}{c|}{23.42} & 88.64 & \multicolumn{1}{c|}{80.80} & 73.71 & 48.04 & \multicolumn{1}{c|}{96.08} & \multicolumn{1}{c|}{85.04} & 82.66 & 40.41 \\
\hline
Predicted Cla. Acc. & \multicolumn{1}{c|}{23.94} & 88.40 & \multicolumn{1}{c|}{80.31} & 74.55 & 45.70 & \multicolumn{1}{c|}{96.88} & \multicolumn{1}{c|}{85.89} & 83.47 & 42.51 \\
\hline
Con. Acc. & \multicolumn{1}{c|}{14.19} & 42.98 & \multicolumn{1}{c|}{88.47} & 80.60 & 98.74 & \multicolumn{1}{c|}{98.43} & \multicolumn{1}{c|}{91.74} & 90.02 & 82.83 \\
\hline
MAE & \multicolumn{1}{c|}{0.52} & 0.24 & \multicolumn{1}{c|}{0.49} & 0.84 & 2.34 & \multicolumn{1}{c|}{0.80} & \multicolumn{1}{c|}{0.85} & 0.81 & 2.50 \\ 
\Xhline{0.8pt}
\end{tabular}%
}
% \vspace{-0.3cm}
\end{table*}
\footnotetext[2]{Since the pipeline of each algorithms to came up with the accuracy evaluation is vastly different, we simplify compare their time complexity at the algorithm category level:
\textit{confidence-based $\textless$ regression-based \textcolor{magenta}{\textbf{(we are here)}} $\textless$ agreement-based $\textless $ distribution statistics $\textless$ self-training.}}

\begin{figure}[t]
    \centering
    \begin{subfigure}{0.235\textwidth}
        \centering
        \includegraphics[width=\textwidth]{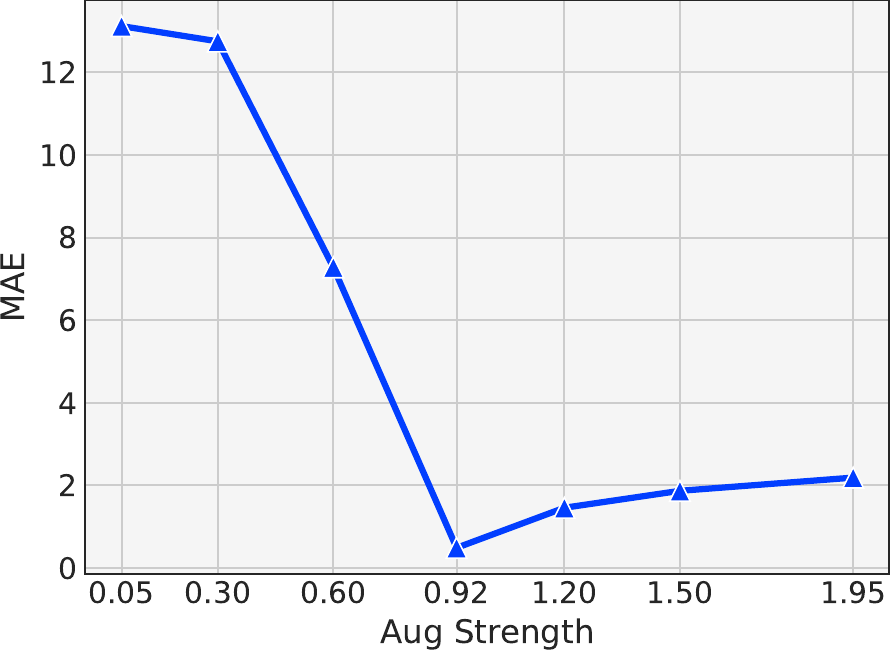}
        % \caption{MAE}
    \end{subfigure}
    \begin{subfigure}{0.235\textwidth}
        \centering
        \includegraphics[width=\textwidth]{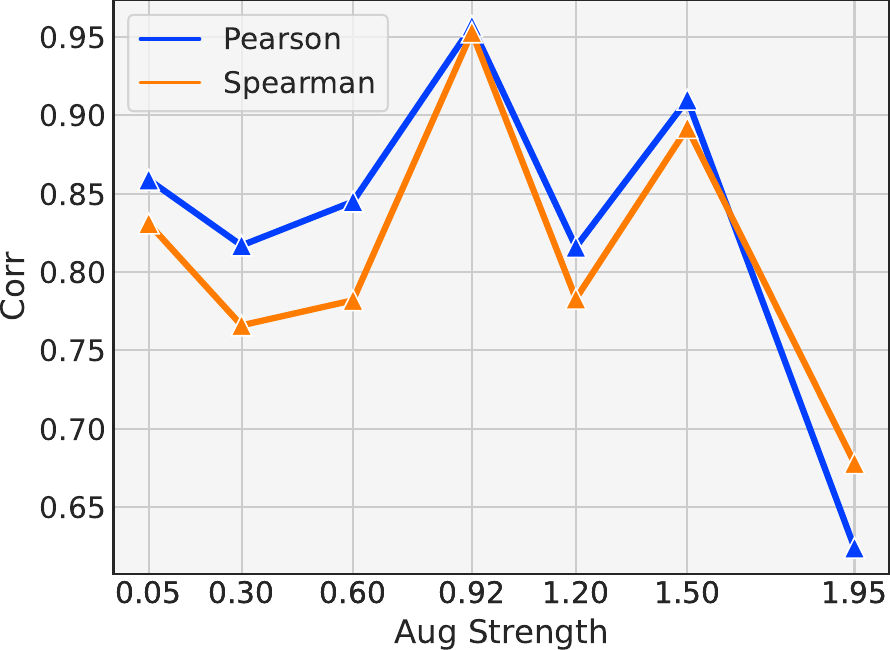}
        % \caption{Corr}
    \end{subfigure}
    \caption{Pearson's correlation ($r$), Spearman's rank correlation ($\rho$) and MAE with different augmentation strength on the RandomResizedCrop operation under CIFAR-10 setup.}
    \label{RandomResizedCrop}
    \vspace{-0.5cm}
\end{figure}

\subsection{Ablation Studies}
\subsubsection{Contrastive Learning Parameters}
As we adopt SimCLR in our framework, we want to know if our overall performance is consistent to its basic settings (i.e. the linear correlation coefficient achieves its maximum and the estimation error achieves its minimum under the best training parameters of SimCLR). 
According to \cite{wang2022chaos}, among the data augmentations adopted in SimCLR, RandomResizedCrop is the most important augmentation, and ColorJitter is the second. 
So we study the impact of these two kinds of augmentations on CIFAR-10 in our work.

For \textbf{RandomResizedCrop}, to quantify its influence, we use the augmentation strength defined in \cite{wang2022chaos}. 
For a RandomResizedCrop operator with scale range $[a,b]$, its aug-strength can be defined as $r=(1-b)+(1-a)$. 
In Figure \ref{RandomResizedCrop}, we show that under different augmentation strengths, the accuracy estimation error achieves its minimum at the default strength value 0.92.

\begin{figure}[t]
    \centering
    \begin{subfigure}{0.47\textwidth}
        \centering
        \includegraphics[width=\textwidth]{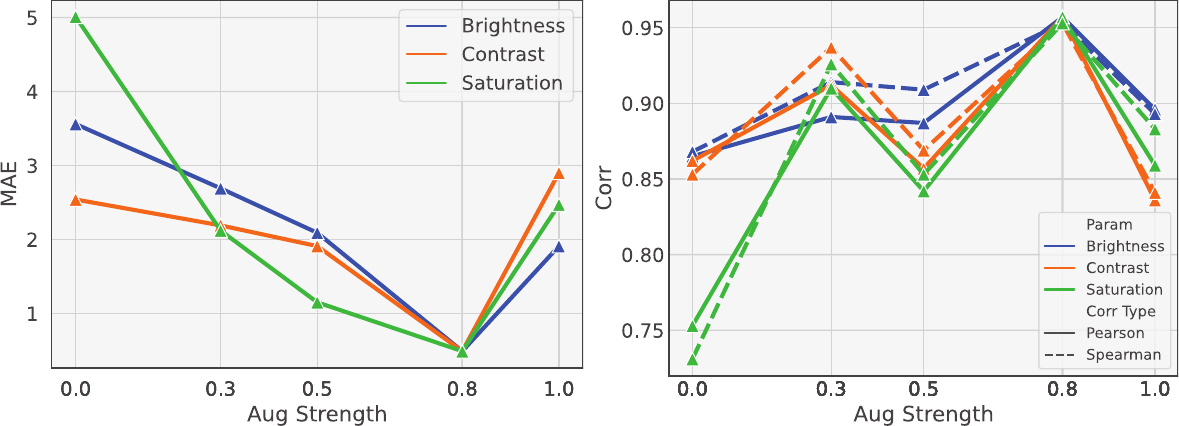}
        \caption{Brightness, Contrast, Saturation}
        \label{BCS_mae}
    \end{subfigure}
    \begin{subfigure}{0.47\textwidth}
        \centering
        \includegraphics[width=\textwidth]{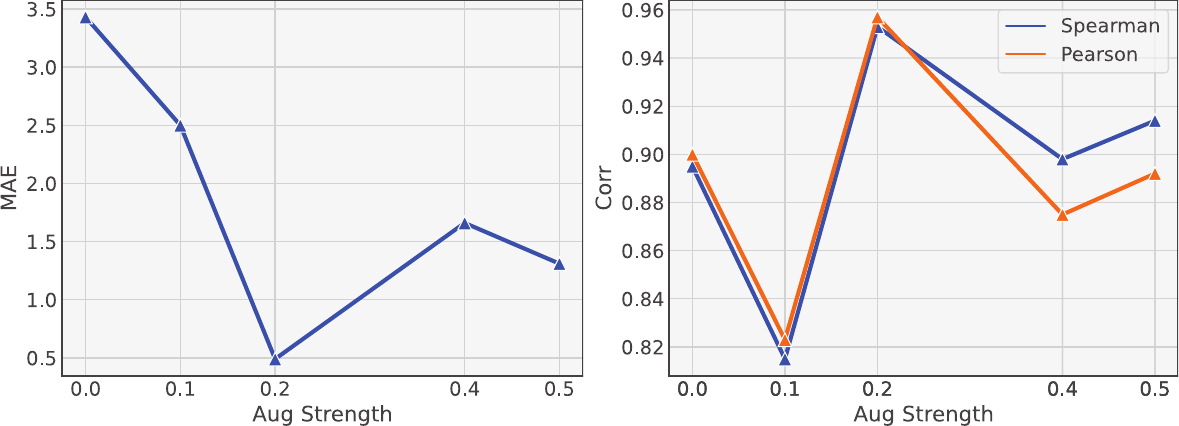}
        \caption{Hue}
        \label{BCS_corr}
    \end{subfigure}
    \caption{Pearson's correlation ($r$), Spearman's rank correlation ($\rho$) and MAE with different augmentation strength on color jittering operations under CIFAR-10 setup.}
    \label{ColorJitter}
\end{figure}

For \textbf{ColorJitter}, we study its parameters: brightness, contrast, saturation and hue, where the augmentation strength is corresponding to the parameter value. Note that all other augmentations in SimCLR are kept default. In Figure \ref{ColorJitter}, for each of these parameters, we plot the changes of the accuracy estimation error under the CIFAR-10 setup. Here we have similar observation that the default parameter values ($Brightness=Contrast=Saturation=0.8, Hue=0.2$) in SimCLR yield best performance.

\begin{figure}[t]
    \centering
    \begin{subfigure}{0.235\textwidth}
        \centering
        \includegraphics[width=\textwidth]{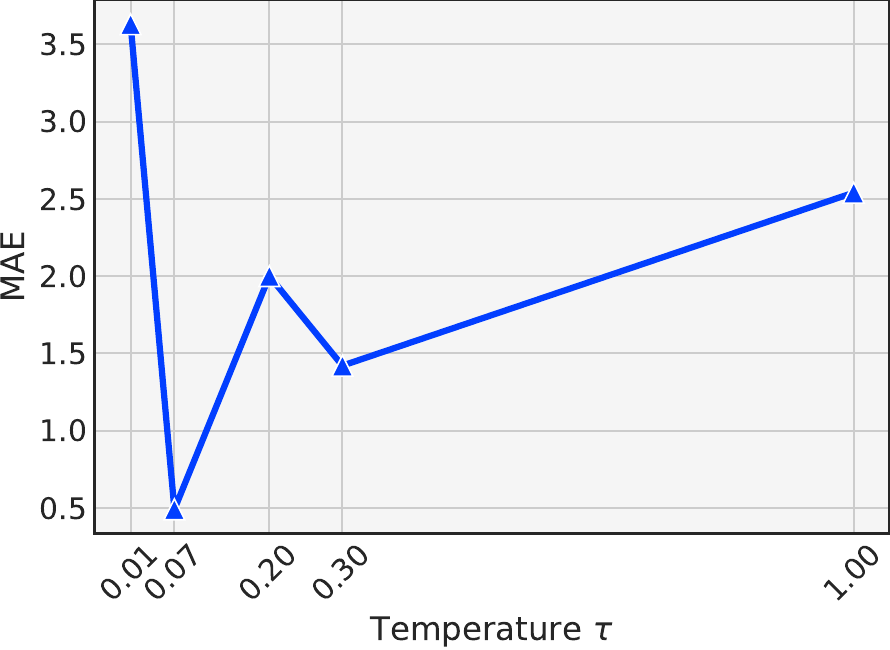}
        % \caption{MAE}
    \end{subfigure}
    \begin{subfigure}{0.235\textwidth}
        \centering
        \includegraphics[width=\textwidth]{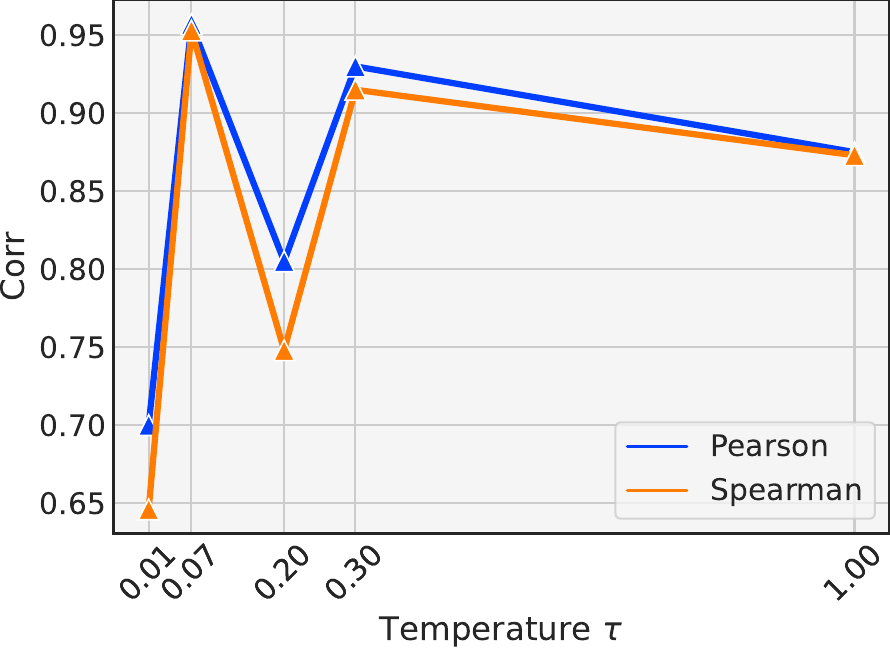}
        % \caption{Corr}
    \end{subfigure}
    \caption{Pearson's correlation ($r$), Spearman's rank correlation ($\rho$) and MAE with different temperature scaling parameters under CIFAR-10 setup.}
    \label{temperature}
\end{figure}

Also, \textbf{temperature scaling} is an important factor during the training process of SimCLR. We study the temperature parameter $\tau$ on CIFAR-10. As Figure \ref{temperature} shows, when using default temperature value $\tau=0.07$, we can obtain best performance for both MAE and correlation coefficient.

Finally, we investigate how to assign appropriate \textbf{task weights} in the proposed multi-task training. 
Here we fix the classification weight to 1.0, and change different CL task weights: 1.0, 0.1, 0.01, 0.001. In Table \ref{weight}, we report the linear correlation score and estimation error. 
The insight is that our framework is more likely to perform well when assigning small weight ($<0.01$) for the CL task.
\begin{table}[t]
\centering
\caption{Pearson's correlation ($r$), Spearman's rank correlation ($\rho$) and MAE with different contrastive learning task weight $\lambda$ under CIFAR-10 setup.}
\label{weight}
\resizebox{0.4\textwidth}{!}{%
\begin{tabular}{c|c|c|c|c}
\Xhline{0.8pt}
CL task weight & 1.0 & 0.1 & 0.01 & 0.001 \\ \hline
$r$ & 0.953 & 0.946 & 0.872 & 0.953 \\ \hline
$\rho$ & 0.957 & 0.933 & 0.860 & 0.957 \\ \hline
MAE & 0.71 & 2.87 & 0.91 & 0.49 \\ 
\Xhline{0.8pt}
\end{tabular}%
}
% \vspace{-0.5cm}
\end{table}

In summary, these empirical results demonstrate that when adopting CL frameworks, keeping default optimal settings is most likely to build strong linear correlation between the CL accuracy and classification accuracy, as well as obtain lowest accuracy estimation error on the final unlabeled test set.
% \vspace{-0.3cm}

\subsubsection{Different Training Settings}
\setlength{\parindent}{1em}\textbf{Different contrastive learning augmentation groups.}
In this paper, we adopt SimCLR, to study if other frameworks can fit in well, we change SimCLR to MoCo-v1 \cite{he2019moco}, MoCo-v2 \cite{chen2020mocov2}, and BYOL \cite{grill2020bootstrap}. 
From Table \ref{aug_groups}, we can observe that on CIFAR-10, the linear correlations are all strong across different CL frameworks ($r>0.86$).
\begin{table}[t]
\centering
\caption{Pearson’s correlation ($r$), Spearman’s rank correlation ($\rho$) and MAE with different CL data augmentation under CIFAR-10 setup.}
\label{aug_groups}
\renewcommand{\arraystretch}{1.0} %adjust line spacing
\resizebox{0.47\textwidth}{!}
{{
\begin{tabular}{c|c|c|c|c}
\Xhline{0.8pt}
CL Data Aug & SimCLR & MoCo-v1 & MoCo-v2 & BYOL \\ 
 \hline
$r$ & 0.957 & 0.868 & 0.893 & 0.922 \\ 
\hline
$\rho$ & 0.953 & 0.864 & 0.897 & 0.902 \\ 
\hline
MAE & 0.49 & 1.20 & 0.88 & 0.59 \\ 
\Xhline{0.8pt}
\end{tabular}%
}}
\end{table}

\begin{figure}[t]
    \centering
    \begin{subfigure}{0.47\textwidth}
        \centering
        \includegraphics[width=\textwidth]{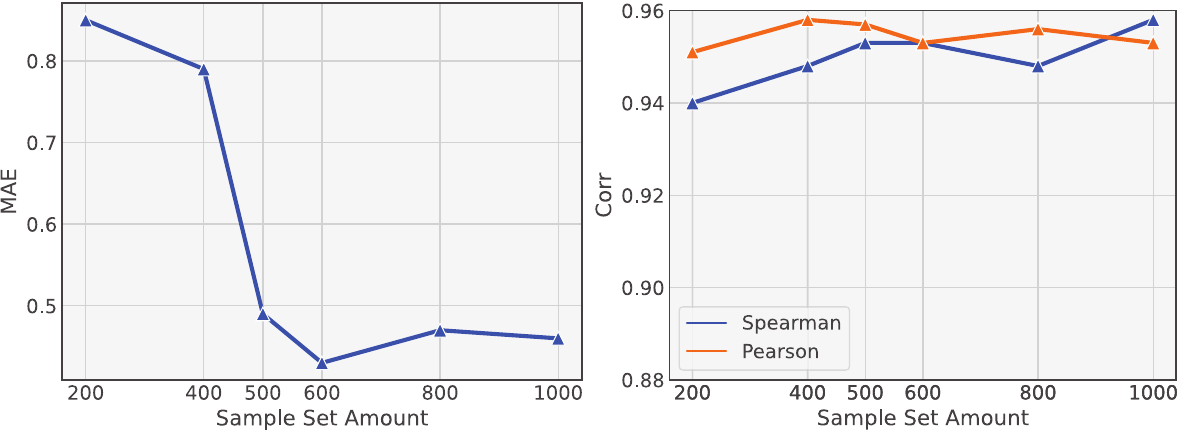}
        \caption{Sample set amount}
    \end{subfigure}
    \begin{subfigure}{0.47\textwidth}
        \centering
        \includegraphics[width=\textwidth]{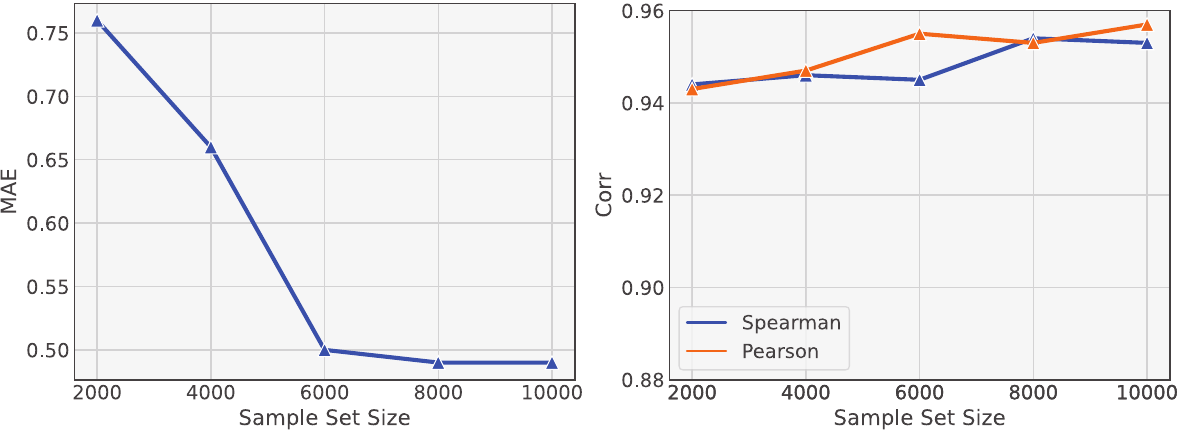}
        \caption{Sample set size}
    \end{subfigure}
    \caption{Pearson's correlation ($r$), Spearman's rank correlation ($\rho$) and MAE with different meta-set size and sample-set size under CIFAR-10 setup.}
    \label{set_size}
    % \vspace{-0.5cm}
\end{figure}

\textbf{Different amounts and sizes of synthetic datasets.}
In this paper, we synthesize many sample sets to build a regression model for accuracy estimation, so we study the influence of the amount of synthetic test sets and the size of each test set. 
Here, we refer to them as \emph{sample set amount} and \emph{sample set size} respectively. 
As Figure \ref{set_size} shows, the linear correlation is quite robust and the estimation error is also robust when there are enough test sets.

\textbf{Different backbones.}
In experimental setup, we use DenseNet-40-12 for CIFAR-10 setup in default. Here we change it to other model structures (ResNet-18, ResNet-34, VGG-11, VGG-19 \cite{simonyan2014very}) to study if the choice of backbone significantly influences the performance. As Table \ref{backbones} shows, our framework is robust against different backbones (strong linear correlation and precise accuracy estimation).
\begin{table}[t]
\centering
\caption{Pearson's correlation ($r$), Spearman's rank correlation ($\rho$) and MAE with different CNN backbones under CIFAR-10 setup.}
\label{backbones}
\renewcommand{\arraystretch}{1.1} %adjust line spacing
\setlength{\tabcolsep}{0.8pt}
\resizebox{0.48\textwidth}{!}{%
\begin{tabular}{c|c|c|c|c|c}
\Xhline{0.8pt}
Backbone & ResNet-18 & ResNet-34 & VGG-11 & VGG-19 & DenseNet-40-12 \\ \hline
$r$ & 0.853 & 0.854 & 0.816 & 0.863 & 0.957 \\ \hline
$\rho$ & 0.858 & 0.853 & 0.787 & 0.849 & 0.953 \\ \hline
MAE & 0.92 & 0.63 & 1.17 & 0.77 & 0.49 \\
\Xhline{0.8pt}
\end{tabular}%
}
\vspace{-0.5cm}
\end{table}

\textbf{Different random seeds.}
To check if the experimental results are robust to the initial random state, we choose different random seeds for training (use 0 as default seed). As Table \ref{seed} shows, the performance of our framework is robust to randomness.
\begin{table}[t]
\centering
\caption{Pearson's correlation ($r$), Spearman's rank correlation ($\rho$) and MAE with different random seeds under CIFAR-10 setup.}
\label{seed}
\resizebox{0.3\textwidth}{!}{%
\begin{tabular}{c|c|c|c}
\Xhline{0.8pt}
Random Seed & 0 & 21 & 42 \\ \hline
$r$ & 0.957 & 0.892 & 0.869 \\ \hline
$\rho$ & 0.953 & 0.894 & 0.845 \\ \hline
MAE & 0.49 & 0.65 & 0.51 \\
\Xhline{0.8pt}
\end{tabular}%
}
\end{table}

\begin{figure}[t]
    \centering
    \includegraphics[width=0.45\textwidth]{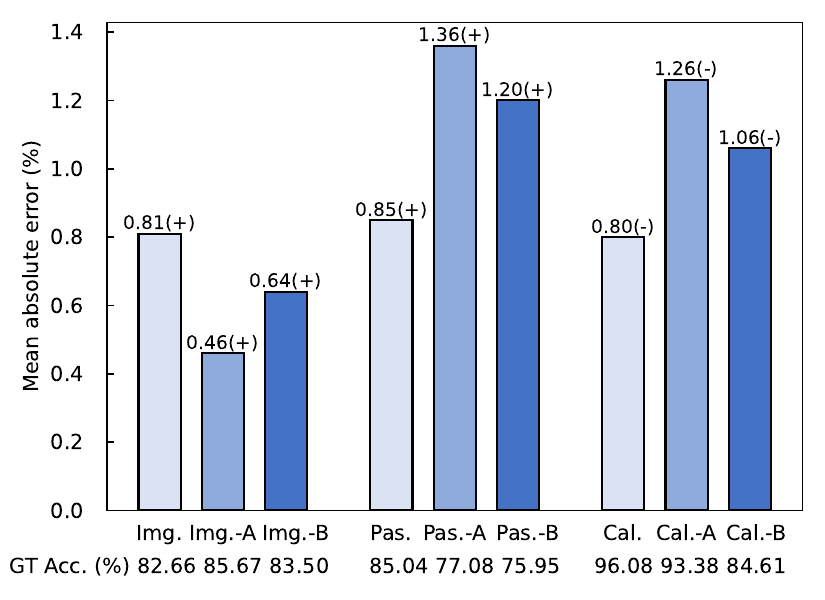}
    \caption{The MAE of linear regressor on transformed test sets (ImageNet, Pascal, and Caltech).
The transformed datasets are denoted by “-A” and “-B” with new transformations such Cutout~\cite{devries2017improved}, Shear, Equalize and ColorTemperature~\cite{cubuk2019autoaugment}. 
(-) / (+) denotes the estimated accuracy is
lower and higher than the ground-truth accuracy, respectively.}
    \label{transformed_test}
    \vspace{-0.5cm}
\end{figure}

\begin{figure*}[t]
    \centering
    \begin{subfigure}{0.33\textwidth}
        \centering
        \includegraphics[width=\textwidth]{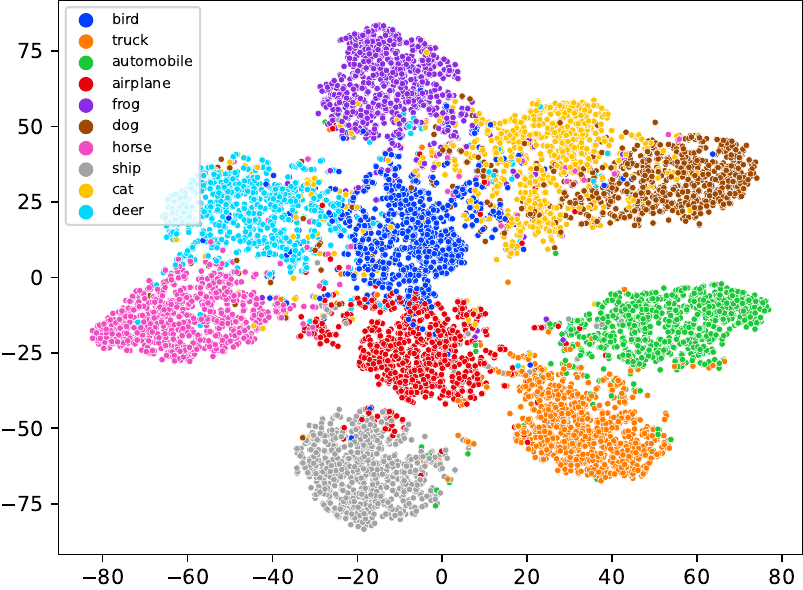}
        \caption{Only Classification}
    \end{subfigure}
    \begin{subfigure}{0.33\textwidth}
        \centering
        \includegraphics[width=\textwidth]{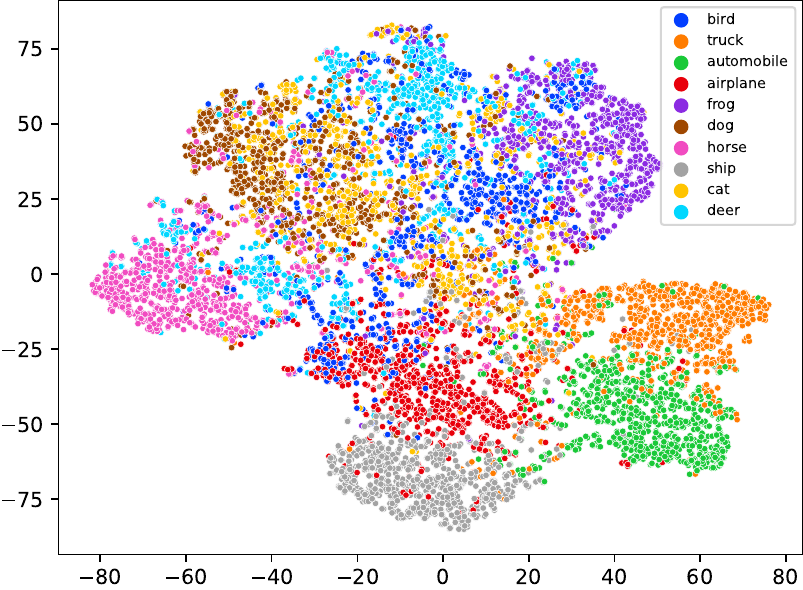}
        \caption{Pre-train + Fine-tune}
    \end{subfigure}
    \begin{subfigure}{0.33\textwidth}
    \centering
    \includegraphics[width=\textwidth]{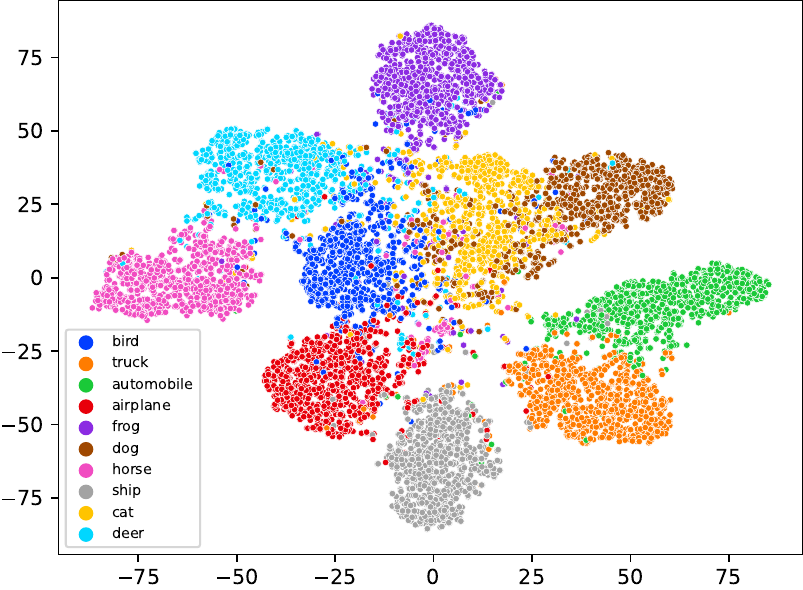}
    \caption{Multi-task}
    \end{subfigure}
    \caption{T-SNE visualization of the classification head features on CIFAR-10 validation set. Different colors correspond to different classes.}
    \label{feature_cluster}
    \vspace{-0.3cm}
\end{figure*}

\section{Method Interpretability}
\textbf{Regressor Robustness on transformed test sets.}
To further check the regressor robustness in the unobserved testing environment.
We rigorously shift three natural datasets (ImageNet, Pascal and Caltech) with new transformations that do not overlap with the various transformations constructed when fitting regression curves.
Specifically, we use Cutout~\cite{devries2017improved}, Shear, Equalize and ColorTemperature~\cite{cubuk2019autoaugment} to generate ImageNet-A$/$B, Pascal-A$/$B, Caltach-A$/$B.
We note the following observations from Figure \ref{transformed_test}. 
First, the classifier accuracies will fluctuate after these test sets are shifted. Second, even in these unseen test sets cases undergoing new transformations, our method consistently achieves superior results. 
We conjecture that this phenomenon stems from contrastive learning performing well across a broad spectrum of underlying distributions.

\textbf{Underlying relationship between contrastive accuracy and classification accuracy.}
In common practice, contrastive learning is often used as a pre-training task, which has been proved effective for learning visual representations for downstream classification task. 
In this work, we adopt SimCLR in multi-task training. Here we compare the two ways in Table \ref{manner}. 
We can observe that both of them can yield strong linear correlation and precise accuracy estimation, while the multi-task way is better. 
This finding further reveals the underlying relationship between contrastive accuracy and classification accuracy, regardless of the training way.

\begin{table}[t]
\centering
\caption{Comparison between the pre-train+fine-tune (Pre+Fine) and multi-task training on CIFAR-10.}
\label{manner}
\renewcommand{\arraystretch}{1.2} %adjust line spacing
\resizebox{0.48\textwidth}{!}{%
\begin{tabular}{c|c|c|c|c|c|c}
\Xhline{0.8pt}
Paradigm & $r$ & $\rho$ & $Acc_{con}$ & $Acc_{cla}$ & $\hat{Acc_{cls}}$ & MAE \\ \hline
\begin{tabular}[c]{@{}c@{}}Pre+Fine\end{tabular} & 0.900 & 0.928 & 94.87 & 61.20 & 60.20 & 1.00 \\ \hline
Multi-task & 0.957 & 0.953 & 88.47 & 80.80 & 80.41 & 0.49 \\ 
\Xhline{0.8pt}
\end{tabular}%
}
\end{table}

\begin{figure}[t]
    \centering
    \begin{subfigure}{0.235\textwidth}
        \centering
        \includegraphics[width=\textwidth]{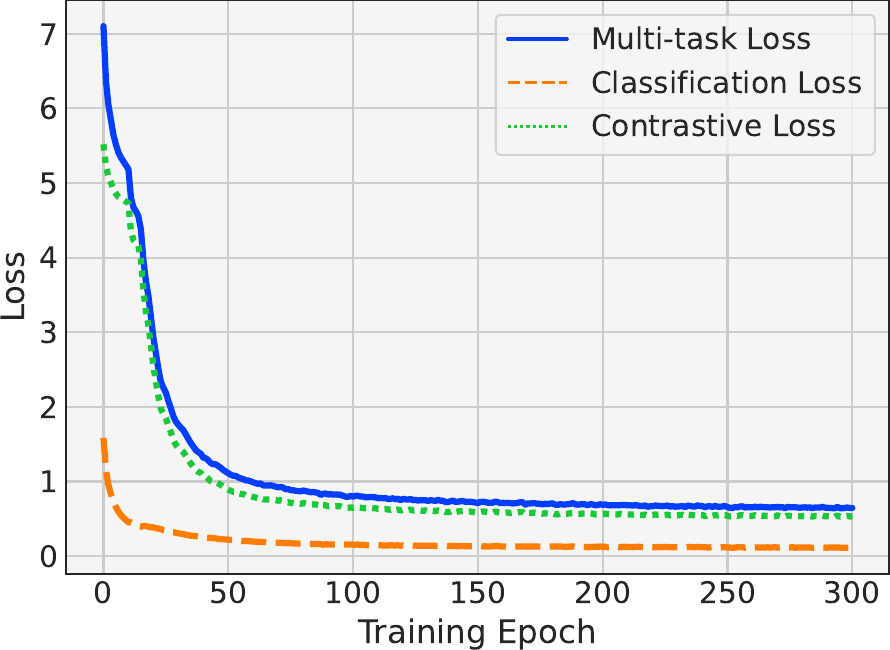}
        \caption{Loss}
    \end{subfigure}
    \begin{subfigure}{0.235\textwidth}
        \centering
        \includegraphics[width=\textwidth]{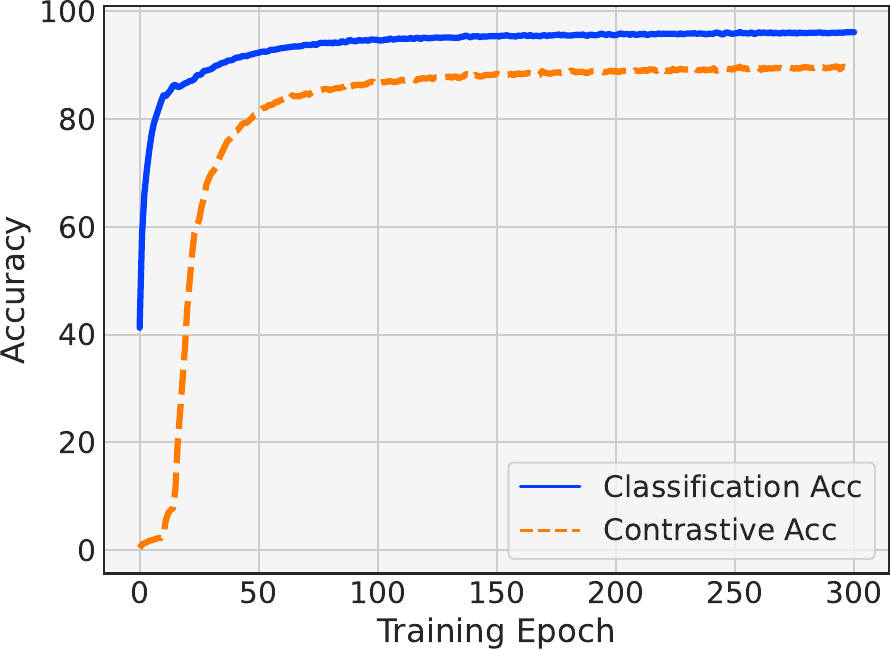}
        \caption{Accuracy}
    \end{subfigure}
    \caption{Learning curve plot of accuracy and loss in training phase.}
    \label{learning_curve}
    \vspace{-0.4cm}
\end{figure}

\textbf{Multi-task training learns better features for AutoEval.}
Intuitively, strong linear correlation bewteen contrastive accuracy and classification accuracy can be built because they both learn class-wise visual representations. 
What kind of image features will be learned in multi-task training?
In Figure \ref{feature_cluster} and \ref{learning_curve}, we plot some intermediate results during the training process.
Notably, compared to pre-train + fine-tune, multi-task yields better feature clusters, which well corresponds to the results in Table \ref{manner}.
This further justifies the usage of contrastive learning in a multi-task paradigm is indeed feasible.

\section{Conclusion}
In this paper, we propose a novel framework \name{} for estimating classifier accuracy on invisible test sets without ground-truth labels. 
We find that there is a strong linear correlation between contrastive accuracy and classification accuracy and give the theoretical analysis to support this discovery.
Thus, our work indicates that it is feasible to estimate classifier accuracy by self-supervised contrastive accuracy using linear regression.
Extensive experimental results show that our method achieves new SOTA results for AutoEval by outperforming previous works.
In future, we will explore the feasibility of other self-supervised learning tasks in AutoEval, and extend \name{} to other fields such as natural language processing and graph learning tasks. 

\section*{Acknowledgement}
This work is majorly supported by the Fundamental Research Funds for the Central Universities (No. 226-2022-00028), and in part by the NSFC Grants (No. 62206247). 
RP,  HW and JZ also thank the sponsorship by the CAAI-Huawei Open Fund.

\section*{Limitation}
Our method is grounded on an assumption that approximates the image distribution of unknown data environments via image transformations applied to the proposed synthetic sample sets. 
However, this might encounter certain intricate real-world cases where not be able to work.
For example, there could be non-negligible samples in testing sets whose classes have never appeared in the training label space. 
Consequently, even though we will still predict an estimated accuracy, it is essentially unavailable.
Despite the seemingly extreme situation, this issue could be well alleviated by employing out-of-distribution detection techniques~\cite{hsu2020generalized,liu2020energy,lee2018simple,liang2017enhancing} to help detect and reject such samples.
Furthermore, our method is not a plug-and-play solution for model evaluation due to the co-training strategy.
Nonetheless, our method serves as a general technique with the potential for extension across various fields, supported by the widespread deployment of contrastive learning work.

{\small
\bibliographystyle{ieee_fullname}
\bibliography{main}
\nocite{*}
}

\clearpage
\appendix

\section{Details of Dataset Setup}
In our work, we consider both natural and synthetic distribution shifts in empirical evaluation. The details of the dataset settings are shown in Table \ref{table9}.

\begin{table*}[t]
\centering
\caption{Details of the datasets considered in our paper, where the validation set that has not undergone data transformation is used as the seed set.}\label{table9}
\resizebox{\textwidth}{!}{%
\begin{tabular}{c|c|c}
\hline
Train set (source) & Valid set (source) & Unseen test set (target) \\ \hline
MNIST (train) & MNIST (valid) & USPS, SVHN \\ \hline
CIFAR-10 (train) & CIFAR-10 (valid) & CIFAR-10.1, 95 CIFAR10-C (Fog and Motion blur, etc.) \\ \hline
CIFAR-100 (train) & CIFAR-100 (valid) & 95 CIFAR-100-C (Fog and Motion blur, etc.) \\ \hline
COCO 2014 (train) & COCO 2014 (valid) & Caltech256 (test), PASCAL VOC 2007 (test), ImageNet (test) \\ \hline
TinyImageNet (train) & TinyImageNet (valid) & 95 TinyImageNet-C  (Fog and Motion blur, etc.) \\ \hline
\end{tabular}%
}
\end{table*}

\section{Baseline Methods}
\textbf{Prediction Score (Pred)}~\cite{hendrycks2016baseline}
It is defined as the maximum softmax output of the model. If the prediction score of a sample is greater than a given threshold $\tau$, it is regarded correct.

\textbf{Entropy Score (Entropy)}~\cite{kendall2017uncertainties}
It is defined as the normalized entropy of softmax outputs (normalized by $\log K$, where $K$ is the number of classes. If the entropy score of a sample is less than a given threshold $\tau$, it is regarded correct.

\textbf{Proxy Risk}~\cite{chuang2020estimating}
This work propose a set of domain-invariant predictors as a proxy for the unknown, true target labels. 
They train the check model and fine-tune it to maximize the disagreement using a separate target training dataset sampled from the distribution of the test dataset. If the check model gives a different prediction, it is regarded as an error prediction of the given model.

\textbf{Ensemble Average Confidence (Ens. AC)}~\cite{lakshminarayanan2017simple}
This method involves training multiple neural networks with different initializations and architectures, and then combining their predictions to obtain a probabilistic estimate of the target variable. 
The model’s accuracy is estimated by the average confidence calculated from the model ensemble.

\textbf{Ensemble Method $\mathcal{T}_{RI}$ (Ens. RI)}~\cite{chen2021detecting}
This method uses the same training algorithm as for the given model to train a model ensemble from different random initialization (RI). 
If the model gives a different prediction, it is regarded as an error prediction of the given model.

\textbf{Ensemble Method $\mathcal{T}_{RM}$ (Ens. RM)}~\cite{chen2021detecting}
This method is also based on model ensemble similar to the Ens RI mentioned above, but designed with the representation matching (RM) technique for domain adaptation, which can potentially improve the accuracy of the ensemble on some test inputs related to the training data.

\textbf{Frechet}~\cite{deng2020labels}
This method first synthesizes many test sets. 
And then, it computes the frechet Distance (FD) between the training set and each of the test set. 
Using the (FD, acc) value pairs, it can build a regression model to estimate the model’s accuracy on an unlabeled test set.

\textbf{Frechet + $\mu$ + $\sigma$}~\cite{deng2020labels}
Similar to frechet mentioned above, but adds the mean and variance values to the Frechet Distance and train a neural network regression to estimate the testing performance of unlabeled set.

\textbf{Semi-Structured Dataset Representations (SSDR)}~\cite{sun2021label}
Similar to the frechet Distance based method mentioned above, it uses a semi-structured dataset feature to regress the model’s accuracy.

\textbf{Average Confidence (AC)}~\cite{hendrycks2016baseline}
It uses the average of the model’s confidence (maximum softmax output) as the model’s accuracy on the test set.

\textbf{Difference of Confidence (DoC)}~\cite{guillory2021predicting}
This method uses the difference of confidences on source and target distribution to regress the model’s accuracy.

\textbf{Importance-re-weighting (IM)}~\cite{chen2021mandoline}
This method estimates the model’s accuracy on target data by the importance ratios, by using this, the trained model’s accuracy can be converted to the accuracy on the unlabeled target test set.

\textbf{Generalization Disagreement Equality (GDE)}~\cite{jiang2021assessing}
This method first trains two models, which are trained on the same training set but with different initialization or different data ordering. 
Then, it simultaneously uses the two models to predict, if their predictions disagree, it’s considered as an error prediction.

\textbf{Average Thresholded Confidence (ATC-MC and ATC-NE)}~\cite{garg2022leveraging}
This method proposes average thresholded confidence, which learns a threshold on a score of model confidence on validation source data and predicts target domain accuracy as the fraction of unlabeled target points that receive a score above that threshold. 
ATC-MC uses the mean confidence as the score, while ATC-NE uses the negative entropy.

\section{Training Details}
Overall, we train classification along with SimCLR in a multi-task way. 
Here are some detailed training parameters under different setups.
\textbf{MNIST}
We train LeNet-5 on MNIST. 
We choose the Adam optimizer, with learning rate $3e^{-4}$, and train 700 epochs with batch size 2048. 

\textbf{CIFAR-10 and CIFAR-100}
For CIFAR-10 and CIFAR-100, the model architecture we use is DenseNet-40-12 (40 layers with growth rate 12). 
In the training phase, we use the SGD optimizer with momentum 0.9, and train 300 epochs with batch size 128. 
The initial learning rate is 0.1, and decay by multiplying 0.1 at epoch 150 and epoch 225.

\textbf{COCO}
We use pre-trained ResNet-50, and train 50 epochs with batch size 128. 
For training, we use the SGD optimizer with momentum 0.9. 
The initial learning rate is $1e^{-3}$, decayed by multiplying 0.1 at epoch 20 and epoch 30.

\textbf{TinyImageNet}
We use pre-trained ResNet-50, and train 50 epochs with batch size 128.
For training, we use the SGD optimizer with momentum 0.9.
The initial learning rate is $5e^{-3}$, decayed by multiplying 0.1 at epoch 20 and epoch 30.

\section{Sample Visualization of Synthetic Sets}
\begin{figure}[t]
    \centering
    \begin{subfigure}{0.15\textwidth}
        \centering
        \includegraphics[width=\textwidth]{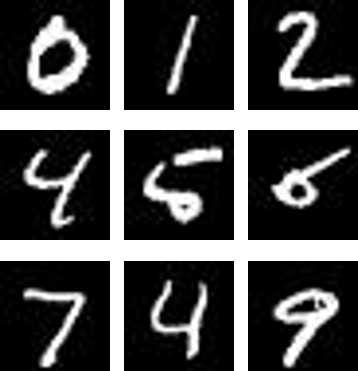}
        \caption{original set}
    \end{subfigure}
    \begin{subfigure}{0.15\textwidth}
        \centering
        \includegraphics[width=\textwidth]{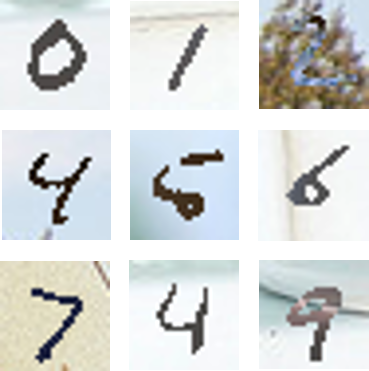}
        \caption{sample set 1}
    \end{subfigure}
    \begin{subfigure}{0.15\textwidth}
        \centering
        \includegraphics[width=\textwidth]{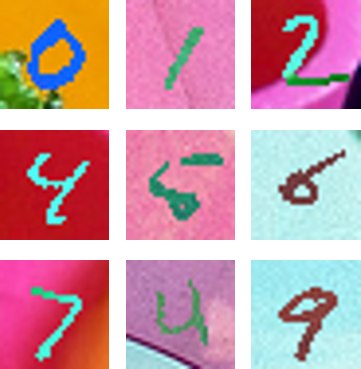}
        \caption{sample set 2}
    \end{subfigure}
    \caption{MNIST sample}
    \label{MNIST sample}
\end{figure}

\begin{figure}[t]
    \centering
    \begin{subfigure}{0.15\textwidth}
        \centering
        \includegraphics[width=\textwidth]{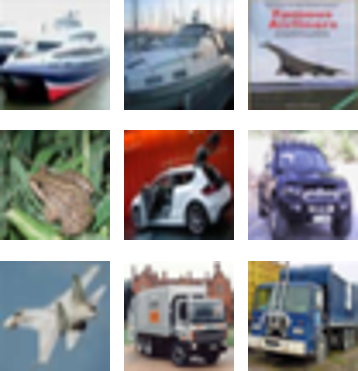}
        \caption{original set}
    \end{subfigure}
    \begin{subfigure}{0.15\textwidth}
        \centering
        \includegraphics[width=\textwidth]{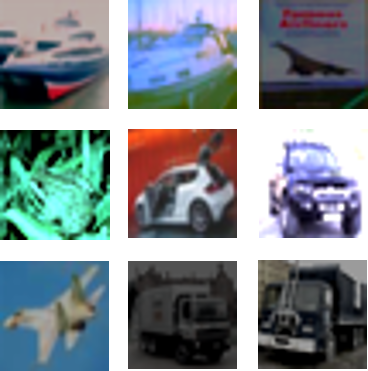}
        \caption{sample set 1}
    \end{subfigure}
    \begin{subfigure}{0.15\textwidth}
        \centering
        \includegraphics[width=\textwidth]{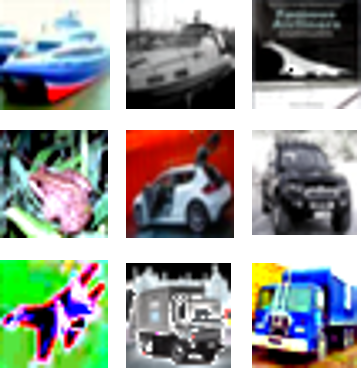}
        \caption{sample set 2}
    \end{subfigure}
    \caption{CIFAR-10 sample}
    \label{CIFAR-10 sample}
    % \vspace{-0.25cm}
\end{figure}

In section \ref{section3_4}, we describe how we synthesize sample sets by applying various transformations on the original seed set. 
Here we provide some visualizations for the generated sample sets (see Figure \ref{MNIST sample}, \ref{CIFAR-10 sample}).

\section{Additional Theoretical Discussion}
Recalling to Theorem \ref{Guarantees for the Optimal Encoder}, we provide some more detailed discussions of this theorem at here, including its basic assumptions and a extended theorems under weaker conditions \cite{wang2022chaos}.

\begin{assumption} \label{Label Consistency}
$\forall x,x^+ \sim p(x,x^+)$, the labels are deterministic(one-hot) and consistent: $p(y|x)=p(y|x^+)$. 
\end{assumption}

Under the premise of satisfying the natural and minimum assumption --- label consistency, we can extend Theorem \ref{Guarantees for the Optimal Encoder} to the situation of any model:

\begin{theorem} \label{Guarantees for General Encoders}
For any model $f \in \mathcal{F}$, its downstream classification risk $\mathcal{L}^\mu_{CE}(f)$ can be bounded by the contrastive learning risk $\mathcal{L}_{NCE}(f)$
\begin{equation}
\begin{aligned}
    &\mathcal{L}_{NCE}(f)-\sqrt{Var(f(x)|y})-\frac{1}{2} \sum_{j=1}^m\sqrt{Var(f_j(x)|y)} \\
    &-\mathcal{O}(M^{-1/2})\leq \mathcal{L}_{CE}^{\mu}(f)+\log(M/K)\leq \\
    &\mathcal{L}_{NCE}(f)+\sqrt{Var(f(x)|y)}+\mathcal{O}(M^{-1/2})
\end{aligned}
\end{equation}
where $\mathcal{L}_{CE}^{\mu}(f)=\mathbb{E}_{p(x,y)}[-\log\frac{\exp({f(x)^T\mu_y})}{\sum_{i=1}^K\exp({f(x)^T\mu_i})}]$, $\log(M/K)$ is a constant, $\mathcal{O}(M^{-1/2})$ denotes the order of the approximation error by using $M$ negative samples, $f_j(x)$ denotes the $j$-th coordinate of $f(x)$, and $Var(f(x)|y)=\mathbb{E}_{p(y)}[\mathbb{E}_{p(x|y)}||f(x)-\mathbb{E}_{p(x|y)}f(x)||^2]$ denotes the conditional variance. 
\end{theorem}

\begin{figure}[t]
    \centering
    \includegraphics[width=0.45\textwidth]{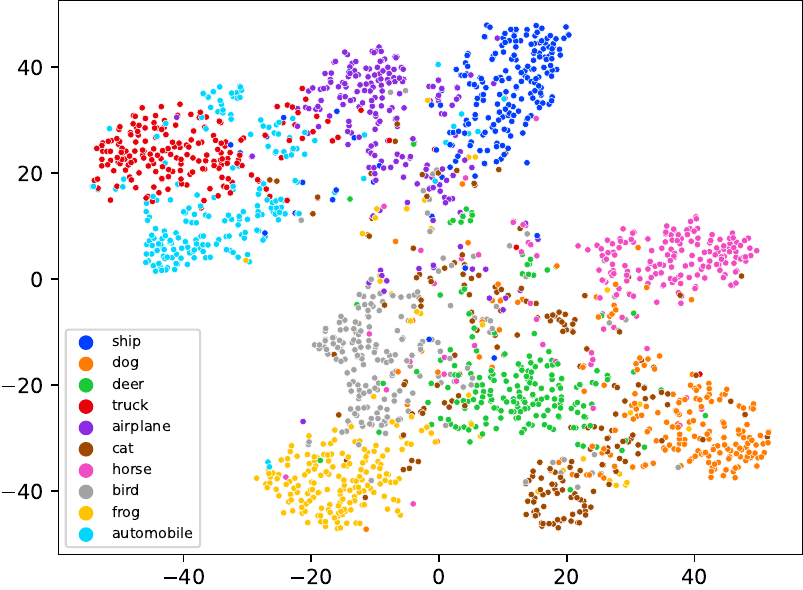}
    \caption{T-SNE clustering visualization of the classification features on CIFAR-10.1 test set. Different colors correspond to different classes.}
    \label{tsne_by_class_test}
    % \vspace{-0.5cm}
\end{figure}

\section{Tightness Analysis of Bounded Theorem}
\begin{figure*}[t]
    \centering
    \begin{subfigure}{0.35\textwidth}
        \centering
        \includegraphics[width=\textwidth]{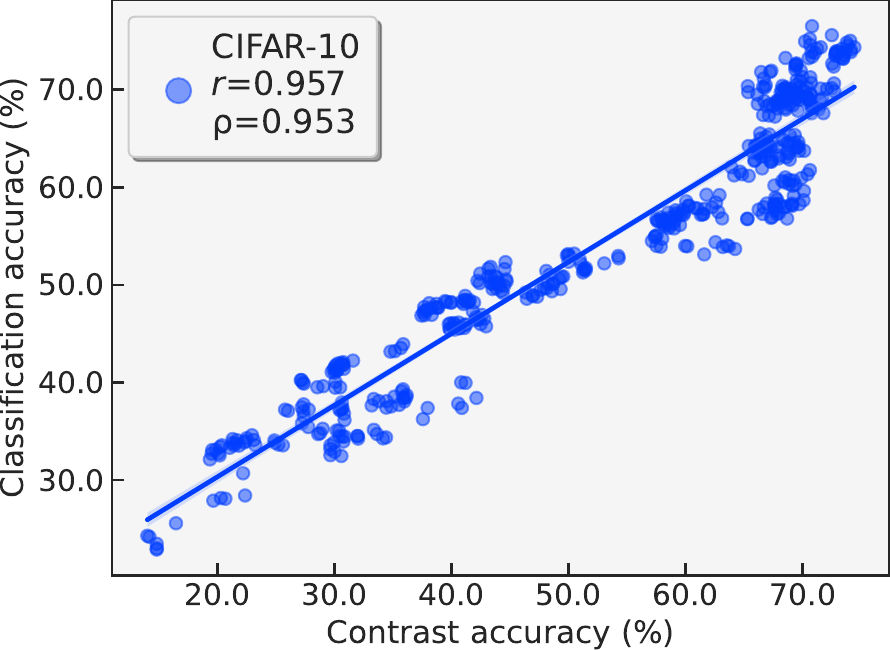}
        \caption{$\lambda=1$}
    \end{subfigure}
    \begin{subfigure}{0.35\textwidth}
        \centering
        \includegraphics[width=\textwidth]{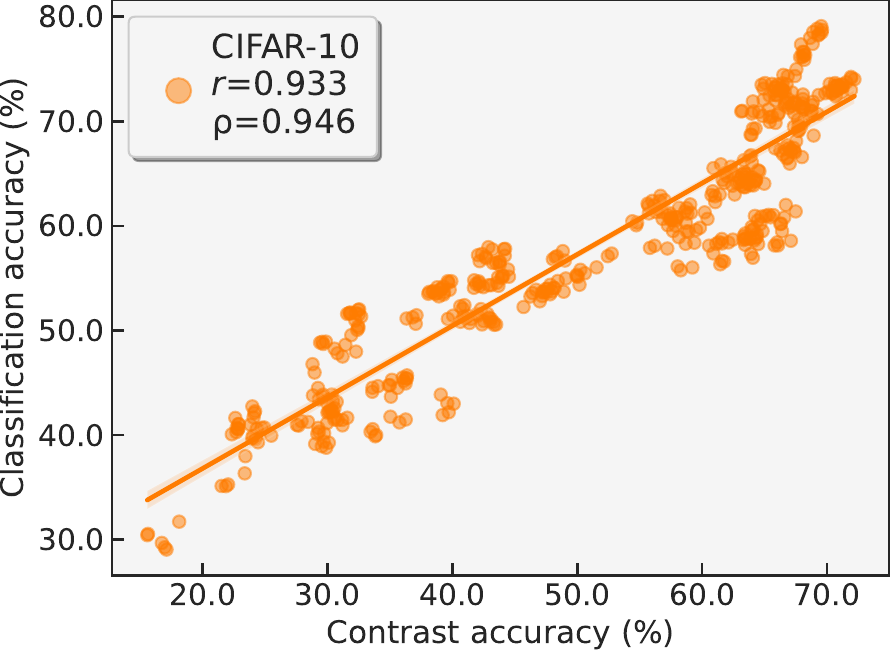}
        \caption{$\lambda=0.1$}
    \end{subfigure}
    \begin{subfigure}{0.35\textwidth}
        \centering
        \includegraphics[width=\textwidth]{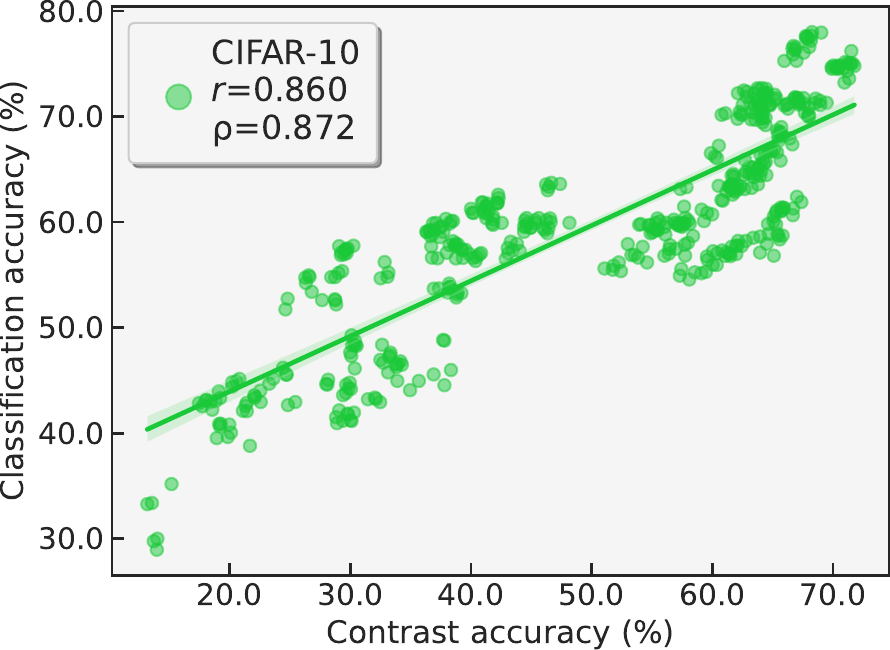}
        \caption{$\lambda=0.01$}
    \end{subfigure}
    \begin{subfigure}{0.35\textwidth}
        \centering
        \includegraphics[width=\textwidth]{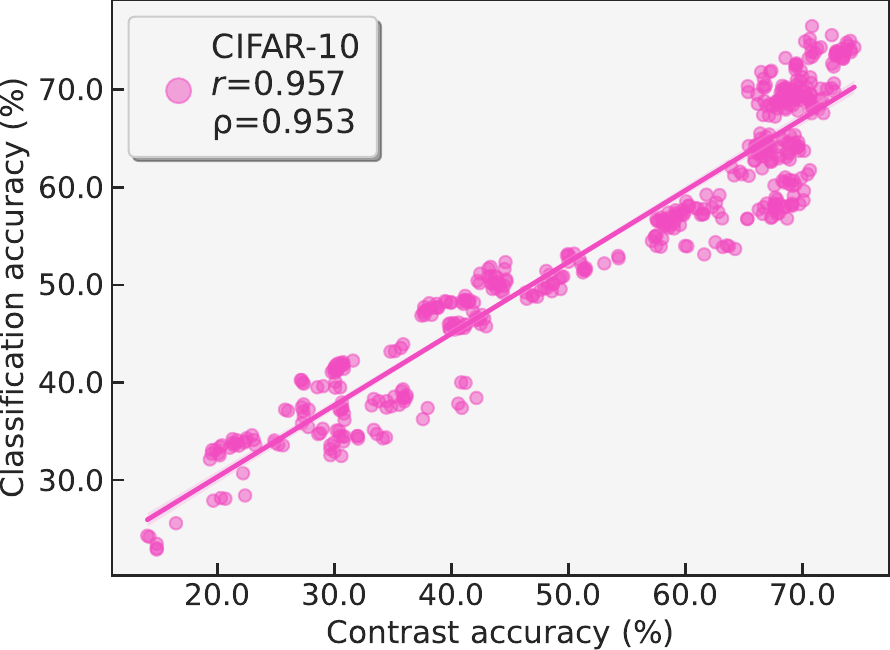}
        \caption{$\lambda=0.001$}
    \end{subfigure}
    
    \caption{Scatter plots of the linear correlation with different contrastive learning task weights ($\lambda$).}
    \label{conloss_weight_app}
\end{figure*}
In this section, we aim to delve into the tightness of the bounds defined by Theorem \ref{Guarantees for General Encoders}, specifically examining whether the variance term under domain shift without fine-tuning on the test set can be neglected.
Revisiting the literature~\cite{wang2022chaos}, eliminating the troublesome variance term must be based on the concurrent fulfillment of the \emph{Intra-class Connectivity} and \emph{Perfect alignment} assumptions.
The \emph{Intra-class Connectivity} is defined as follows: for a given data set $\mathcal{D}$, there exists a appropriate augmentation set $\mathcal{T}$ in which  different intra-class samples can overlap with an aggressive augmentation from $\mathcal{T}$.
The \emph{Perfect alignment} means that the classifier has a minimum InfoNCE Loss. 
To this end, we did an interesting visualization experiment on the CIFAR-10.1 test set. 

In Figure \ref{tsne_by_class_test}, we present an intriguing visualization of the CIFAR-10.1 test set.
Our method concurrently attains high contrastive accuracy (88.47\%) and well intra-class clustering effect in CIFAR-10.1. 
This implies that we are able to fulfill the above-mentioned assumptions of negligible variance term under the shifted test set.
These results further guaranteed that our CL accuracy can being a good indicator of classification accuracy in widely spread unseen test distributions.

\section{Linear correlation with different training settings.}
\begin{figure*}[t]
    \centering
    \begin{subfigure}{0.33\textwidth}
        \centering
        \includegraphics[width=\textwidth]{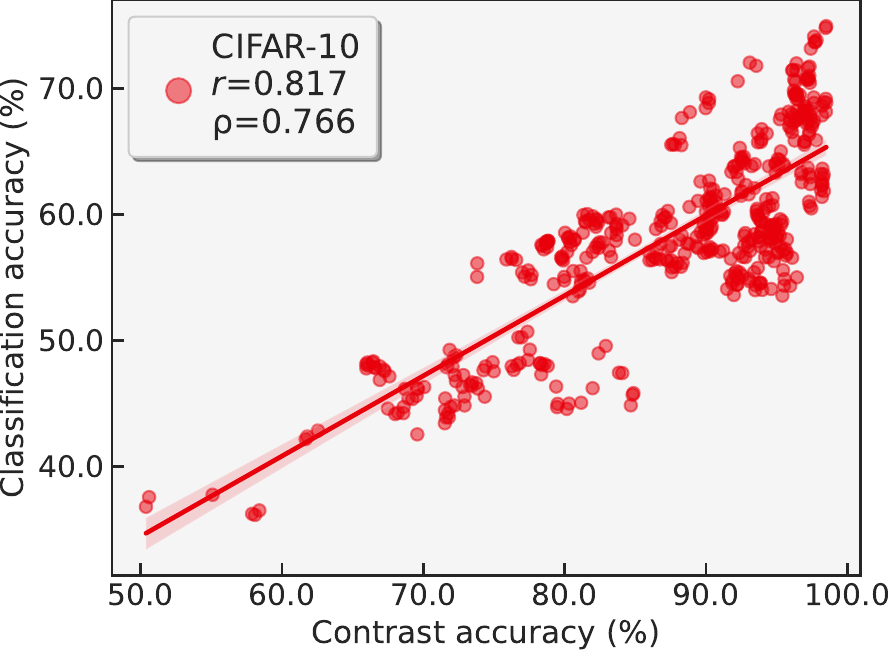}
        \caption{$aug=0.3$}
    \end{subfigure}
    \begin{subfigure}{0.33\textwidth}
        \centering
        \includegraphics[width=\textwidth]{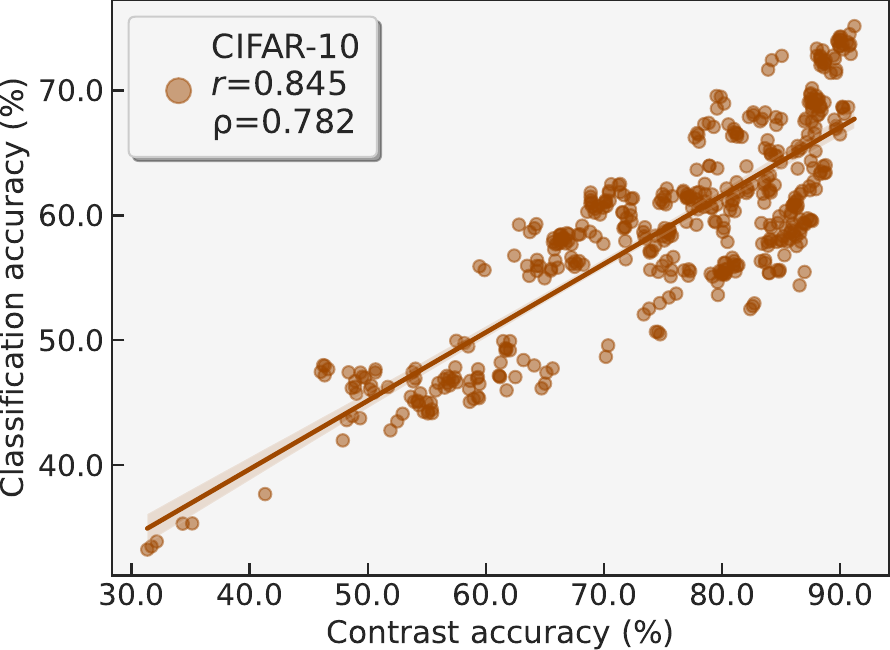}
        \caption{$aug=0.6$}
    \end{subfigure}
    \begin{subfigure}{0.33\textwidth}
        \centering
        \includegraphics[width=\textwidth]{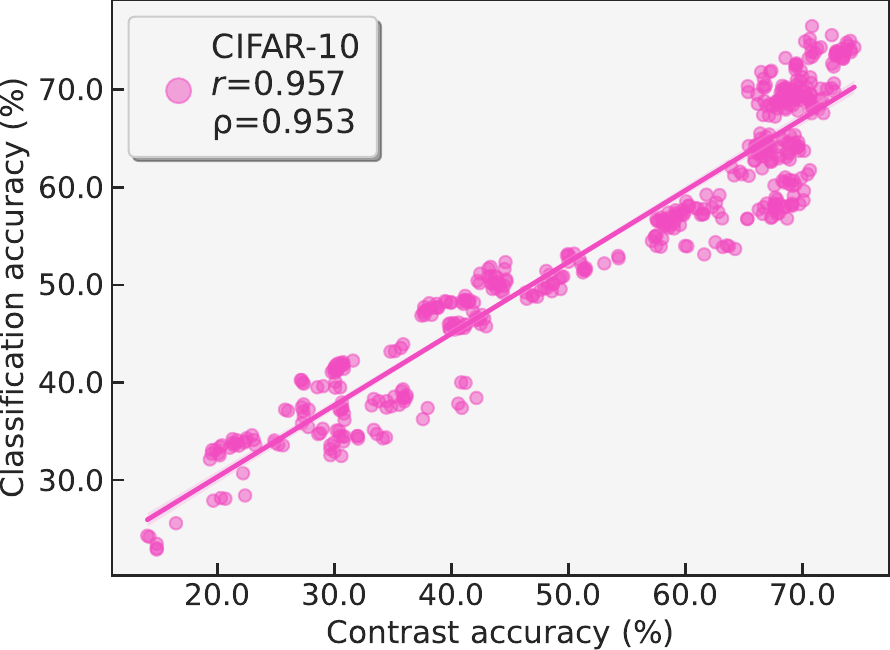}
        \caption{$aug=0.92$}
    \end{subfigure}
    \begin{subfigure}{0.33\textwidth}
        \centering
        \includegraphics[width=\textwidth]{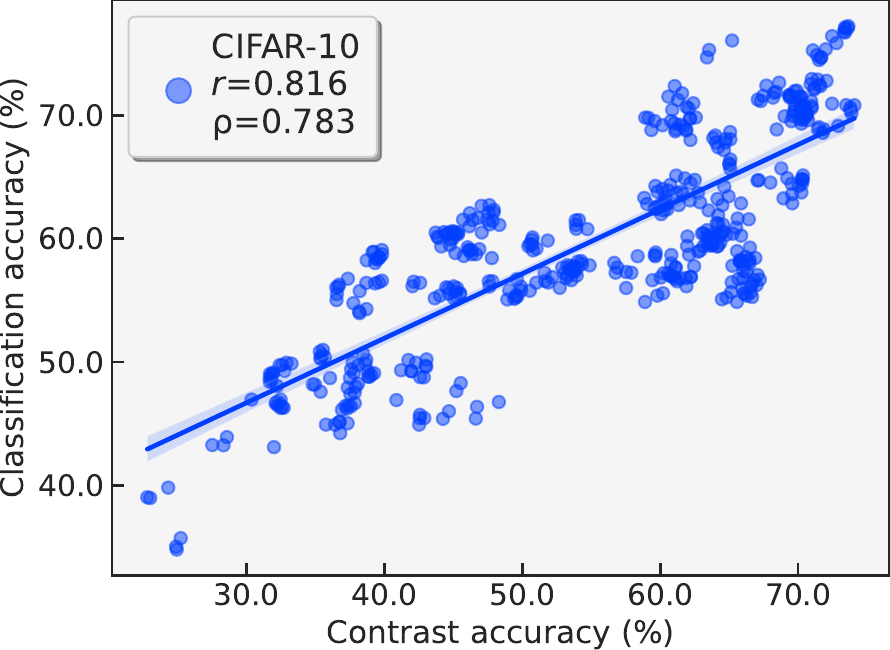}
        \caption{$aug=1.2$}
    \end{subfigure}
    \begin{subfigure}{0.33\textwidth}
        \centering
        \includegraphics[width=\textwidth]{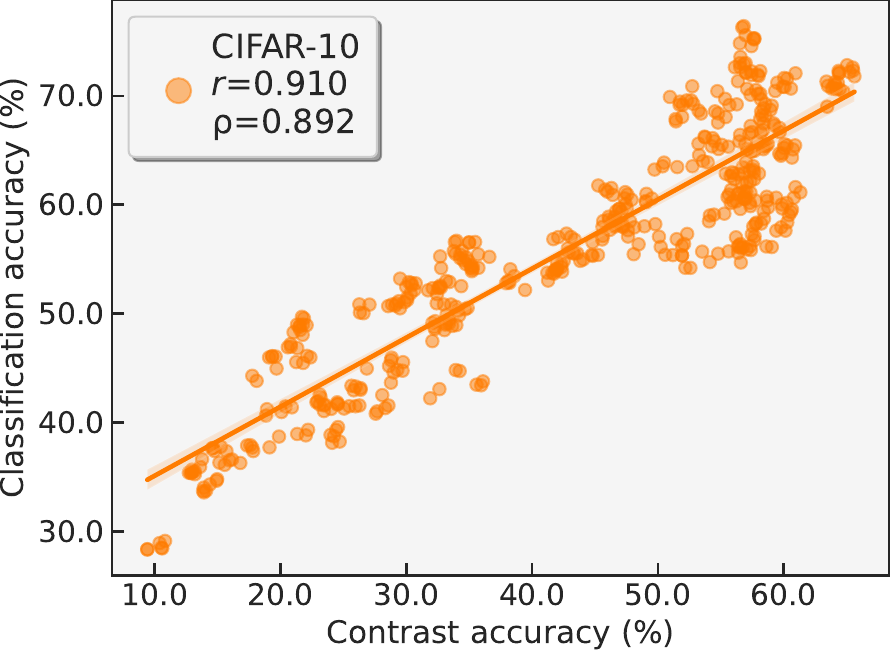}
        \caption{$aug=1.5$}
    \end{subfigure}
    \begin{subfigure}{0.33\textwidth}
        \centering
        \includegraphics[width=\textwidth]{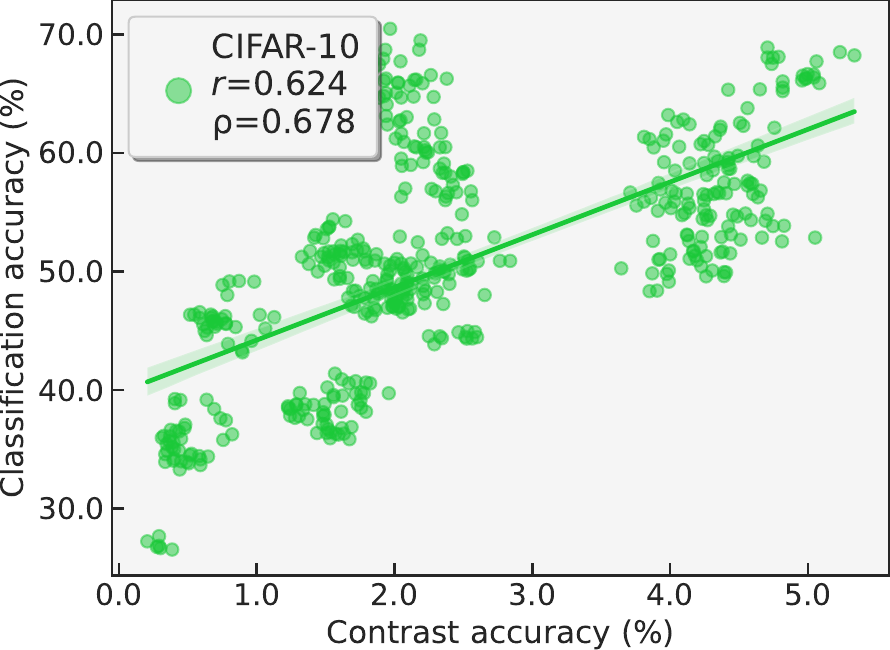}
        \caption{$aug=1.95$}
    \end{subfigure}
    
    \caption{Scatter plots of the linear correlation with different RandomResizedCrop augmentation strengths ($aug$).}
    \label{RRC_app}
\end{figure*}

\begin{figure*}[t]
    \centering
    \begin{subfigure}{0.33\textwidth}
        \centering
        \includegraphics[width=\textwidth]{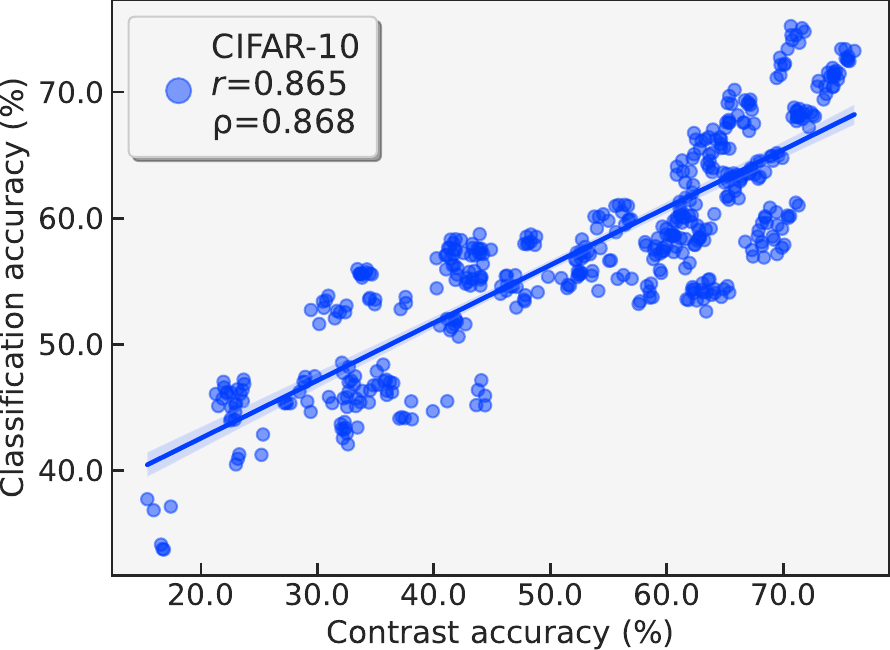}
        \caption{$brightness=0$}
    \end{subfigure}
    \begin{subfigure}{0.33\textwidth}
        \centering
        \includegraphics[width=\textwidth]{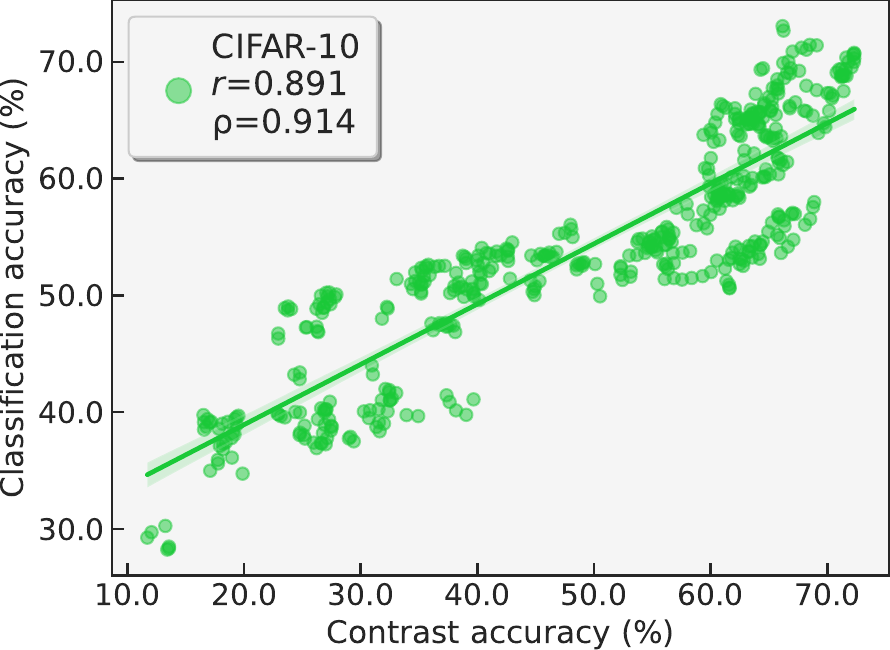}
        \caption{$brightness=0.3$}
    \end{subfigure}
    \begin{subfigure}{0.33\textwidth}
        \centering
        \includegraphics[width=\textwidth]{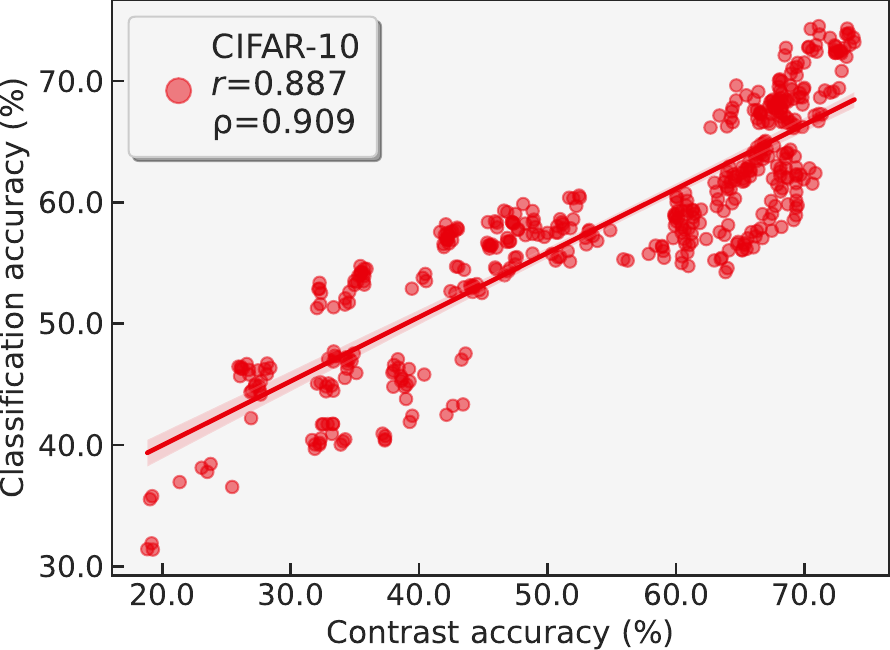}
        \caption{$brightness=0.5$}
    \end{subfigure}
    \begin{subfigure}{0.33\textwidth}
        \centering
        \includegraphics[width=\textwidth]{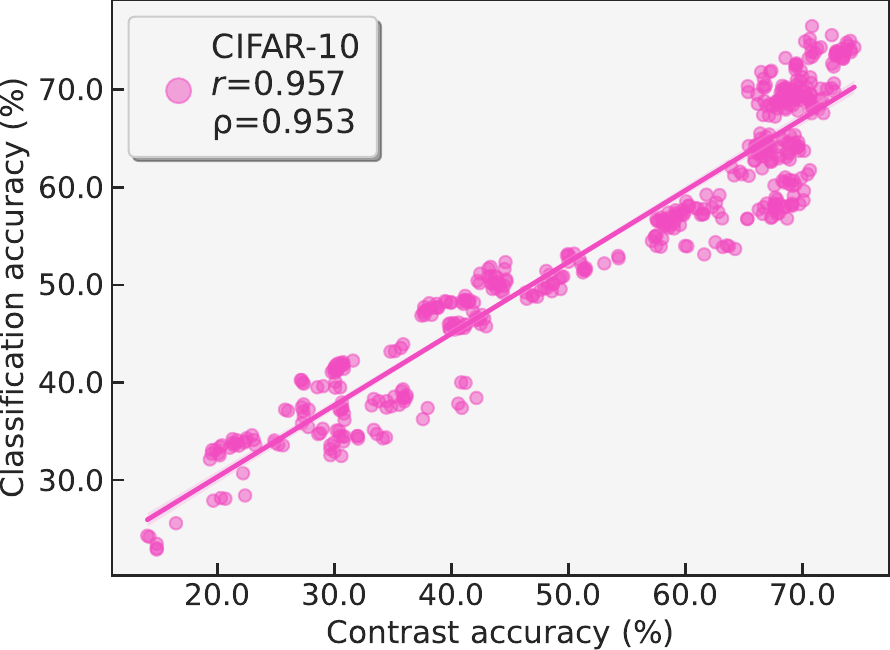}
        \caption{$brightness=0.8$}
    \end{subfigure}
    \begin{subfigure}{0.33\textwidth}
        \centering
        \includegraphics[width=\textwidth]{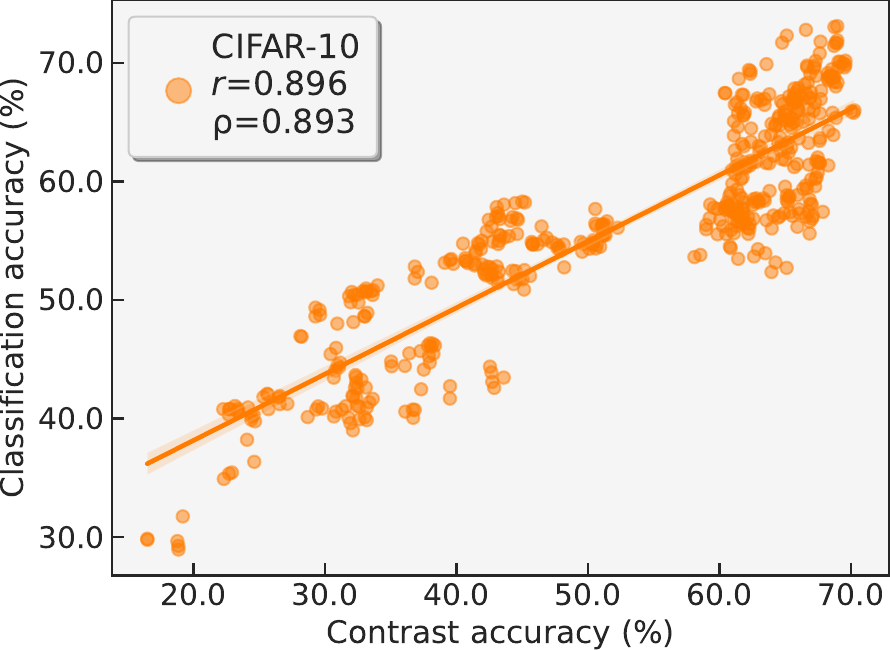}
        \caption{$brightness=1.0$}
    \end{subfigure}
    
    \caption{Scatter plots of the linear correlation with different color jittering parameter $brightness$.}
    \label{brightness_app}
\end{figure*}

\begin{figure*}[t]
    \centering
    \begin{subfigure}{0.33\textwidth}
        \centering
        \includegraphics[width=\textwidth]{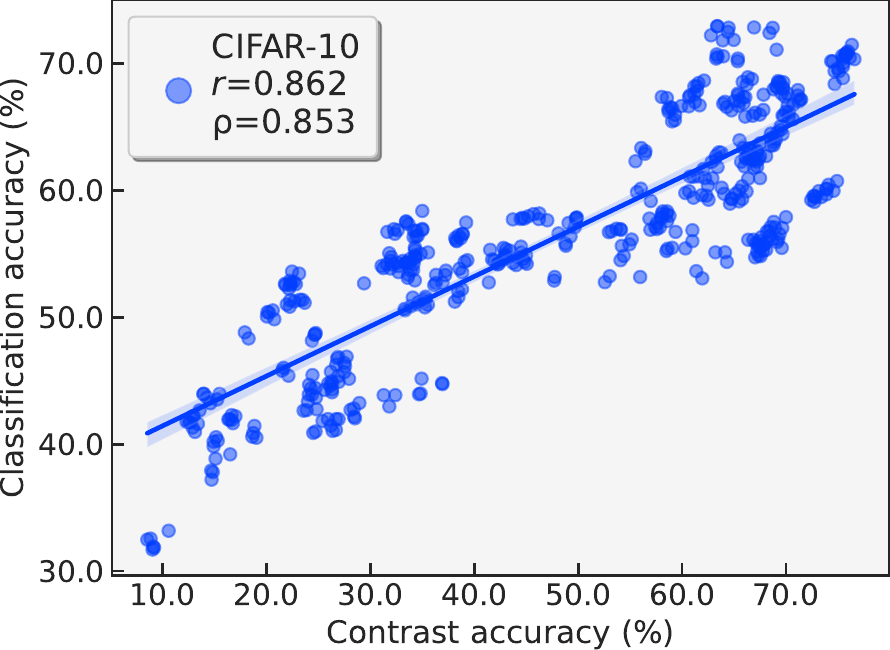}
        \caption{$contrast=0$}
    \end{subfigure}
    \begin{subfigure}{0.33\textwidth}
        \centering
        \includegraphics[width=\textwidth]{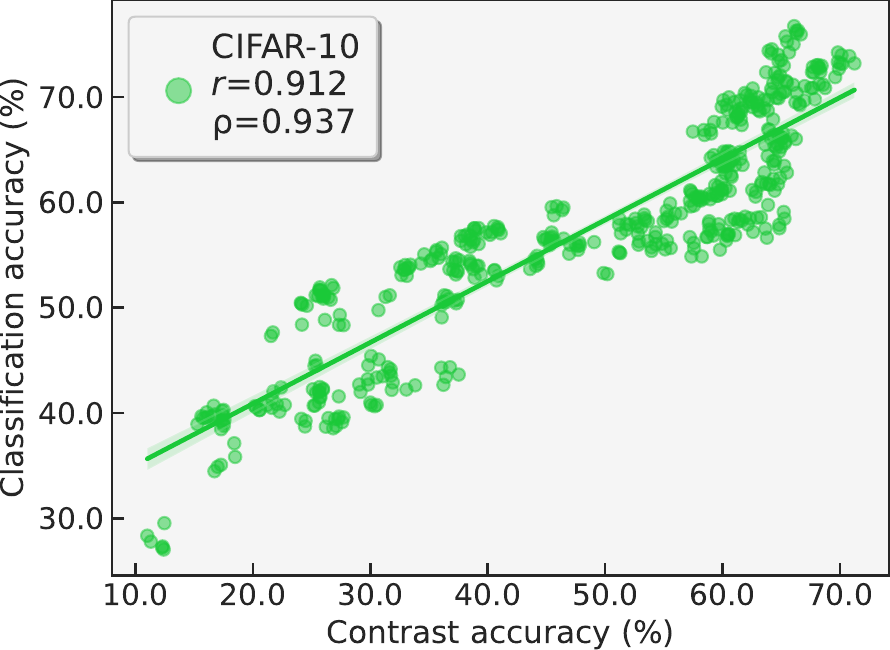}
        \caption{$contrast=0.3$}
    \end{subfigure}
    \begin{subfigure}{0.33\textwidth}
        \centering
        \includegraphics[width=\textwidth]{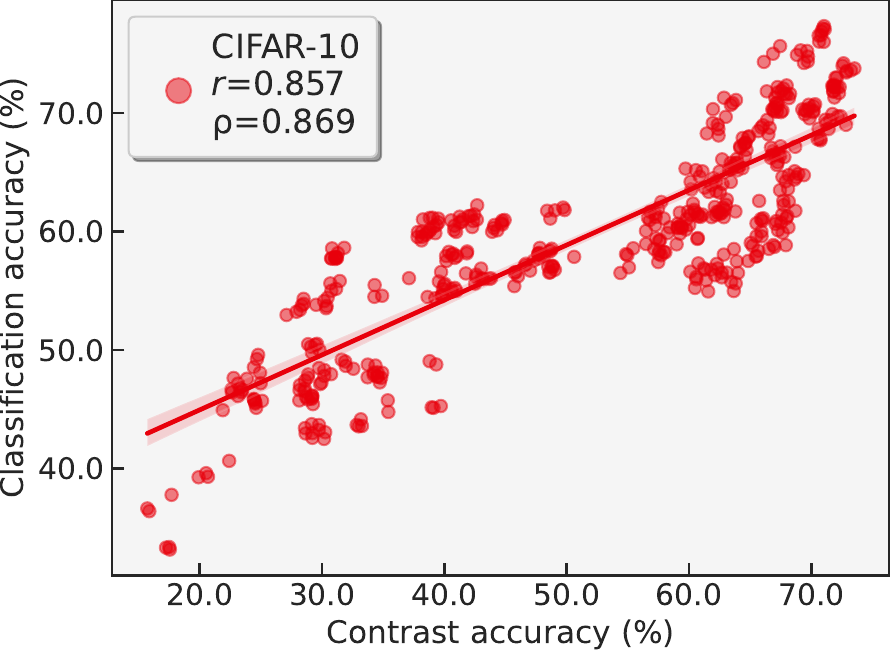}
        \caption{$contrast=0.5$}
    \end{subfigure}
    \begin{subfigure}{0.33\textwidth}
        \centering
        \includegraphics[width=\textwidth]{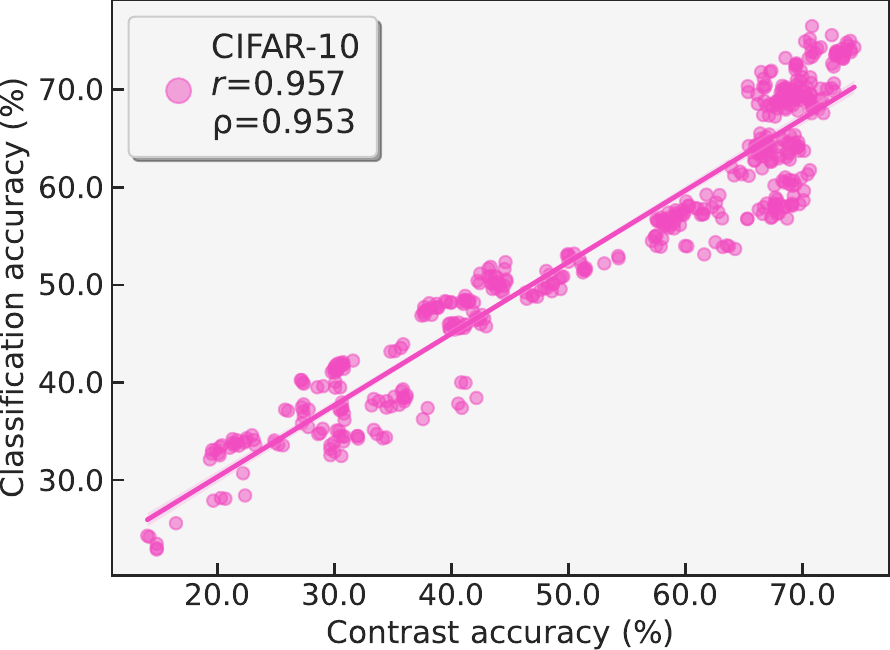}
        \caption{$contrast=0.8$}
    \end{subfigure}
    \begin{subfigure}{0.33\textwidth}
        \centering
        \includegraphics[width=\textwidth]{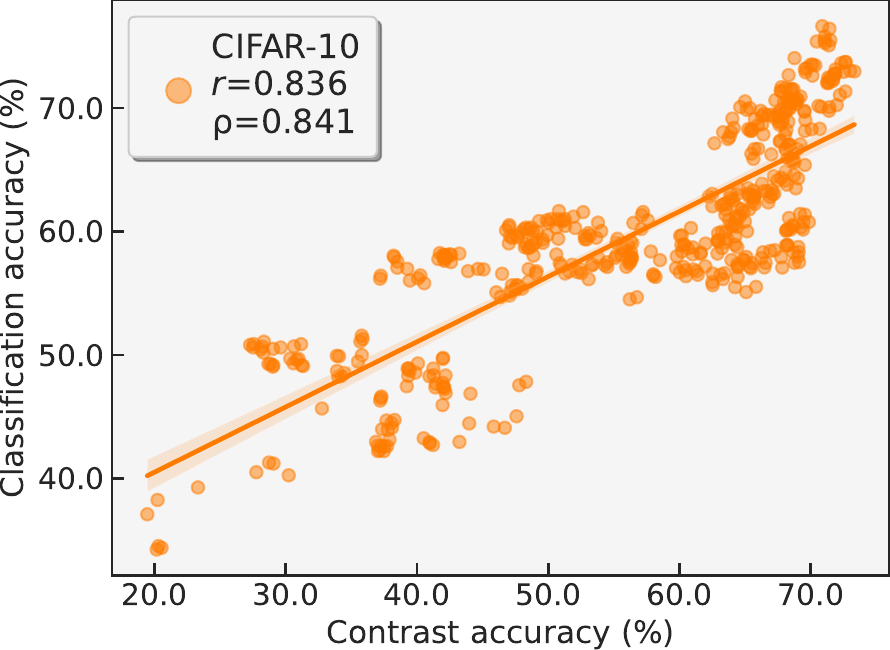}
        \caption{$contrast=1.0$}
    \end{subfigure}
    
    \caption{Scatter plots of the linear correlation with different color jittering parameter $contrast$.}
    \label{contrast_app}
\end{figure*}

\begin{figure*}[t]
    \centering
    \begin{subfigure}{0.33\textwidth}
        \centering
        \includegraphics[width=\textwidth]{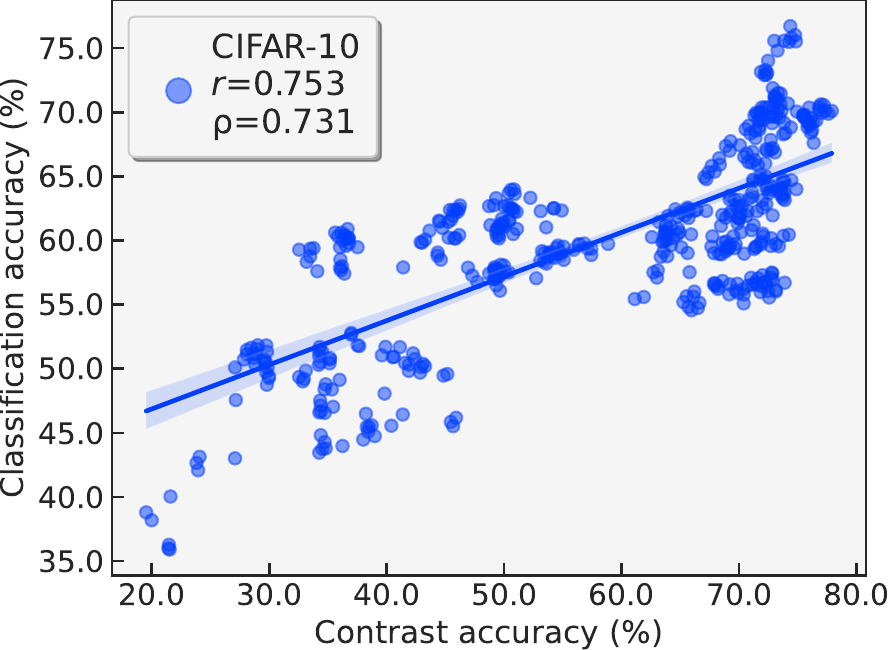}
        \caption{$saturation=0$}
    \end{subfigure}
    \begin{subfigure}{0.33\textwidth}
        \centering
        \includegraphics[width=\textwidth]{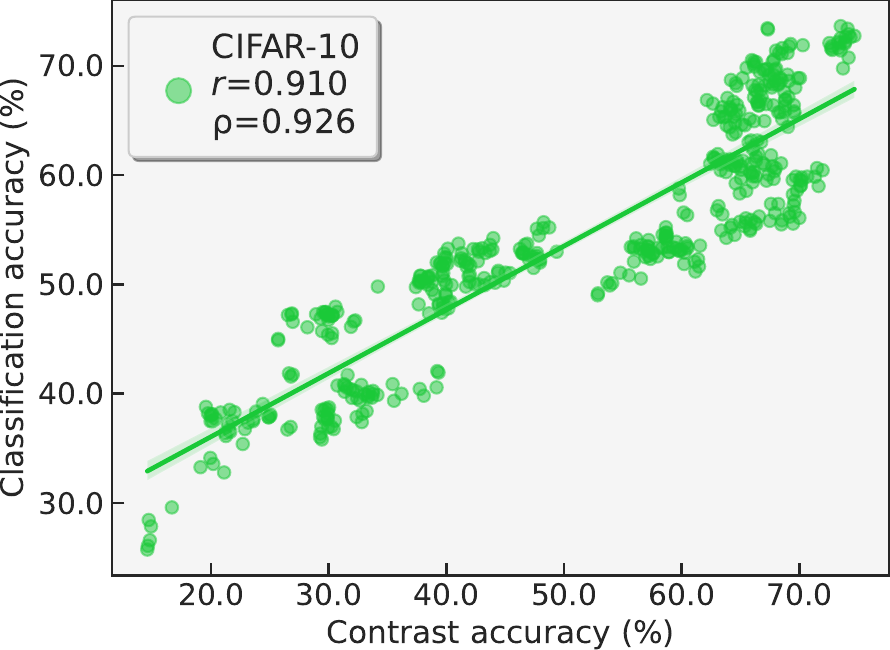}
        \caption{$saturation=0.3$}
    \end{subfigure}
    \begin{subfigure}{0.33\textwidth}
        \centering
        \includegraphics[width=\textwidth]{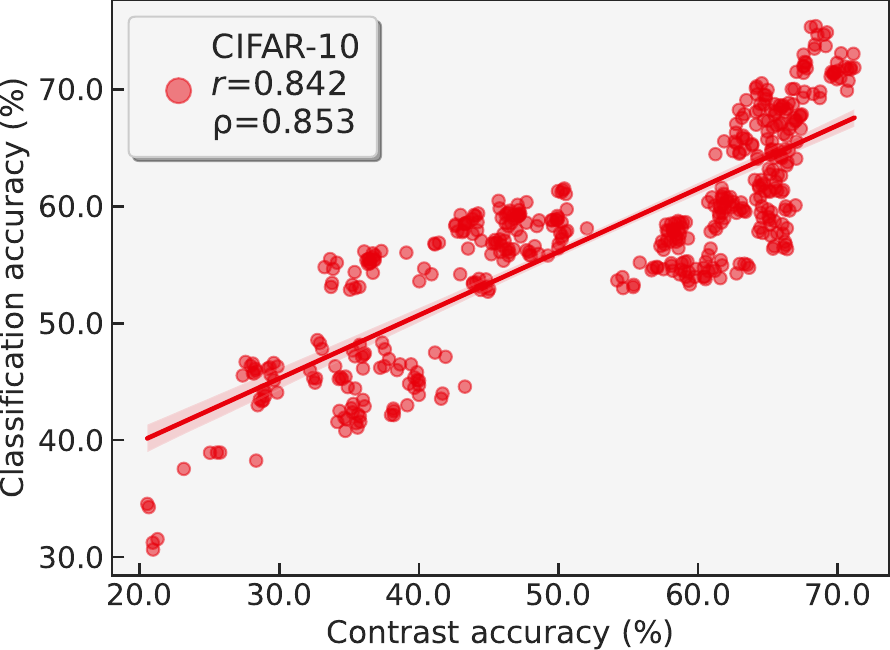}
        \caption{$saturation=0.5$}
    \end{subfigure}
    \begin{subfigure}{0.33\textwidth}
        \centering
        \includegraphics[width=\textwidth]{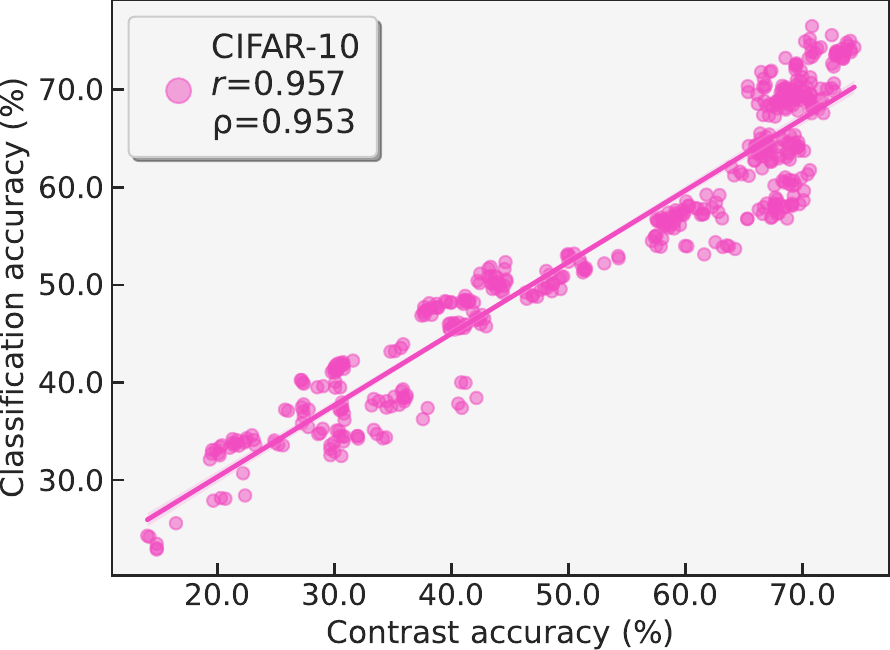}
        \caption{$saturation=0.8$}
    \end{subfigure}
    \begin{subfigure}{0.33\textwidth}
        \centering
        \includegraphics[width=\textwidth]{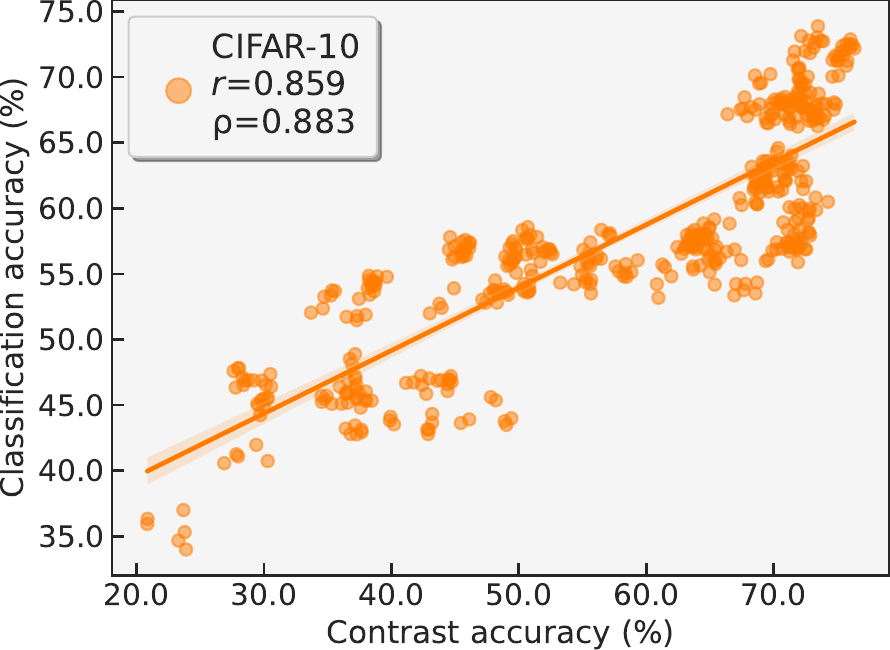}
        \caption{$saturation=1.0$}
    \end{subfigure}
    
    \caption{Scatter plots of the linear correlation with different color jittering parameter $saturation$.}
    \label{saturation_app}
\end{figure*}

\begin{figure*}[t]
    \centering
    \begin{subfigure}{0.33\textwidth}
        \centering
        \includegraphics[width=\textwidth]{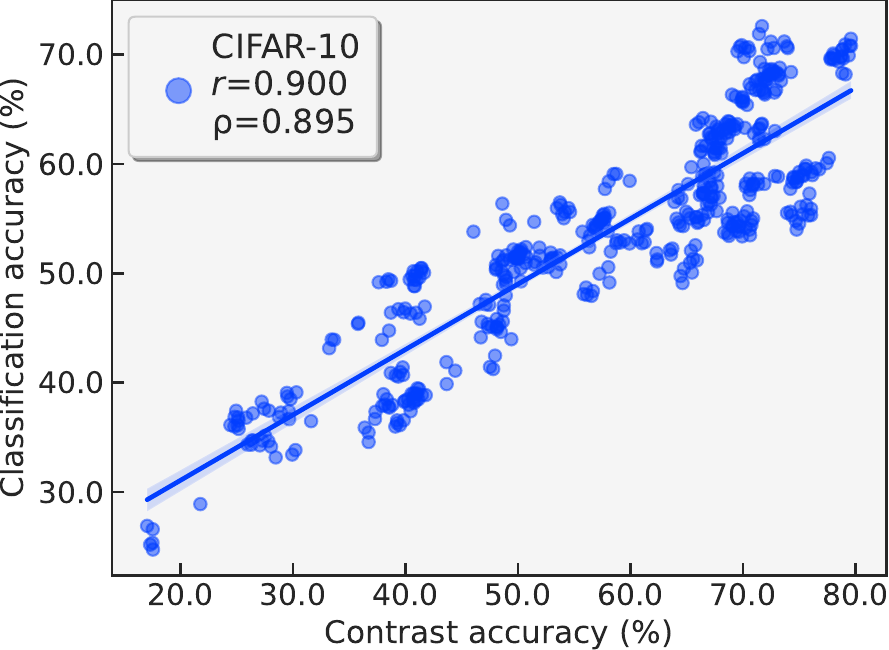}
        \caption{$hue=0$}
    \end{subfigure}
    \begin{subfigure}{0.33\textwidth}
        \centering
        \includegraphics[width=\textwidth]{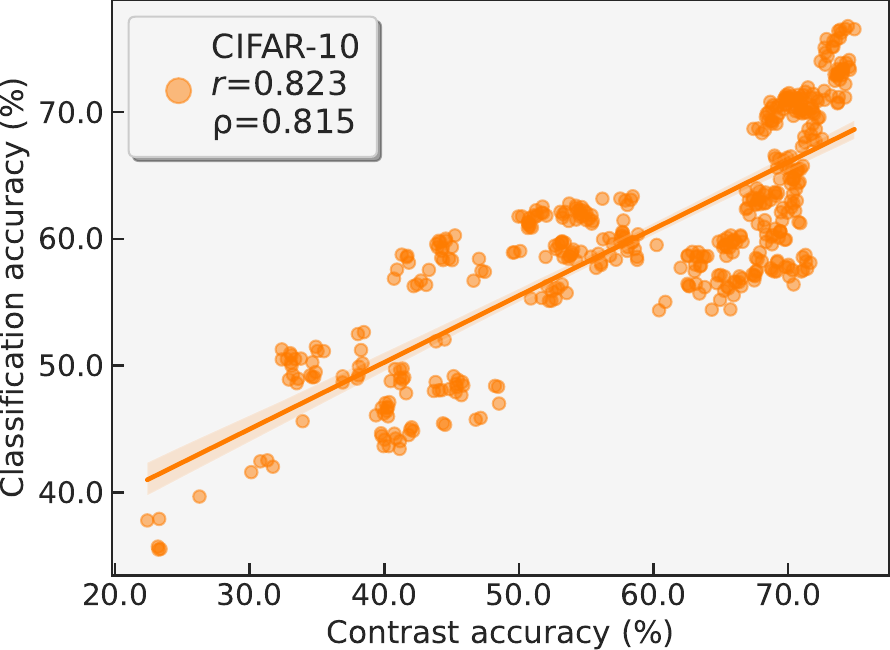}
        \caption{$hue=0.1$}
    \end{subfigure}
    \begin{subfigure}{0.33\textwidth}
        \centering
        \includegraphics[width=\textwidth]{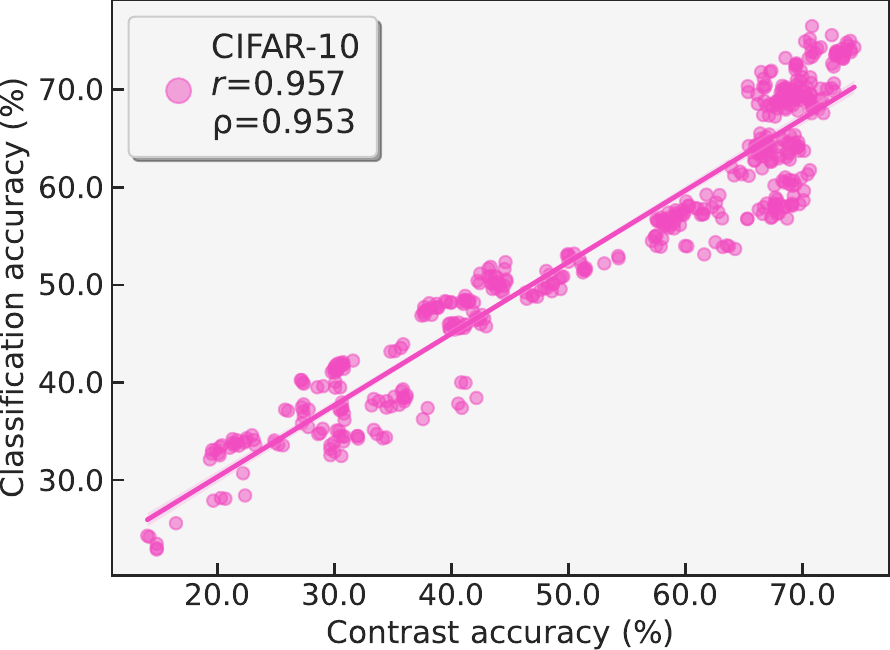}
        \caption{$hue=0.2$}
    \end{subfigure}
    \begin{subfigure}{0.33\textwidth}
        \centering
        \includegraphics[width=\textwidth]{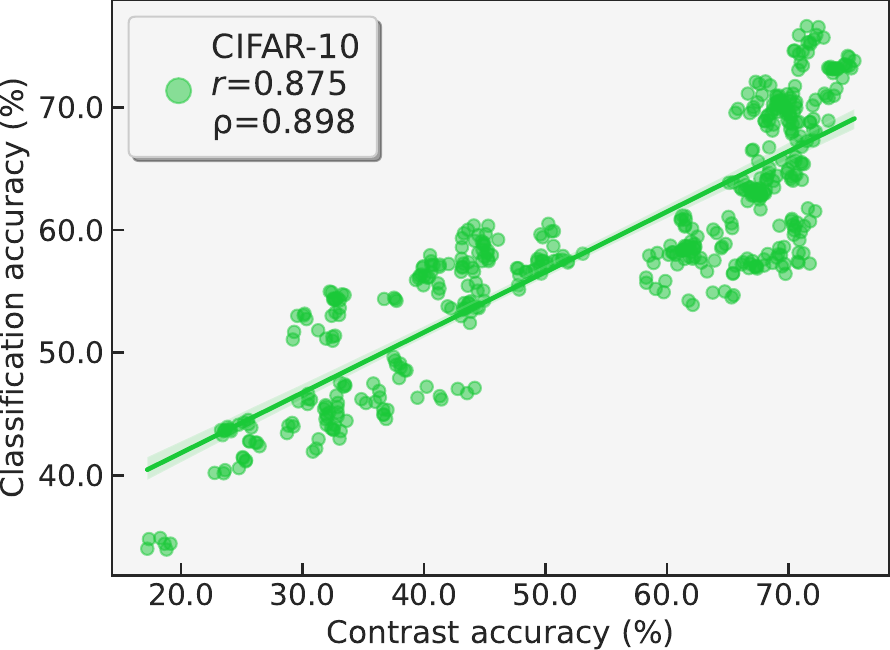}
        \caption{$hue=0.4$}
    \end{subfigure}
    \begin{subfigure}{0.33\textwidth}
        \centering
        \includegraphics[width=\textwidth]{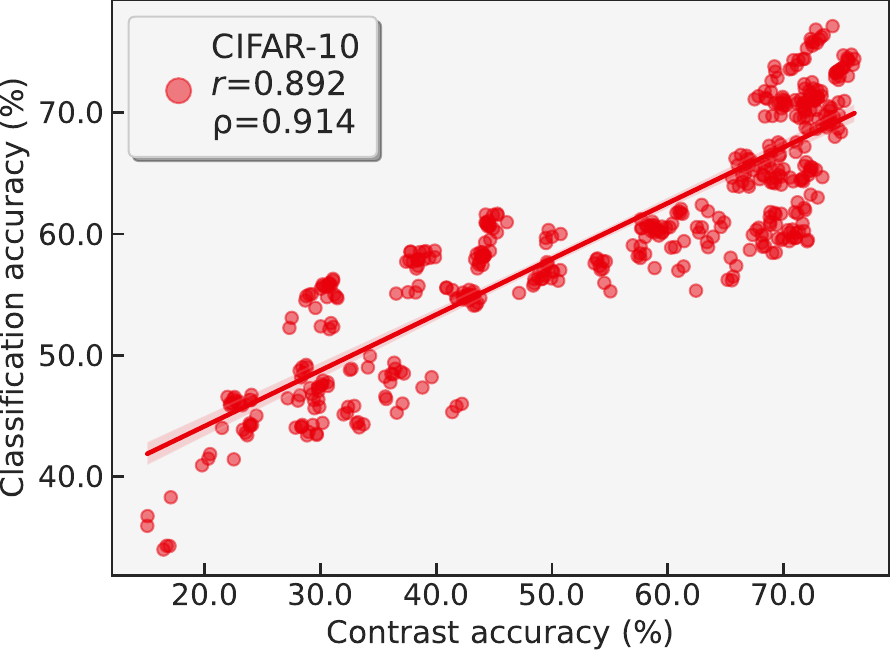}
        \caption{$hue=0.5$}
    \end{subfigure}
    
    \caption{Scatter plots of the linear correlation with different color jittering parameter $hue$.}
    \label{hue_app}
\end{figure*}

\begin{figure*}[t]
    \centering
    \begin{subfigure}{0.33\textwidth}
        \centering
        \includegraphics[width=\textwidth]{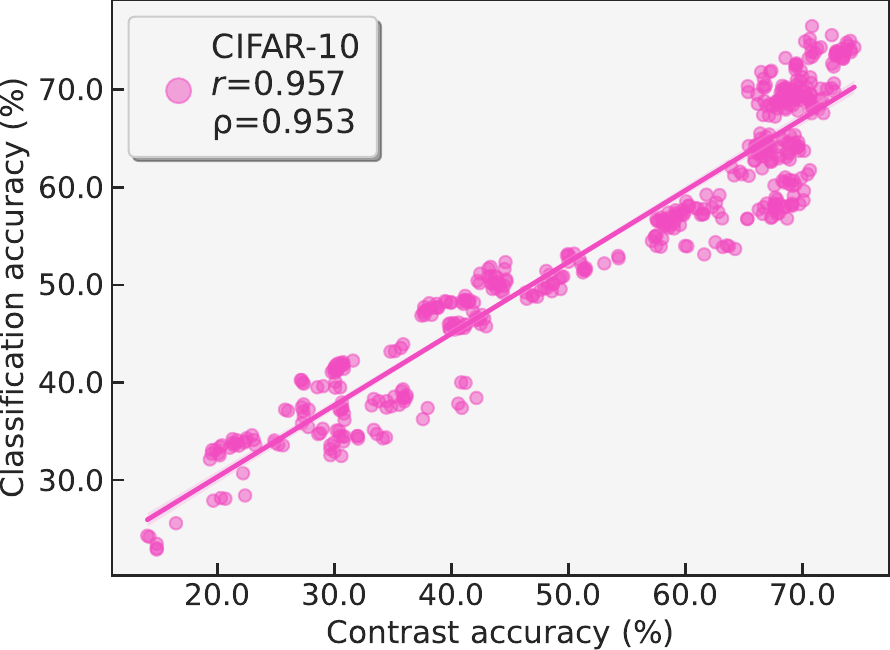}
        \caption{DenseNet-40-12}
    \end{subfigure}
    \begin{subfigure}{0.33\textwidth}
        \centering
        \includegraphics[width=\textwidth]{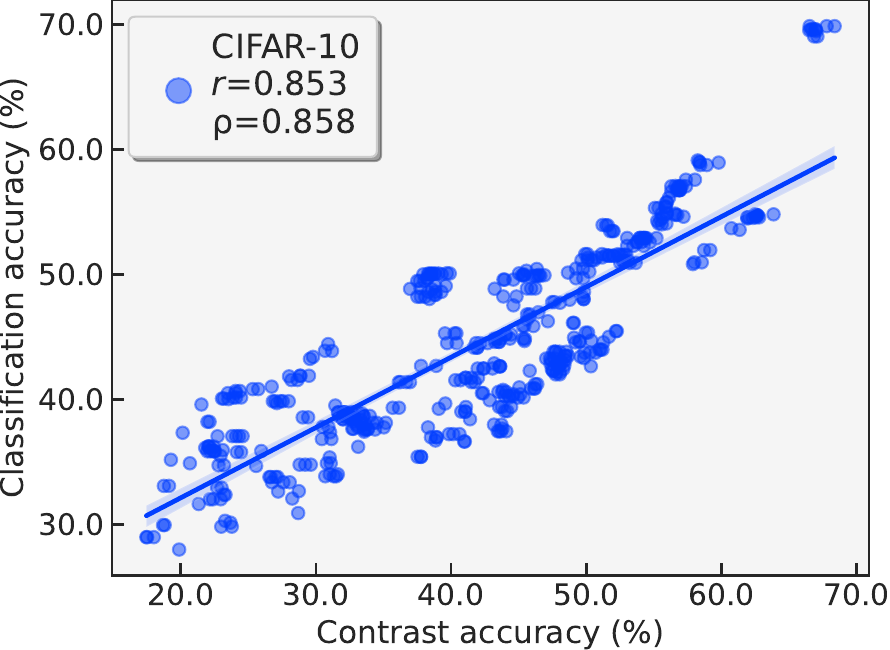}
        \caption{ResNet-18}
    \end{subfigure}
    \begin{subfigure}{0.33\textwidth}
        \centering
        \includegraphics[width=\textwidth]{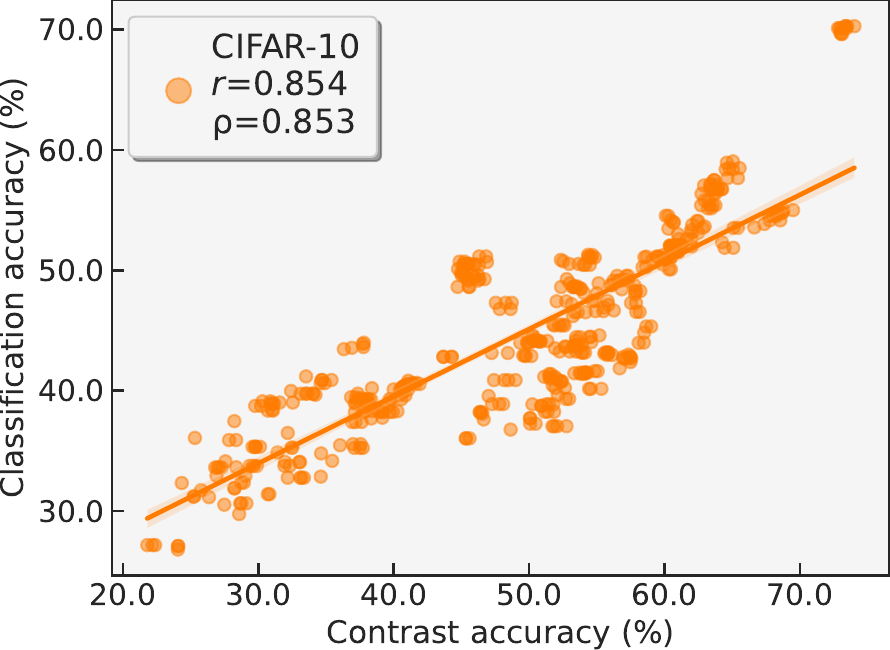}
        \caption{ResNet-34}
    \end{subfigure}
    \begin{subfigure}{0.33\textwidth}
        \centering
        \includegraphics[width=\textwidth]{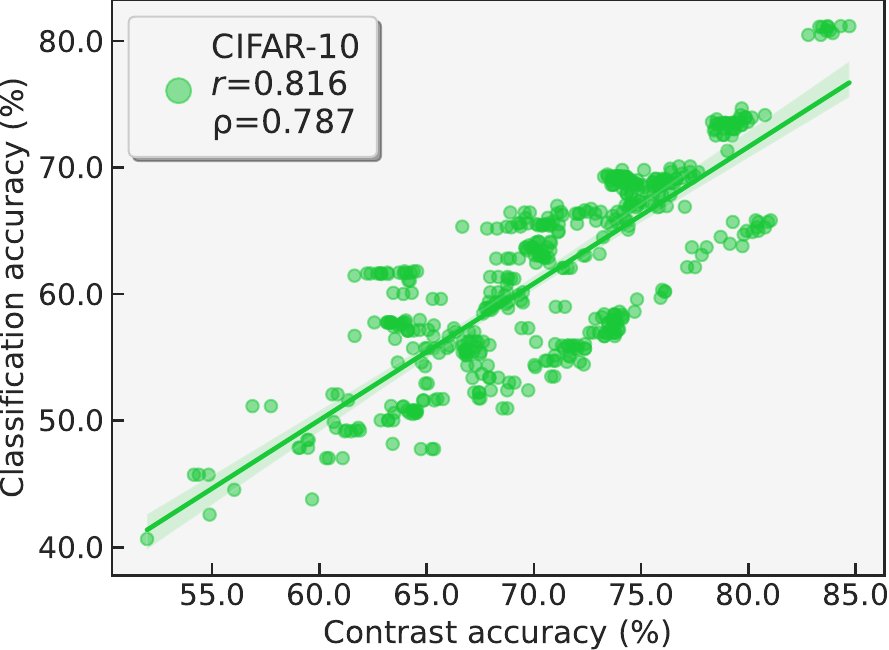}
        \caption{VGG-11}
    \end{subfigure}
    \begin{subfigure}{0.33\textwidth}
        \centering
        \includegraphics[width=\textwidth]{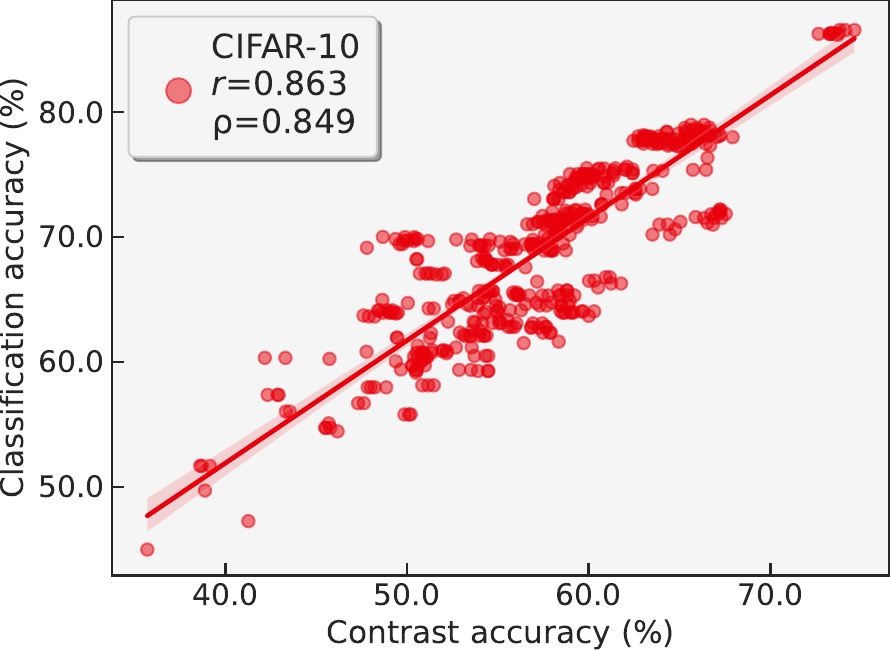}
        \caption{VGG-19}
    \end{subfigure}
    
    \caption{Scatter plots of the linear correlation with different backbones.}
    \label{backbone_app}
\end{figure*}

\begin{figure*}[t]
    \centering
    \begin{subfigure}{0.33\textwidth}
        \centering
        \includegraphics[width=\textwidth]{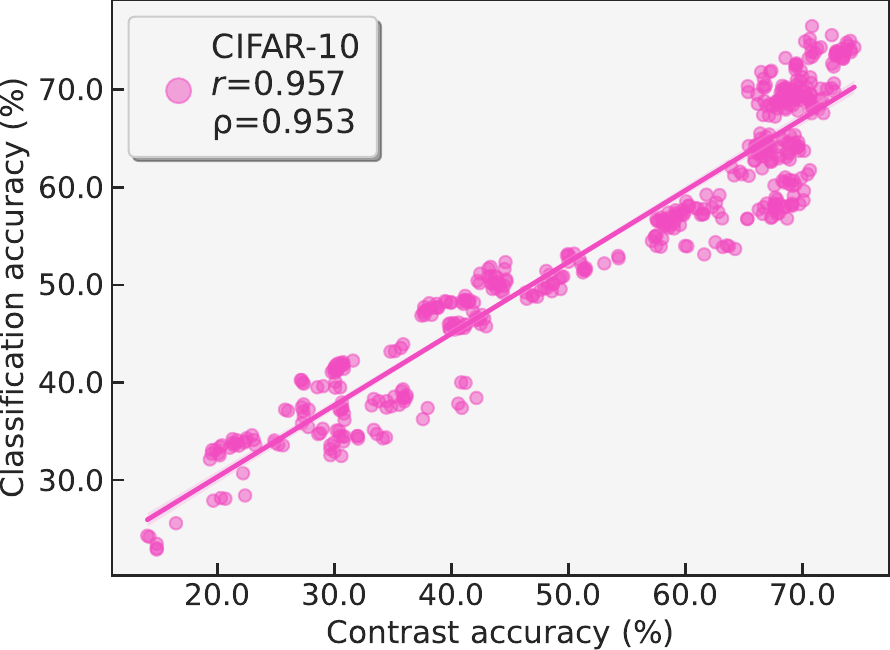}
        \caption{$seed=0$}
    \end{subfigure}
    \begin{subfigure}{0.33\textwidth}
        \centering
        \includegraphics[width=\textwidth]{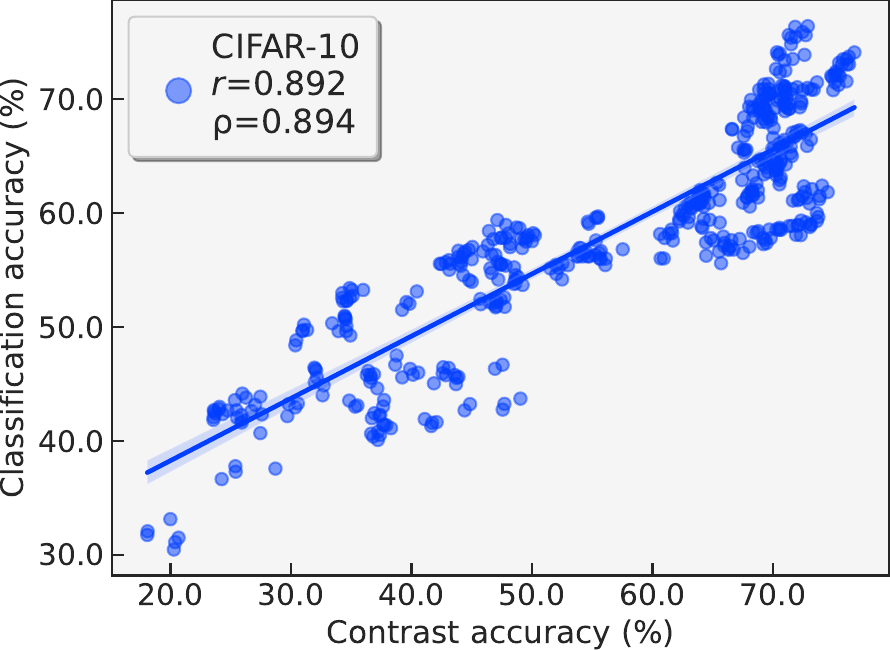}
        \caption{$seed=21$}
    \end{subfigure}
    \begin{subfigure}{0.33\textwidth}
        \centering
        \includegraphics[width=\textwidth]{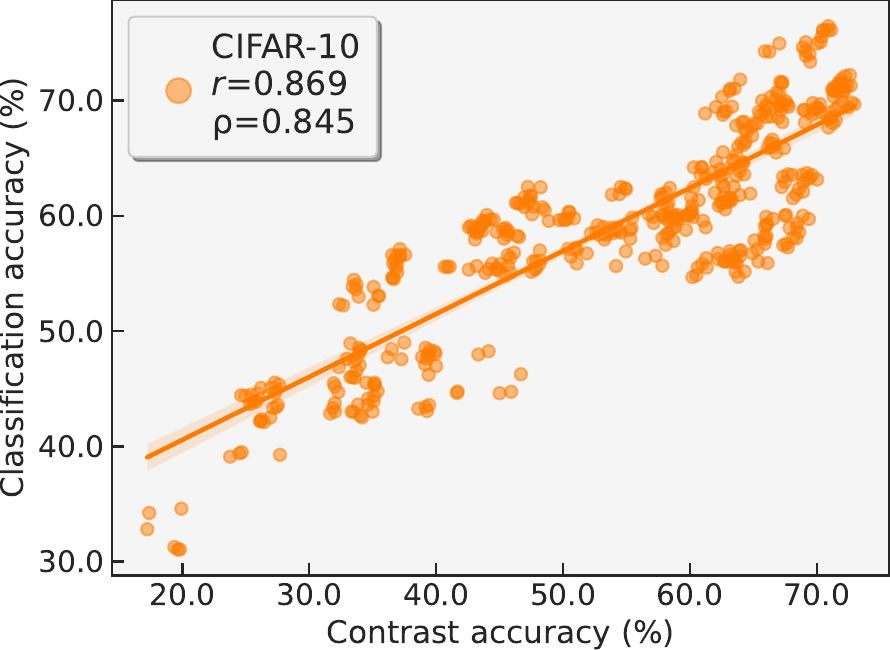}
        \caption{$seed=42$}
    \end{subfigure}
    
    \caption{Scatter plots of the linear correlation with different random seeds.}
    \label{seed_app}
\end{figure*}

\begin{figure*}[t]
    \centering
    \begin{subfigure}{0.33\textwidth}
        \centering
        \includegraphics[width=\textwidth]{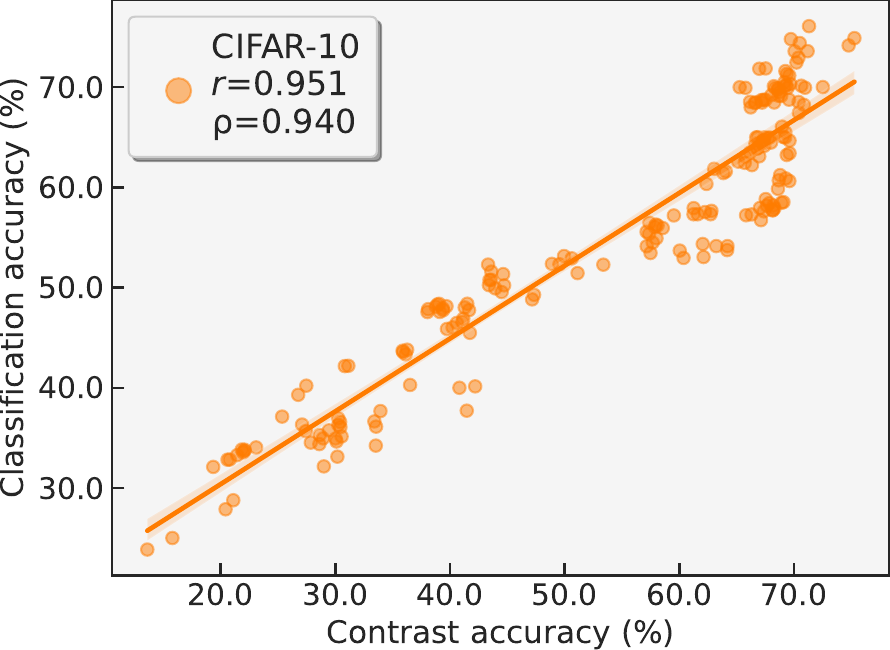}
        \caption{$num=200$}
    \end{subfigure}
    \begin{subfigure}{0.33\textwidth}
        \centering
        \includegraphics[width=\textwidth]{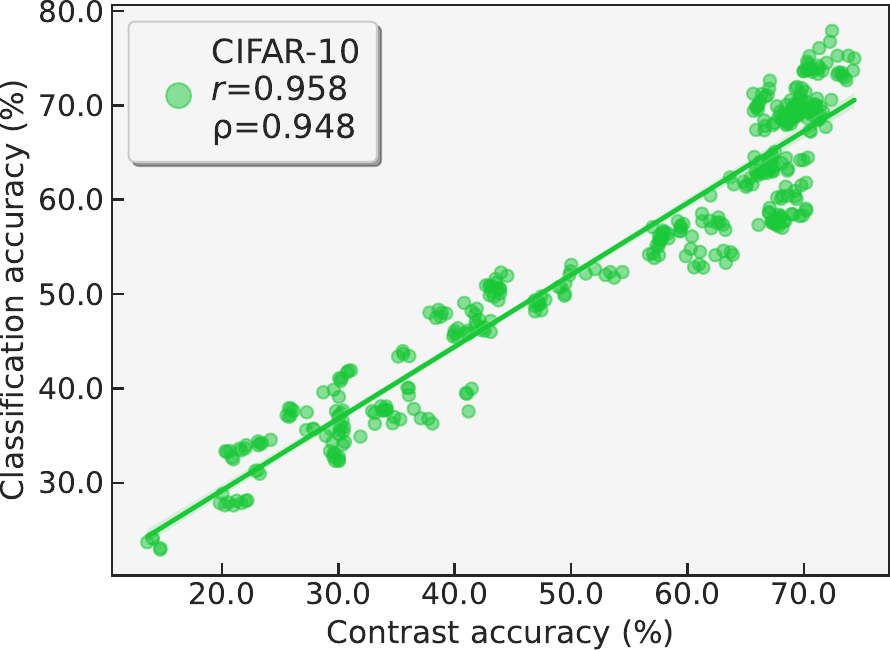}
        \caption{$num=400$}
    \end{subfigure}
    \begin{subfigure}{0.33\textwidth}
        \centering
        \includegraphics[width=\textwidth]{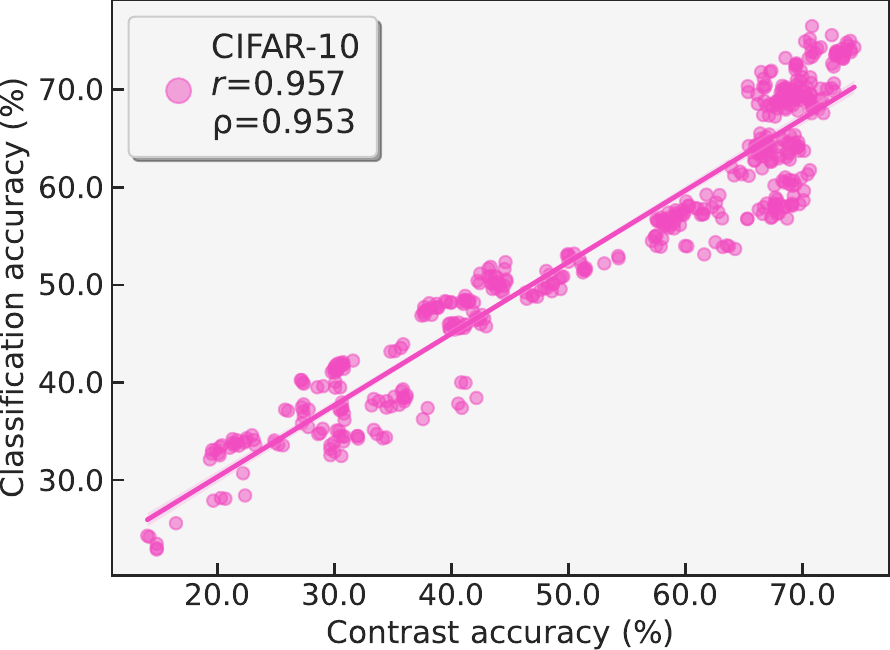}
        \caption{$num=500$}
    \end{subfigure}
    \begin{subfigure}{0.33\textwidth}
        \centering
        \includegraphics[width=\textwidth]{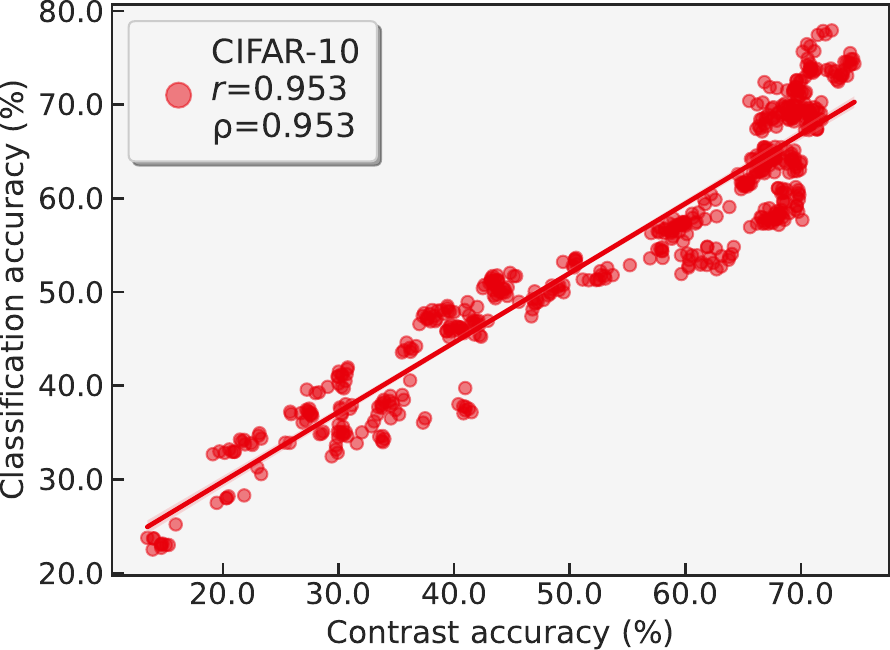}
        \caption{$num=600$}
    \end{subfigure}
    \begin{subfigure}{0.33\textwidth}
        \centering
        \includegraphics[width=\textwidth]{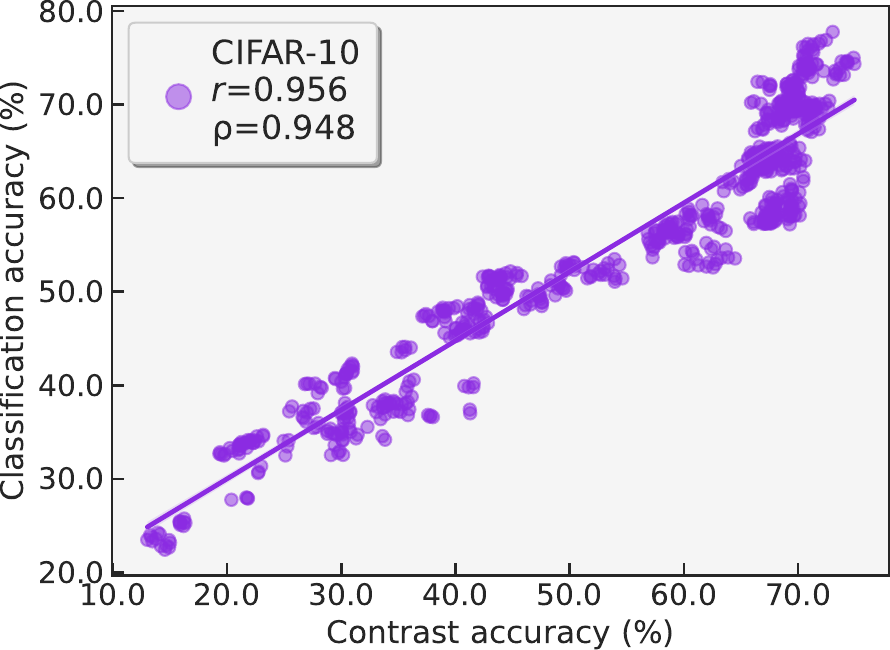}
        \caption{$num=800$}
    \end{subfigure}
    \begin{subfigure}{0.33\textwidth}
        \centering
        \includegraphics[width=\textwidth]{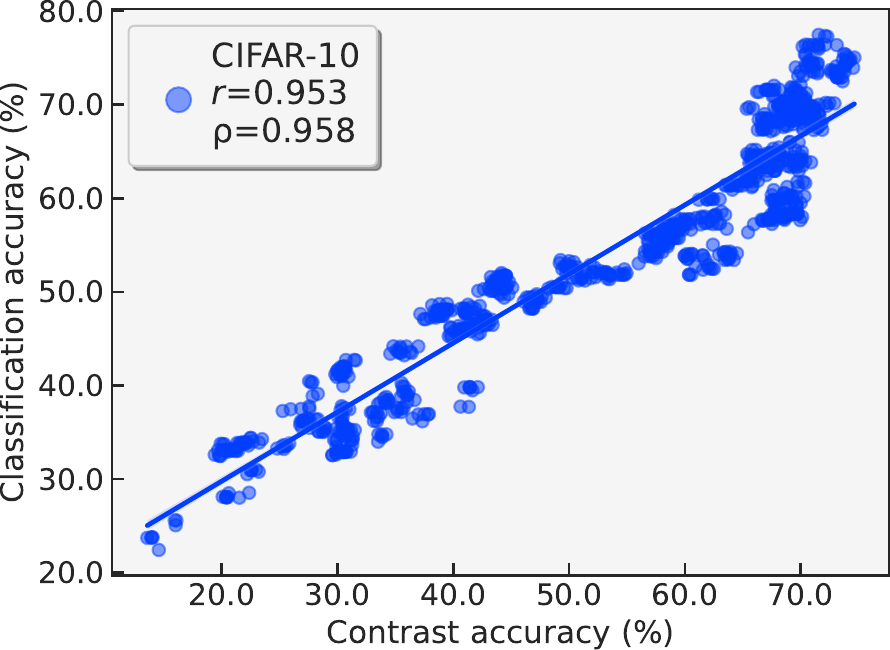}
        \caption{$num=1000$}
    \end{subfigure}
    
    \caption{Scatter plots of the linear correlation with different sample set amounts $num$ (each sample set contains 10000 images).}
    \label{sample_set_num_app}
\end{figure*}

\begin{figure*}[t]
    \centering
    \begin{subfigure}{0.33\textwidth}
        \centering
        \includegraphics[width=\textwidth]{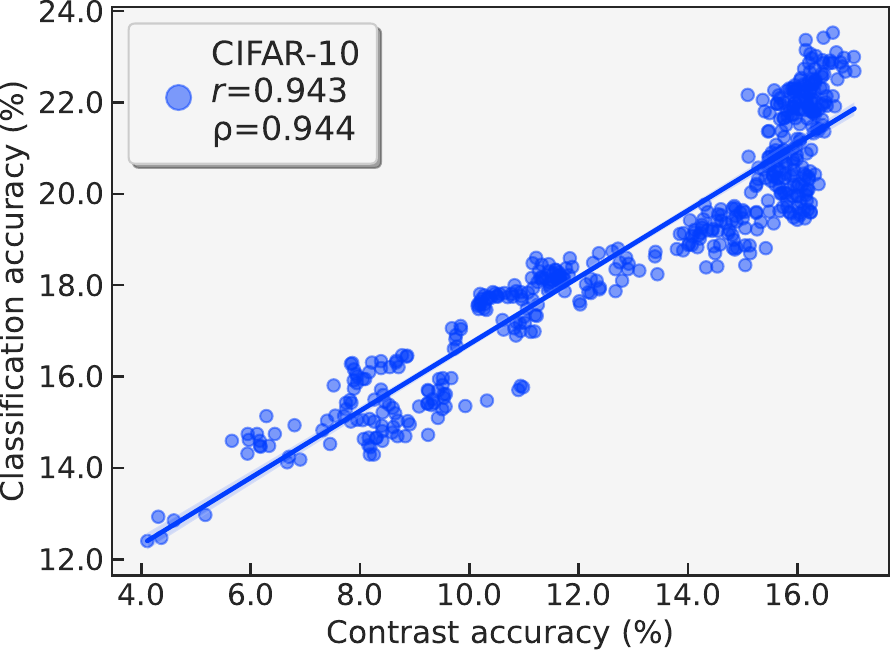}
        \caption{$size=2000$}
    \end{subfigure}
    \begin{subfigure}{0.33\textwidth}
        \centering
        \includegraphics[width=\textwidth]{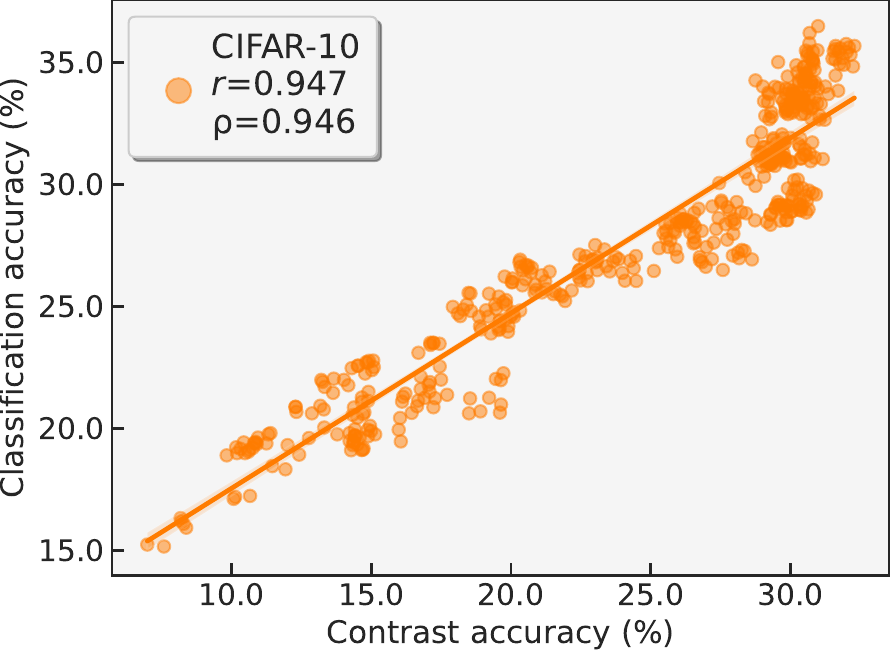}
        \caption{$size=4000$}
    \end{subfigure}
    \begin{subfigure}{0.33\textwidth}
        \centering
        \includegraphics[width=\textwidth]{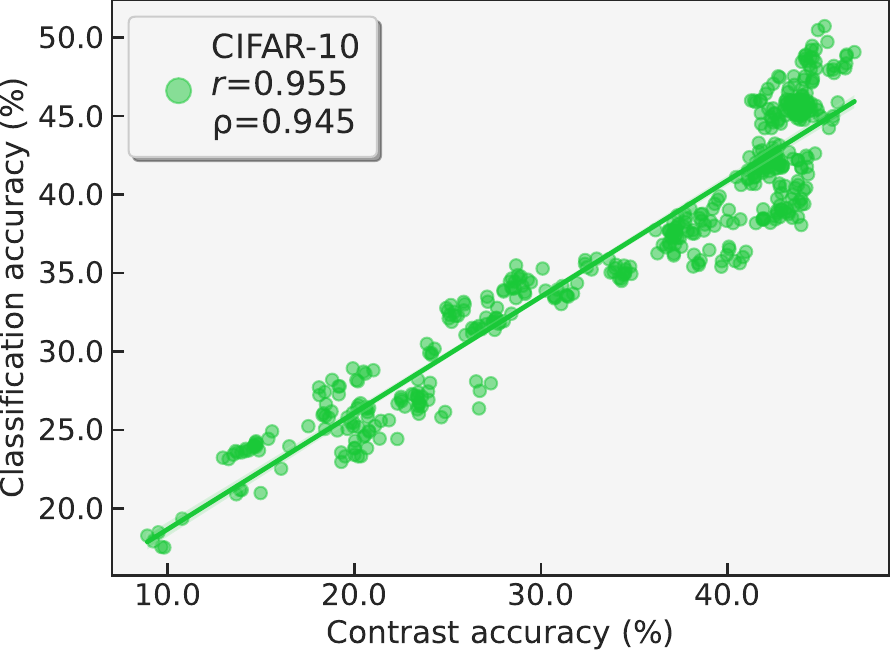}
        \caption{$size=6000$}
    \end{subfigure}
    \begin{subfigure}{0.33\textwidth}
        \centering
        \includegraphics[width=\textwidth]{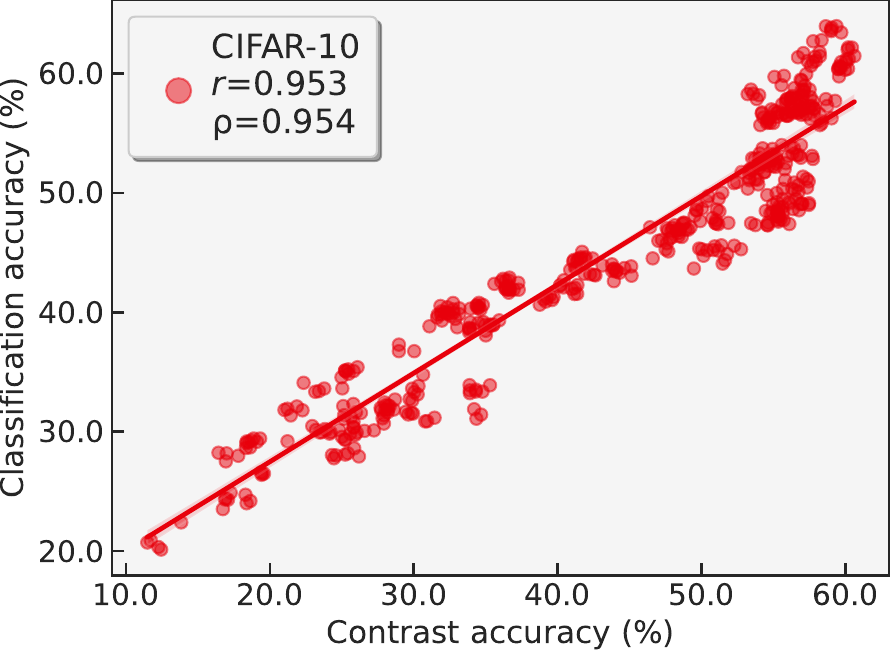}
        \caption{$size=8000$}
    \end{subfigure}
    \begin{subfigure}{0.33\textwidth}
        \centering
        \includegraphics[width=\textwidth]{metaset500_sampleset10k.pdf}
        \caption{$size=10000$}
    \end{subfigure}
    
    \caption{Scatter plots of the linear correlation with different sample set sizes $size$ (using 500 sample sets).}
    \label{sample_set_size_app}
\end{figure*}

\begin{figure*}[t]
    \centering
    \begin{subfigure}{0.33\textwidth}
        \centering
        \includegraphics[width=\textwidth]{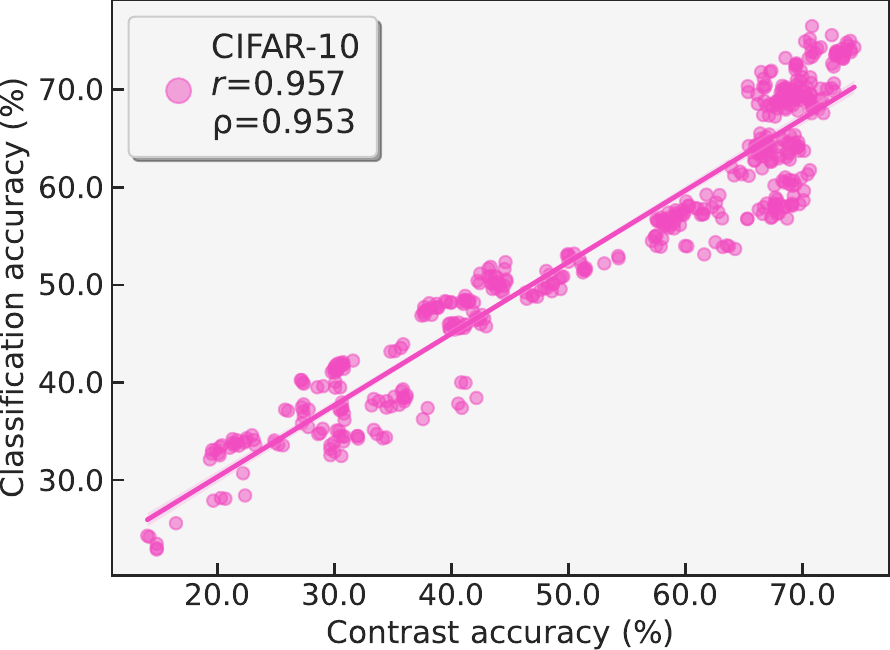}
        \caption{Multi-task}
    \end{subfigure}
    \begin{subfigure}{0.33\textwidth}
        \centering
        \includegraphics[width=\textwidth]{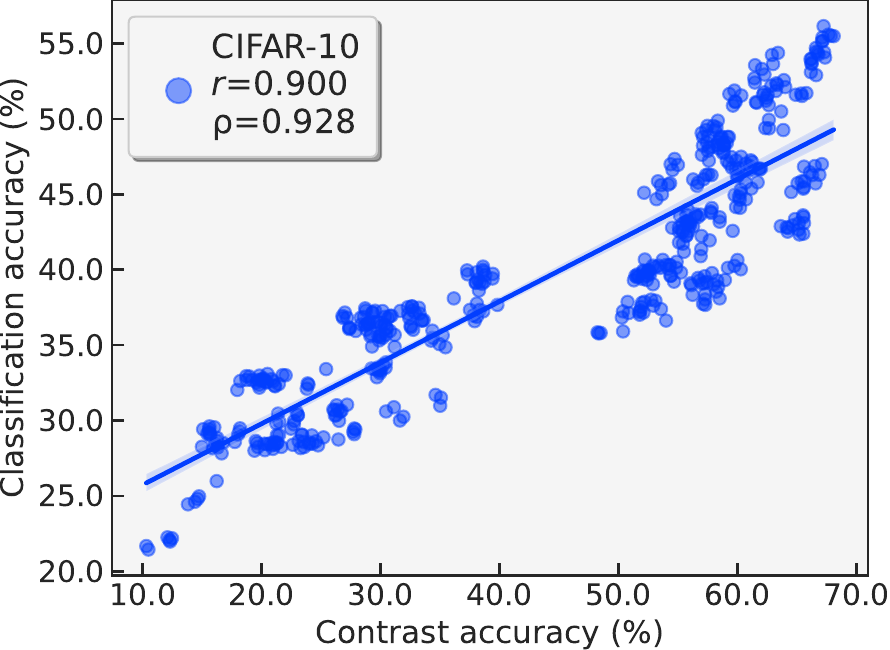}
        \caption{Pretain + fine-tune}
    \end{subfigure}
    
    \caption{Scatter plots of the linear correlation with training ways.}
    \label{training_ways}
\end{figure*}

In this section, we display the scatter plots of linear correlations under different training settings, as follows in Figure \ref{conloss_weight_app}, \ref{RRC_app}, \ref{brightness_app}, \ref{contrast_app}, \ref{saturation_app}, \ref{hue_app}, \ref{backbone_app}, \ref{seed_app}, \ref{sample_set_num_app}, \ref{sample_set_size_app}, \ref{training_ways}.

\end{document}